\pdfoutput=1    
\documentclass{article}

\usepackage[numbers]{natbib}
\usepackage[final]{neurips_2019} 

\usepackage[utf8]{inputenc}

\usepackage[labelfont=bf]{caption} 

\usepackage{bold-extra} 

\usepackage{textcomp} 

\usepackage{changepage} 

\usepackage{float} 
\usepackage{afterpage} 

\usepackage{array} 

\usepackage{hyperref} 

\usepackage[
	activate={true,nocompatibility},
	tracking=true
]{microtype}

\usepackage{xcolor}

\definecolor{myLinkColor}{rgb}{0.18,0.39,0.62}
\hypersetup{
    colorlinks=true,
    linkcolor=myLinkColor,
    filecolor=myLinkColor,
    urlcolor=myLinkColor,
    citecolor=myLinkColor,
}

\usepackage[disable]{todonotes} 
\usepackage{graphicx} 
\usepackage{tabularx} 
\usepackage{booktabs} 

\usepackage[titles]{tocloft}
\usepackage{caption} 
\usepackage{chngcntr}
\counterwithin{figure}{section}
\counterwithin{table}{section}

\title{Language Models are Few-Shot Learners}
\author{
Tom B. Brown\thanks{Equal contribution}
\And
Benjamin Mann\footnotemark[1]
\And
Nick Ryder\footnotemark[1]
\And
Melanie Subbiah\footnotemark[1]
\AND
Jared Kaplan\thanks{Johns Hopkins University, OpenAI\newline\newline Author contributions \hyperref[sec:contributions]{listed at end of paper}.}
\And
Prafulla Dhariwal
\And
Arvind Neelakantan
\And
Pranav Shyam
\And
Girish Sastry
\And
Amanda Askell
\And
Sandhini Agarwal
\And
Ariel Herbert-Voss
\And
Gretchen Krueger
\And
Tom Henighan
\And
Rewon Child
\And
Aditya Ramesh
\And
Daniel M. Ziegler
\And
Jeffrey Wu
\And
Clemens Winter
\AND
Christopher Hesse
\And
Mark Chen
\And
Eric Sigler
\And
Mateusz Litwin
\And
Scott Gray
\AND
Benjamin Chess
\And
Jack Clark
\And
Christopher Berner
\AND
Sam McCandlish
\And
Alec Radford
\And
Ilya Sutskever
\And
Dario Amodei 
\AND
\\
{\large OpenAI}
}
\date{2020}

\begin{document}

\maketitle
\begin{abstract}
   Recent work has demonstrated substantial gains on many NLP tasks and benchmarks by pre-training on a large corpus of text followed by fine-tuning on a specific task. While typically task-agnostic in architecture, this method still requires task-specific fine-tuning datasets of thousands or tens of thousands of examples. By contrast, humans can generally perform a new language task from only a few examples or from simple instructions – something which current NLP systems still largely struggle to do. Here we show that scaling up language models greatly improves task-agnostic, few-shot performance, sometimes even reaching competitiveness with prior state-of-the-art fine-tuning approaches. Specifically, we train GPT-3, an autoregressive language model with 175 billion parameters, 10x more than any previous non-sparse language model, and test its performance in the few-shot setting.  For all tasks, GPT-3 is applied without any gradient updates or fine-tuning, with tasks and few-shot demonstrations specified purely via text interaction with the model.  GPT-3 achieves strong performance on many NLP datasets, including translation, question-answering, and cloze tasks, as well as several tasks that require on-the-fly reasoning or domain adaptation, such as unscrambling words, using a novel word in a sentence, or performing 3-digit arithmetic. At the same time, we also identify some datasets where GPT-3's few-shot learning still struggles, as well as some datasets where GPT-3 faces methodological issues related to training on large web corpora. Finally, we find that GPT-3 can generate samples of news articles which human evaluators have difficulty distinguishing from articles written by humans.  We discuss broader societal impacts of this finding and of GPT-3 in general.
\end{abstract}

\newpage
\setcounter{tocdepth}{2}
\tableofcontents
\newpage

\setcounter{footnote}{0}
\makeatletter
\let\@fnsymbol\@arabic
\makeatother

%
%
\section{Introduction}
\label{section:Introduction}
Recent years have featured a trend towards pre-trained language representations in NLP systems, applied in increasingly flexible and task-agnostic ways for downstream transfer.  First, single-layer representations were learned using word vectors \cite{mikolov2013efficient, pennington2014glove} and fed to task-specific architectures, then RNNs with multiple layers of representations and contextual state were used to form stronger representations \cite{dai2015semi, mccann2017learned, peters2018dissecting} (though still applied to task-specific architectures), and more recently pre-trained recurrent or transformer language models \cite{vaswani2017attention}  have been directly fine-tuned, entirely removing the need for task-specific architectures \cite{radford2018gpt1, devlin2018bert, howard2018universal}.

\begin{figure}[b!]
\centering\includegraphics[width=1.0\linewidth]{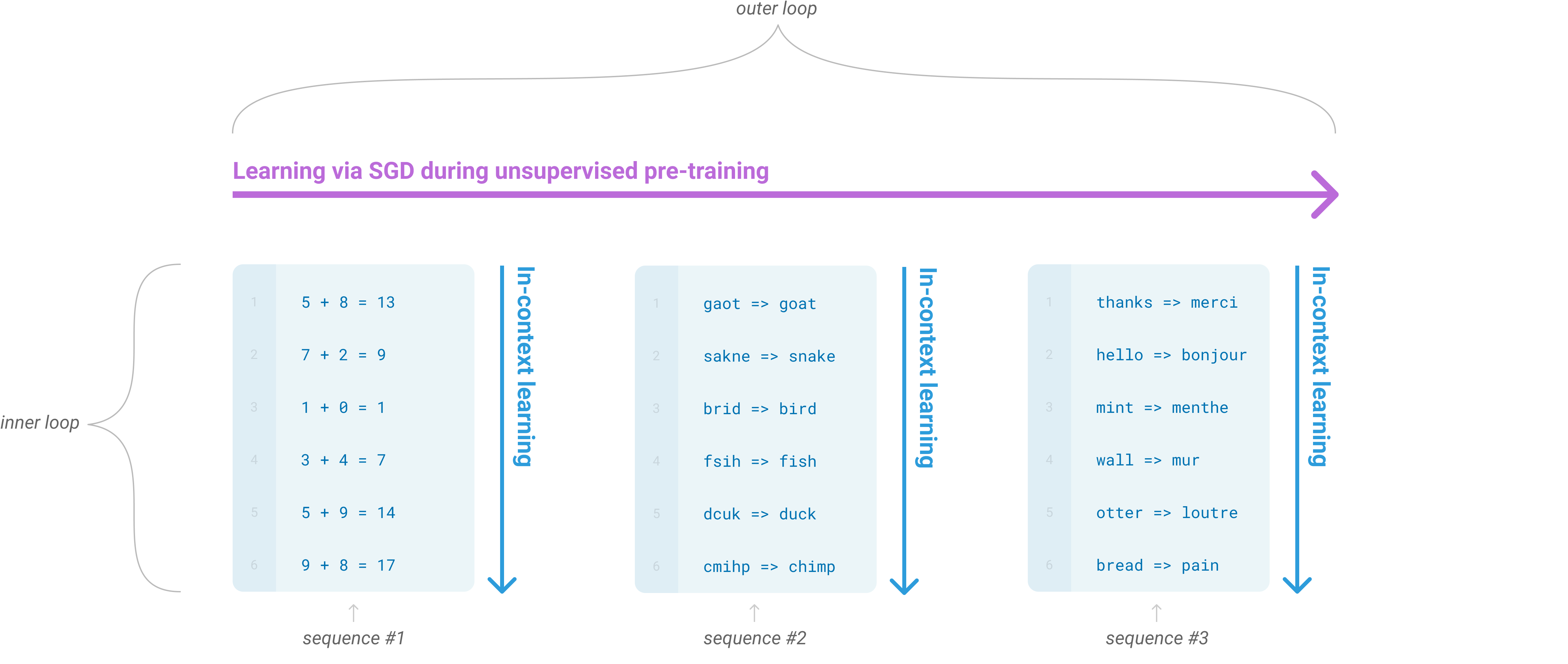}
\caption{\textbf{Language model meta-learning.} During unsupervised pre-training, a language model develops a broad set of skills and pattern recognition abilities. It then uses these abilities at inference time to rapidly adapt to or recognize the desired task. We use the term ``in-context learning" to describe the inner loop of this process, which occurs within the forward-pass upon each sequence. The sequences in this diagram are not intended to be representative of the data a model would see during pre-training, but are intended to show that there are sometimes repeated sub-tasks embedded within a single sequence.
}
\label{figure:metalearning}
\end{figure}

This last paradigm has led to substantial progress on many challenging NLP tasks such as reading comprehension, question answering, textual entailment, and many others, and has continued to advance based on new architectures and algorithms \cite{raffel2019t5, liu2019roberta, yang2019xlnet, lan2019albert}. However, a major limitation to this approach is that while the architecture is task-agnostic, there is still a need for task-specific datasets and task-specific fine-tuning: to achieve strong performance on a desired task typically requires fine-tuning on a dataset of thousands to hundreds of thousands of examples specific to that task.  Removing this limitation would be desirable, for several reasons.

First, from a practical perspective, the need for a large dataset of labeled examples for every new task limits the applicability of language models.  There exists a very wide range of possible useful language tasks, encompassing anything from correcting grammar, to generating examples of an abstract concept, to critiquing a short story.  For many of these tasks it is difficult to collect a large supervised training dataset, especially when the process must be repeated for every new task.

Second, the potential to exploit spurious correlations in training data fundamentally grows with the expressiveness of the model and the narrowness of the training distribution. This can create problems for the pre-training plus fine-tuning paradigm, where models are designed to be large to absorb information during pre-training, but are then fine-tuned on very narrow task distributions. For instance \cite{hendrycks2020pretrained} observe that larger models do not necessarily generalize better out-of-distribution. There is evidence that suggests that the generalization achieved under this paradigm can be poor because the model is overly specific to the training distribution and does not generalize well outside it \cite{yogatama2019learning, mccoy2019right}. Thus, the performance of fine-tuned models on specific benchmarks, even when it is nominally at human-level, may exaggerate actual performance on the underlying task \cite{gururangan2018annotation, niven2019probing}.

Third, humans do not require large supervised datasets to learn most language tasks -- a brief directive in natural language (e.g. ``please tell me if this sentence describes something happy or something sad'') or at most a tiny number of demonstrations (e.g. ``here are two examples of people acting brave; please give a third example of bravery'') is often sufficient to enable a human to perform a new task to at least a reasonable degree of competence.  Aside from pointing to a conceptual limitation in our current NLP techniques, this adaptability has practical advantages -- it allows humans to seamlessly mix together or switch between many tasks and skills, for example performing addition during a lengthy dialogue.  To be broadly useful, we would someday like our NLP systems to have this same fluidity and generality.

One potential route towards addressing these issues is meta-learning\footnote{In the context of language models this has sometimes been called ``zero-shot transfer'', but this term is potentially ambiguous: the method is ``zero-shot'' in the sense that no gradient updates are performed, but it often involves providing inference-time demonstrations to the model, so is not truly learning from zero examples.  To avoid this confusion, we use the term ``meta-learning'' to capture the inner-loop / outer-loop structure of the general method, and the term ``in context-learning" to refer to the inner loop of meta-learning. We further specialize the description to ``zero-shot", ``one-shot", or ``few-shot" depending on how many demonstrations are provided at inference time.  These terms are intended to remain agnostic on the question of whether the model learns new tasks from scratch at inference time or simply recognizes patterns seen during training -- this is an important issue which we discuss later in the paper, but ``meta-learning'' is intended to encompass both possibilities, and simply describes the inner-outer loop structure.} -- which in the context of language models means the model develops a broad set of skills and pattern recognition abilities at training time, and then uses those abilities at inference time to rapidly adapt to or recognize the desired task (illustrated in Figure \ref{figure:metalearning}). Recent work \cite{radford2019language} attempts to do this via what we call ``in-context learning", using the text input of a pretrained language model as a form of task specification: the model is conditioned on a natural language instruction and/or a few demonstrations of the task and is then expected to complete further instances of the task simply by predicting what comes next.

While it has shown some initial promise, this approach still achieves results far inferior to fine-tuning -- for example \cite{radford2019language} achieves only 4\% on Natural Questions, and even its 55 F1 CoQa result is now more than 35 points behind the state of the art.  Meta-learning clearly requires substantial improvement in order to be viable as a practical method of solving language tasks.

Another recent trend in language modeling may offer a way forward.  In recent years the capacity of transformer language models has increased substantially, from 100 million parameters \cite{radford2018gpt1}, to 300 million parameters \cite{devlin2018bert}, to 1.5 billion parameters \cite{radford2019language}, to 8 billion parameters \cite{shoeybi2019megatronlm}, 11 billion parameters \cite{raffel2019t5}, and finally 17 billion parameters \cite{turing_17m}.  Each increase has brought improvements in text synthesis and/or downstream NLP tasks, and there is evidence suggesting that log loss, which correlates well with many downstream tasks, follows a smooth trend of improvement with scale \cite{kaplan2020scaling}.  Since in-context learning involves absorbing many skills and tasks within the parameters of the model, it is plausible that in-context learning abilities might show similarly strong gains with scale.

\begin{figure}
\centering\includegraphics[width=0.8\linewidth]{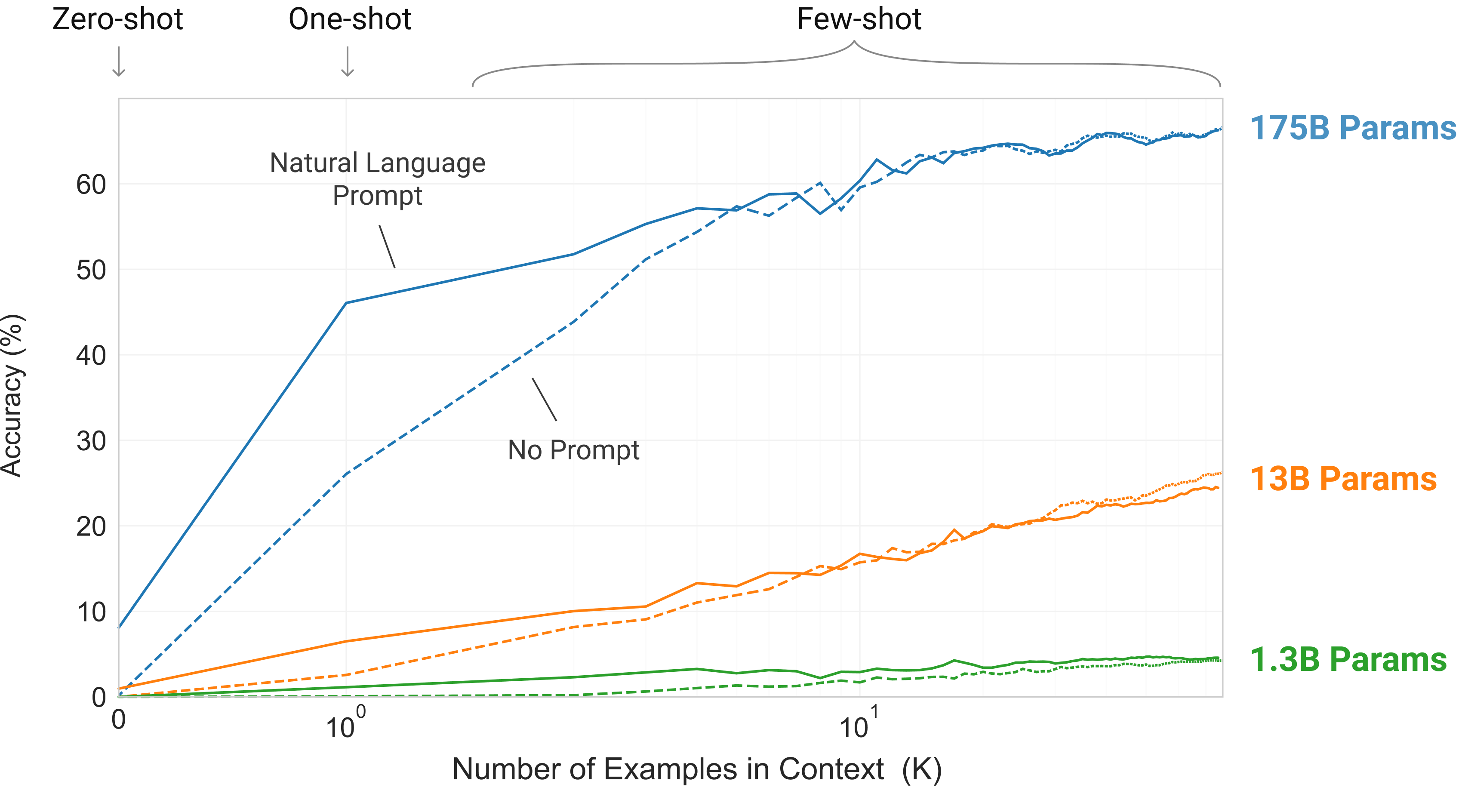}
\caption{\textbf{Larger models make increasingly efficient use of in-context information.}~~~We show in-context learning performance on a simple task requiring the model to remove random symbols from a word, both with and without a natural language task description (see Sec. \ref{section:Word_Scrambling_and_Manipulation_Tasks}). The steeper ``in-context learning curves'' for large models demonstrate improved ability to learn a task from contextual information. We see qualitatively similar behavior across a wide range of tasks.}
\label{graph:scramble_prompted}
\end{figure}

\begin{figure}
\begin{center}
\includegraphics[width=0.7\linewidth]{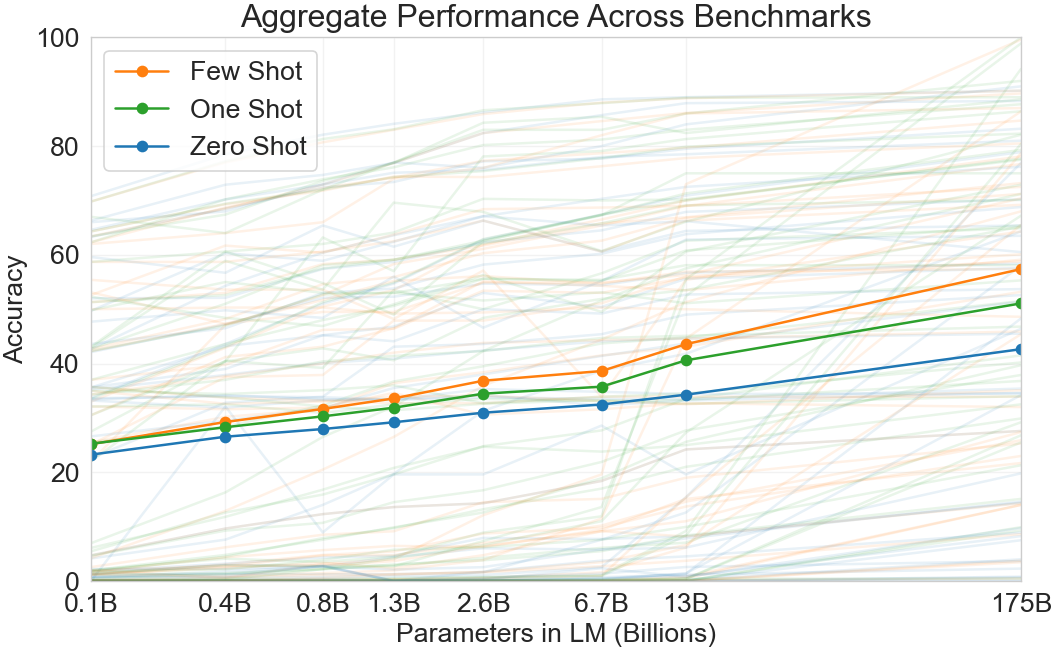}
\end{center}
\caption{\textbf{Aggregate performance for all 42 accuracy-denominated benchmarks}~~~While zero-shot performance improves steadily with model size, few-shot performance increases more rapidly, demonstrating that larger models are more proficient at in-context learning. See Figure \ref{graph:superglue_analysis} for a more detailed analysis on SuperGLUE, a standard NLP benchmark suite.}
\label{figure:aggregate_performance}
\end{figure}

In this paper, we test this hypothesis by training a 175 billion parameter autoregressive language model, which we call GPT-3, and measuring its in-context learning abilities.  Specifically, we evaluate GPT-3 on over two dozen NLP datasets, as well as several novel tasks designed to test rapid adaptation to tasks unlikely to be directly contained in the training set.  For each task, we evaluate GPT-3 under 3 conditions: (a) ``few-shot learning'', or in-context learning where we allow as many demonstrations as will fit into the model’s context window (typically 10 to 100), (b) ``one-shot learning'', where we allow only one demonstration, and (c) ``zero-shot'' learning, where no demonstrations are allowed and only an instruction in natural language is given to the model.  GPT-3 could also in principle be evaluated in the traditional fine-tuning setting, but we leave this to future work.

 Figure \ref{graph:scramble_prompted} illustrates the conditions we study, and shows few-shot learning of a simple task requiring the model to remove extraneous symbols from a word. Model performance improves with the addition of a natural language task description, and with the number of examples in the model's context, $K$. Few-shot learning also improves dramatically with model size.  Though the results in this case are particularly striking, the general trends with both model size and number of examples in-context hold for most tasks we study.  We emphasize that these ``learning'' curves involve no gradient updates or fine-tuning, just increasing numbers of demonstrations given as conditioning.

Broadly, on NLP tasks GPT-3 achieves promising results in the zero-shot and one-shot settings, and in the the few-shot setting is sometimes competitive with or even occasionally surpasses state-of-the-art (despite state-of-the-art being held by fine-tuned models).  For example, GPT-3 achieves 81.5 F1 on CoQA in the zero-shot setting, 84.0 F1 on CoQA in the one-shot setting, 85.0 F1 in the few-shot setting.  Similarly, GPT-3 achieves 64.3\% accuracy on TriviaQA in the zero-shot setting, 68.0\% in the one-shot setting, and 71.2\% in the few-shot setting, the last of which is state-of-the-art relative to fine-tuned models operating in the same closed-book setting.

GPT-3 also displays one-shot and few-shot proficiency at tasks designed to test rapid adaption or on-the-fly reasoning, which include unscrambling words, performing arithmetic, and using novel words in a sentence after seeing them defined only once.  We also show that in the few-shot setting, GPT-3 can generate synthetic news articles which human evaluators have difficulty distinguishing from human-generated articles.

At the same time, we also find some tasks on which few-shot performance struggles, even at the scale of GPT-3.  This includes natural language inference tasks like the ANLI dataset, and some reading comprehension datasets like RACE or QuAC.  By presenting a broad characterization of GPT-3's strengths and weaknesses, including these limitations, we hope to stimulate study of few-shot learning in language models and draw attention to where progress is most needed.  

A heuristic sense of the overall results can be seen in Figure \ref{figure:aggregate_performance}, which aggregates the various tasks (though it should not be seen as a rigorous or meaningful benchmark in itself).

We also undertake a systematic study of ``data contamination'' -- a growing problem when training high capacity models on datasets such as Common Crawl, which can potentially include content from test datasets simply because such content often exists on the web. In this paper we develop systematic tools to measure data contamination and quantify its distorting effects.  Although we find that data contamination has a minimal effect on GPT-3's performance on most datasets, we do identify a few datasets where it could be inflating results, and we either do not report results on these datasets or we note them with an asterisk, depending on the severity.

In addition to all the above, we also train a series of smaller models (ranging from 125 million parameters to 13 billion parameters) in order to compare their performance to GPT-3 in the zero, one and few-shot settings.  Broadly, for most tasks we find relatively smooth scaling with model capacity in all three settings; one notable pattern is that the gap between zero-, one-, and few-shot performance often grows with model capacity, perhaps suggesting that larger models are more proficient meta-learners.

Finally, given the broad spectrum of capabilities displayed by GPT-3, we discuss concerns about bias, fairness, and broader societal impacts, and attempt a preliminary analysis of GPT-3's characteristics in this regard.

The remainder of this paper is organized as follows.  In Section \ref{section:Approach}, we describe our approach and methods for training GPT-3 and evaluating it.  Section \ref{section:Results} presents results on the full range of tasks in the zero-, one- and few-shot settings.  Section \ref{section:measuring_and_preventing_memorization_of_benchmarks} addresses questions of data contamination (train-test overlap).  Section \ref{section:Limitations} discusses limitations of GPT-3.  Section \ref{section:Broader_Impacts} discusses broader impacts.  Section \ref{section:Related Work} reviews related work and Section \ref{section:Conclusion} concludes.

%
%
\section{Approach}
\label{section:Approach}
\begin{figure}
\centering\includegraphics[width=0.8\linewidth]{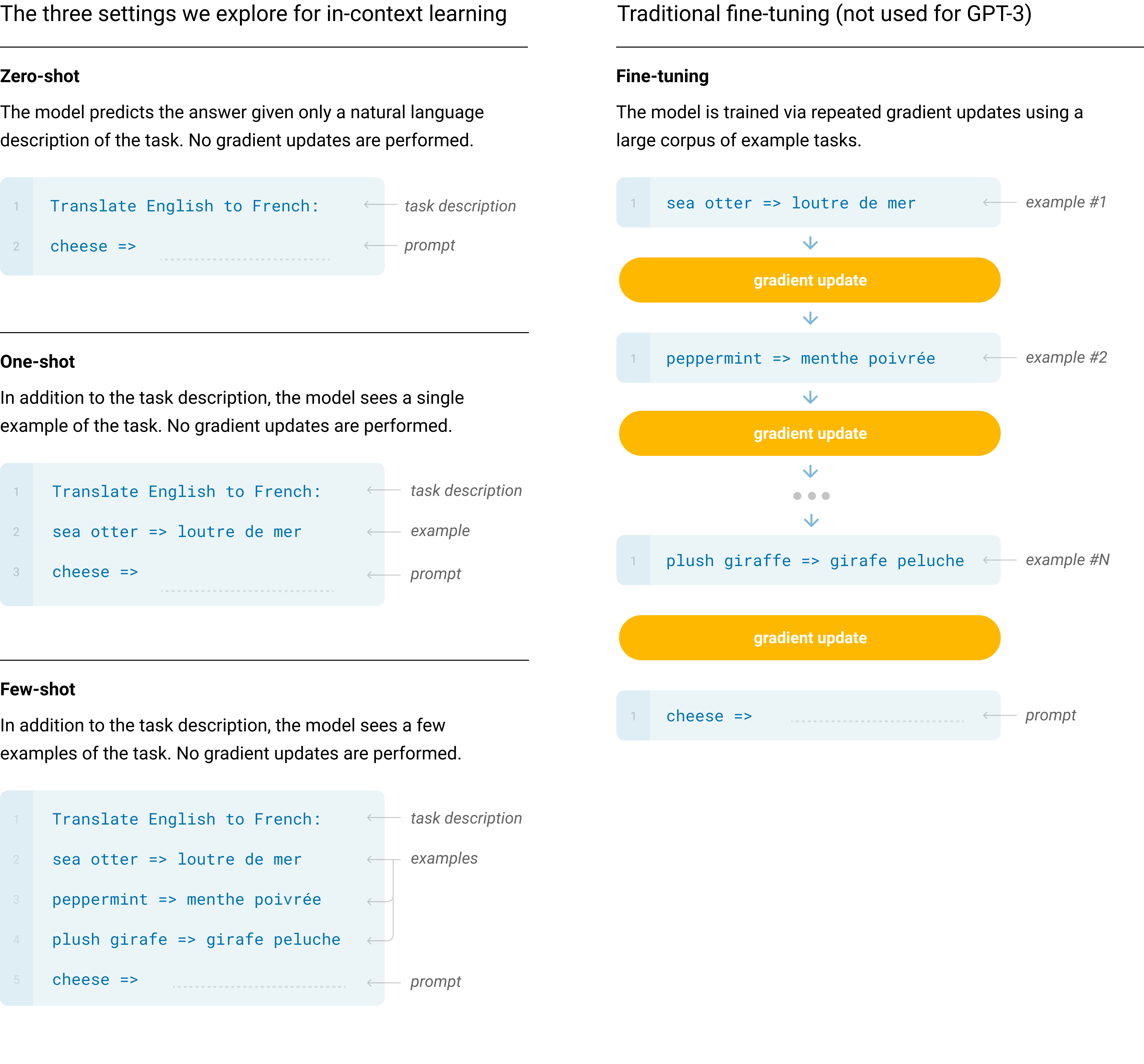}
\caption{\textbf{Zero-shot, one-shot and few-shot, contrasted with traditional fine-tuning}. The panels above show four methods for performing a task with a language model -- fine-tuning is the traditional method, whereas zero-, one-, and few-shot, which we study in this work, require the model to perform the task with only forward passes at test time. We typically present the model with a few dozen examples in the few shot setting. Exact phrasings for all task descriptions, examples and prompts can be found in Appendix \ref{appendix:task_phrasing}.
}
\label{figure:eval strategies}
\end{figure}

Our basic pre-training approach, including model, data, and training, is similar to the process described in \cite{radford2019language}, with relatively straightforward scaling up of the model size, dataset size and diversity, and length of training.  Our use of in-context learning is also similar to \cite{radford2019language}, but in this work we systematically explore different settings for learning within the context.  Therefore, we start this section by explicitly defining and contrasting the different settings that we will be evaluating GPT-3 on or could in principle evaluate GPT-3 on. These settings can be seen as lying on a spectrum of how much task-specific data they tend to rely on.  Specifically, we can identify at least four points on this spectrum (see Figure \ref{figure:eval strategies} for an illustration):

\begin{itemize}
    \item \textbf{Fine-Tuning (FT)} has been the most common approach in recent years, and involves updating the weights of a pre-trained model by training on a supervised dataset specific to the desired task.  Typically thousands to hundreds of thousands of labeled examples are used.  The main advantage of fine-tuning is strong performance on many benchmarks. The main disadvantages are the need for a new large dataset for every task, the potential for poor generalization out-of-distribution \cite{mccoy2019right}, and the potential to exploit spurious features of the training data \cite{ gururangan2018annotation, niven2019probing}, potentially resulting in an unfair comparison with human performance. In this work we do not fine-tune GPT-3 because our focus is on task-agnostic performance, but GPT-3 can be fine-tuned in principle and this is a promising direction for future work.
    \item \textbf{Few-Shot (FS)} is the term we will use in this work to refer to the setting where the model is given a few demonstrations of the task at inference time as conditioning \cite{radford2019language}, but no weight updates are allowed.  As shown in Figure \ref{figure:eval strategies}, for a typical dataset an example has a context and a desired completion (for example an English sentence and the French translation), and few-shot works by giving $K$ examples of context and completion, and then one final example of context, with the model expected to provide the completion.  We typically set $K$ in the range of 10 to 100 as this is how many examples can fit in the model’s context window ($n_{\mathrm{ctx}}=2048$).  The main advantages of few-shot are a major reduction in the need for task-specific data and reduced potential to learn an overly narrow distribution from a large but narrow fine-tuning dataset. The main disadvantage is that results from this method have so far been much worse than state-of-the-art fine-tuned models.  Also, a small amount of task specific data is still required. As indicated by the name, few-shot learning as described here for language models is related to few-shot learning as used in other contexts in ML~\citep{hochreiter2001learning, vinyals2016matching} -- both involve learning based on a broad distribution of tasks (in this case implicit in the pre-training data) and then rapidly adapting to a new task.
    \item \textbf{One-Shot (1S)} is the same as few-shot except that only one demonstration is allowed, in addition to a natural language description of the task, as shown in Figure 1. The reason to distinguish one-shot from few-shot and zero-shot (below) is that it most closely matches the way in which some tasks are communicated to humans.  For example, when asking humans to generate a dataset on a human worker service (for example Mechanical Turk), it is common to give one demonstration of the task.  By contrast it is sometimes difficult to communicate the content or format of a task if no examples are given.
    \item \textbf{Zero-Shot (0S)} is the same as one-shot except that no demonstrations are allowed, and the model is only given a natural language instruction describing the task.  This method provides maximum convenience, potential for robustness, and avoidance of spurious correlations (unless they occur very broadly across the large corpus of  pre-training data), but is also the most challenging setting.  In some cases it may even be difficult for humans to understand the format of the task without prior examples, so this setting is in some cases ``unfairly hard''.  For example, if someone is asked to ``make a table of world records for the 200m dash'', this request can be ambiguous, as it may not be clear exactly what format the table should have or what should be included (and even with careful clarification, understanding precisely what is desired can be difficult).  Nevertheless, for at least some settings zero-shot is closest to how humans perform tasks -- for example, in the translation example in Figure \ref{figure:eval strategies}, a human would likely know what to do from just the text instruction.
\end{itemize}

Figure \ref{figure:eval strategies} shows the four methods using the example of translating English to French.  In this paper we focus on zero-shot, one-shot and few-shot, with the aim of comparing them not as competing alternatives, but as different problem settings which offer a varying trade-off between performance on specific benchmarks and sample efficiency.  We especially highlight the few-shot results as many of them are only slightly behind state-of-the-art fine-tuned models.  Ultimately, however, one-shot, or even sometimes zero-shot, seem like the fairest comparisons to human performance, and are important targets for future work.

Sections \ref{section:Model and Architectures}-\ref{section:Training Process} below give details on our models, training data, and training process respectively.  
Section \ref{section:Evaluation} discusses the details of how we do few-shot, one-shot, and zero-shot evaluations.

    \subsection{Model and Architectures}
    \label{section:Model and Architectures}
    We use the same model and architecture as GPT-2 \cite{radford2019language}, including the modified initialization, pre-normalization, and reversible tokenization described therein, with the exception that we use alternating dense and locally banded sparse attention patterns in the layers of the transformer, similar to the Sparse Transformer \cite{child2019generating}. To study the dependence of ML performance on model size, we train 8 different sizes of model, ranging over three orders of magnitude from 125 million parameters to 175 billion parameters, with the last being the model we call GPT-3.  Previous work \cite{kaplan2020scaling} suggests that with enough training data, scaling of validation loss should be approximately a smooth power law as a function of size; training models of many different sizes allows us to test this hypothesis both for validation loss and for downstream language tasks.

\begin{table}
    
    \begin{adjustwidth}{-.8in}{-.8in}
    \begin{center}
        
        \begin{tabular}{l c c c c c c c}
        \toprule
        Model Name & $n_{\mathrm{params}}$ & $n_{\mathrm{layers}}$ & $d_{\mathrm{model}}$ & $n_{\mathrm{heads}}$ & $d_{\mathrm{head}}$  & Batch Size & Learning Rate \\
        \midrule
GPT-3 Small & 125M & 12 & 768 & 12 & 64 & 0.5M & $6.0 \times 10^{-4}$ \\                              
GPT-3 Medium & 350M & 24 & 1024 & 16 & 64 & 0.5M & $3.0 \times 10^{-4}$ \\                             
GPT-3 Large & 760M & 24 & 1536 & 16 & 96 & 0.5M & $2.5 \times 10^{-4}$ \\                             
GPT-3 XL & 1.3B & 24 & 2048 & 24 & 128 & 1M & $2.0 \times 10^{-4}$ \\                            
GPT-3 2.7B   & 2.7B & 32 & 2560 & 32 & 80 & 1M & $1.6 \times 10^{-4}$ \\                             
GPT-3 6.7B   & 6.7B & 32 & 4096 & 32 & 128 & 2M & $1.2 \times 10^{-4}$ \\                            
GPT-3 13B  & 13.0B & 40 & 5140 & 40 & 128 & 2M & $1.0 \times 10^{-4}$ \\                           
GPT-3 175B or ``GPT-3'' & 175.0B & 96 & 12288 & 96 & 128 & 3.2M & $0.6 \times 10^{-4}$ \\
        \bottomrule
        \end{tabular}
    \end{center}
    \end{adjustwidth}
    \caption{Sizes, architectures, and learning hyper-parameters (batch size in tokens and learning rate) of the models which we trained. All models were trained for a total of 300 billion tokens.}
    \label{table:param}
\end{table}

Table \ref{table:param} shows the sizes and architectures of our 8 models.  Here $n_{\mathrm{params}}$ is the total number of trainable parameters, $n_{\mathrm{layers}}$ is the total number of layers, $d_{\mathrm{model}}$ is the number of units in each bottleneck layer (we always have the feedforward layer four times the size of the bottleneck layer, $d_{\mathrm{ff}}$ $= 4 \ast d_{\mathrm{model}}$), and $d_{\mathrm{head}}$ is the dimension of each attention head.  All models use a context window of $n_{\mathrm{ctx}}=2048$ tokens.  We partition the model across GPUs along both the depth and width dimension in order to minimize data-transfer between nodes. The precise architectural parameters for each model are chosen based on computational efficiency and load-balancing in the layout of models across GPU’s.  Previous work \cite{kaplan2020scaling} suggests that validation loss is not strongly sensitive to these parameters within a reasonably broad range.

\begin{figure}
\centering\includegraphics[width=0.9\linewidth]{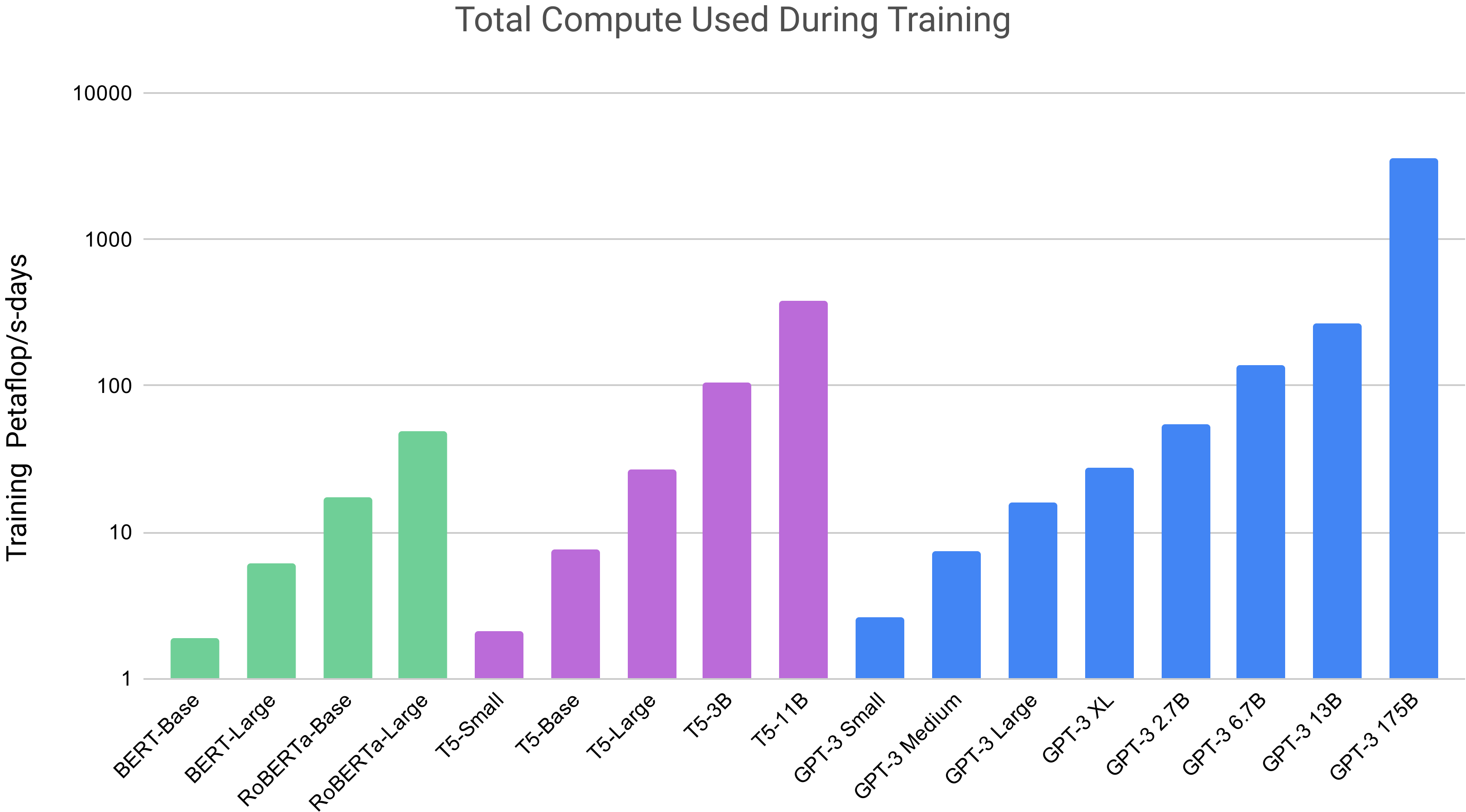}
\caption{\textbf{Total compute used during training}. Based on the analysis in Scaling Laws For Neural Language Models \cite{kaplan2020scaling} we train much larger models on many fewer tokens than is typical. As a consequence, although GPT-3 3B is almost 10x larger than RoBERTa-Large (355M params), both models took roughly 50 petaflop/s-days of compute during pre-training. Methodology for these calculations can be found in Appendix \ref{appendix:total_compute_calculations}.
}
\label{figure:training flops}
\end{figure}

    \subsection{Training Dataset}
    \label{section:Training Dataset}
    Datasets for language models have rapidly expanded, culminating in the Common Crawl dataset\footnote{\url{https://commoncrawl.org/the-data/}} \cite{raffel2019t5} constituting nearly a trillion words.  This size of dataset is sufficient to train our largest models without ever updating on the same sequence twice. However, we have found that unfiltered or lightly filtered versions of Common Crawl tend to have lower quality than more curated datasets.  Therefore, we took 3 steps to improve the average quality of our datasets: (1) we downloaded and filtered a version of CommonCrawl based on similarity to a range of high-quality reference corpora, (2) we performed fuzzy deduplication at the document level, within and across datasets, to prevent redundancy and preserve the integrity of our held-out validation set as an accurate measure of overfitting, and (3) we also added known high-quality reference corpora to the training mix to augment CommonCrawl and increase its diversity.

Details of the first two points (processing of Common Crawl) are described in Appendix \ref{appendix:common_crawl_filtering}. For the third, we added several curated high-quality datasets, including an expanded version of the WebText dataset \cite{radford2019language}, collected by scraping links over a longer period of time, and first described in \cite{kaplan2020scaling}, two internet-based books corpora (Books1 and Books2) and English-language Wikipedia.

Table \ref{table:dataset} shows the final mixture of datasets that we used in training.  The CommonCrawl data was downloaded from 41 shards of monthly CommonCrawl covering 2016 to 2019, constituting 45TB of compressed plaintext before filtering and 570GB after filtering, roughly equivalent to 400 billion byte-pair-encoded tokens.    Note that during training, datasets are not sampled in proportion to their size, but rather datasets we view as higher-quality are sampled more frequently, such that CommonCrawl and Books2 datasets are sampled less than once during training, but the other datasets are sampled 2-3 times. This essentially accepts a small amount of overfitting in exchange for higher quality training data.

\begin{table}

    \begin{center}
        \begin{tabular}{lccc}
        \toprule
        Dataset & \shortstack{Quantity \\ (tokens)} & \shortstack{Weight in \\training mix} & \shortstack{Epochs elapsed when \\ training for 300B tokens} \\ 
        \midrule
        Common Crawl (filtered) & 410 billion & 60\% & 0.44\\ 
        WebText2 & 19 billion & 22\% &  2.9 \\
        Books1 & 12 billion & 8\% & 1.9 \\ 
        Books2 & 55 billion & 8\% & 0.43 \\ 
        Wikipedia & 3 billion & 3\% & 3.4 \\ 
        \bottomrule
        \end{tabular}
    \end{center}
    \caption{\textbf{Datasets used to train GPT-3}.  ``Weight in training mix'' refers to the fraction of examples during training that are drawn from a given dataset, which we intentionally do not make proportional to the size of the dataset.  As a result, when we train for 300 billion tokens, some datasets are seen up to 3.4 times during training while other datasets are seen less than once.}
    \label{table:dataset}
\end{table}

A major methodological concern with language models pretrained on a broad swath of internet data, particularly large models with the capacity to memorize vast amounts of content, is potential contamination of downstream tasks by having their test or development sets inadvertently seen during pre-training.  To reduce such contamination, we searched for and attempted to remove any overlaps with the development and test sets of all benchmarks studied in this paper.  Unfortunately, a bug in the filtering caused us to ignore some overlaps, and due to the cost of training it was not feasible to retrain the model. In Section \ref{section:measuring_and_preventing_memorization_of_benchmarks} we characterize the impact of the remaining overlaps, and in future work we will more aggressively remove data contamination. 

    \subsection{Training Process}
    \label{section:Training Process}
    As found in \cite{kaplan2020scaling, mcc2018batchsize}, larger models can typically use a larger batch size, but require a smaller learning rate. We measure the gradient noise scale during training and use it to guide our choice of batch size \cite{mcc2018batchsize}. Table \ref{table:param} shows the parameter settings we used. To train the larger models without running out of memory, we use a mixture of model parallelism within each matrix multiply and model parallelism across the layers of the network.   All models were trained on V100 GPU’s on part of a high-bandwidth cluster provided by Microsoft.  Details of the training process and hyperparameter settings are described in Appendix \ref{appendix:model_training}.

    \subsection{Evaluation}
    \label{section:Evaluation}
    For few-shot learning, we evaluate each example in the evaluation set by randomly drawing $K$ examples from that task’s training set as conditioning, delimited by 1 or 2 newlines depending on the task.  For LAMBADA and Storycloze there is no supervised training set available so we draw conditioning examples from the development set and evaluate on the test set.  For Winograd (the original, not SuperGLUE version) there is only one dataset, so we draw conditioning examples directly from it.

$K$ can be any value from 0 to the maximum amount allowed by the model’s context window, which is $n_{\mathrm{ctx}}=2048$ for all models and typically fits $10$ to $100$ examples.  Larger values of $K$ are usually but not always better, so when a separate development and test set are available, we experiment with a few values of $K$ on the development set and then run the best value on the test set.  For some tasks (see Appendix \ref{appendix:task_phrasing}) we also use a natural language prompt in addition to (or for $K=0$, instead of) demonstrations.

On tasks that involve choosing one correct completion from several options (multiple choice), we provide $K$ examples of context plus correct completion, followed by one example of context only, and compare the LM likelihood of each completion.  For most tasks we compare the per-token likelihood (to normalize for length), however on a small number of datasets (ARC, OpenBookQA, and RACE) we gain additional benefit as measured on the development set by normalizing by the unconditional probability of each completion, by computing $\frac{P(\mathrm{completion} | \mathrm{context})}{P(\mathrm{completion} | \mathrm{answer\_context})}$, where $\mathrm{answer\_context}$ is the string \texttt{"Answer: "} or \texttt{"A: "} and is used to prompt that the completion should be an answer but is otherwise generic.

On tasks that involve binary classification, we give the options more semantically meaningful names (e.g. ``True" or ``False" rather than 0 or 1) and then treat the task like multiple choice; we also sometimes frame the task similar to what is done by \cite{raffel2019t5} (see Appendix \ref{appendix:task_phrasing}) for details.

On tasks with free-form completion, we use beam search with the same parameters as \cite{raffel2019t5}: a beam width of 4 and a length penalty of $\alpha = 0.6$.  We score the model using F1 similarity score, BLEU, or exact match, depending on what is standard for the dataset at hand.

Final results are reported on the test set when publicly available, for each model size and learning setting (zero-, one-, and few-shot).  When the test set is private, our model is often too large to fit on the test server, so we report results on the development set.  We do submit to the test server on a small number of datasets (SuperGLUE, TriviaQA, PiQa) where we were able to make submission work, and we submit only the 200B few-shot results, and report development set results for everything else.

%
%
\section{Results}
\label{section:Results}
\begin{figure}
\centering\includegraphics[width=0.7\linewidth]{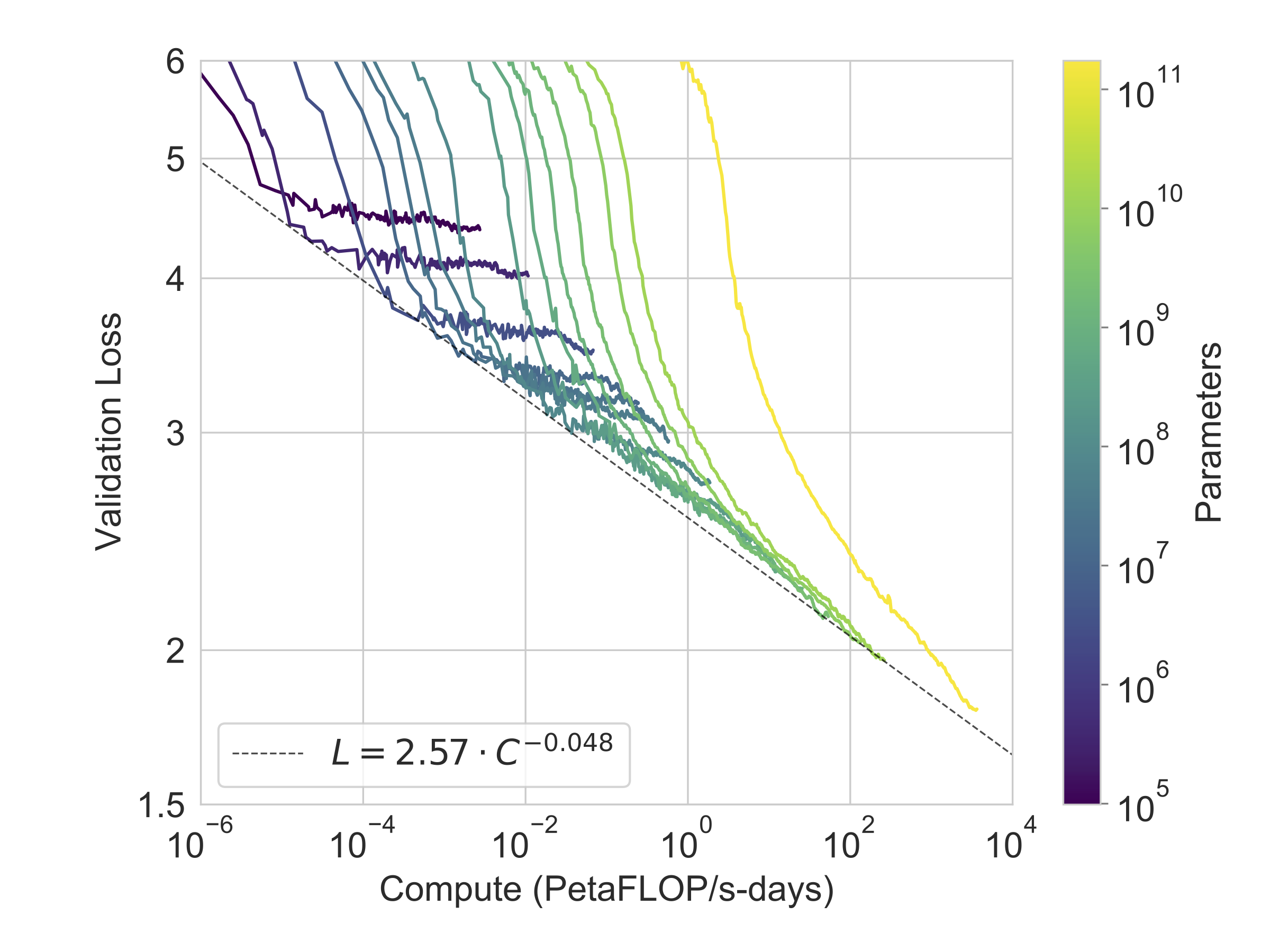}
\caption{\textbf{Smooth scaling of performance with compute.} Performance (measured in terms of cross-entropy validation loss) follows a power-law trend with the amount of compute used for training. The power-law behavior observed in \cite{kaplan2020scaling} continues for an additional two orders of magnitude with only small deviations from the predicted curve.  For this figure, we exclude embedding parameters from compute and parameter counts.}
\label{graph:compute}
\end{figure}

In Figure \ref{graph:compute} we display training curves for the 8 models described in Section \ref{section:Approach}. For this graph we also include 6 additional extra-small models with as few as 100,000 parameters. As observed in \cite{kaplan2020scaling}, language modeling performance follows a power-law when making efficient use of training compute. After extending this trend by two more orders of magnitude, we observe only a slight (if any) departure from the power-law. One might worry that these improvements in cross-entropy loss come only from modeling spurious details of our training corpus. However, we will see in the following sections that improvements in cross-entropy loss lead to consistent performance gains across a broad spectrum of natural language tasks.

Below, we evaluate the 8 models described in Section \ref{section:Approach} (the 175 billion parameter parameter GPT-3 and 7 smaller models) on a wide range of datasets. We group the datasets into 9 categories representing roughly similar tasks.  

In Section \ref{section:Language Modeling, Cloze, and Completion Tasks} we evaluate on traditional language modeling tasks and tasks that are similar to language modeling, such as Cloze tasks and sentence/paragraph completion tasks.  In Section \ref{section:Closed Book Question Answering / Knowledge Base Tasks} we evaluate on ``closed book'' question answering tasks: tasks which require using the information stored in the model’s parameters to answer general knowledge questions.  In Section \ref{section:Translation} we evaluate the model’s ability to translate between languages (especially one-shot and few-shot).  In Section \ref{section:Winograd-Style_Tasks} we evaluate the model’s performance on Winograd Schema-like tasks.  In Section \ref{section:Common_Sense_Reasoning} we evaluate on datasets that involve commonsense reasoning or question answering.  In Section \ref{section:Reading_Comprehension} we evaluate on reading comprehension tasks, in Section \ref{section:SuperGLUE} we evaluate on the SuperGLUE benchmark suite, and in \ref{section:ANLI} we briefly explore NLI.  Finally, in Section \ref{section:Synthetic_and_Qualitative_Tasks}, we invent some additional tasks designed especially to probe in-context learning abilities -- these tasks focus on on-the-fly reasoning, adaptation skills, or open-ended text synthesis. We evaluate all tasks in the few-shot, one-shot, and zero-shot settings.

    \subsection{Language Modeling, Cloze, and Completion Tasks}
    \label{section:Language Modeling, Cloze, and Completion Tasks}
    In this section we test GPT-3’s performance on the traditional task of language modeling, as well as related tasks that involve predicting a single word of interest, completing a sentence or paragraph, or choosing between possible completions of a piece of text.

        \subsubsection{Language Modeling}
        \label{section:Language Modeling}
        We calculate zero-shot perplexity on the Penn Tree Bank (PTB) \cite{marcus1994penn} dataset measured in \cite{radford2019language}. We omit the 4 Wikipedia-related tasks in that work because they are entirely contained in our training data, and we also omit the one-billion word benchmark due to a high fraction of the dataset being contained in our training set.  PTB escapes these issues due to predating the modern internet. Our largest model sets a new SOTA on PTB by a substantial margin of 15 points, achieving a perplexity of 20.50. Note that since PTB is a traditional language modeling dataset it does not have a clear separation of examples to define one-shot or few-shot evaluation around, so we measure only zero-shot.

\begin{table}
    
    \centering
        \begin{tabular}{ll}
        \toprule
        Setting & PTB  \\ 
        \midrule
        SOTA (Zero-Shot) & 35.8\textsuperscript{\textit{a}} \\
        GPT-3 Zero-Shot & \textbf{20.5}\\
        \bottomrule
        \end{tabular}
        
    \caption{\textbf{Zero-shot results on PTB language modeling dataset.} Many other common language modeling datasets are omitted because they are derived from Wikipedia or other sources which are included in GPT-3's training data. \textsuperscript{\textit{a}}\cite{radford2019language}}
    \label{table:language}
\end{table}
        
        \subsubsection{LAMBADA}
        \label{section:LAMBADA}
        \begin{table}
    \centering
    \begin{center}
        \begin{tabular}{l c c c c c c}
        \toprule
        Setting & \shortstack{LAMBADA\\(acc)} & \shortstack{LAMBADA\\(ppl)} & \shortstack{StoryCloze\\(acc)} & \shortstack{HellaSwag\\(acc)}\\ 
        \midrule
        SOTA & 68.0\textsuperscript{\textit{a}} & 8.63\textsuperscript{\textit{b}}  & \textbf{91.8}\textsuperscript{\textit{c}} & \textbf{85.6}\textsuperscript{\textit{d}}\\ 
        GPT-3 Zero-Shot & \textbf{76.2} & \textbf{3.00}   & 83.2 & 78.9\\
        GPT-3 One-Shot & \textbf{72.5} &\textbf{3.35} &  84.7 & 78.1\\
        GPT-3 Few-Shot & \textbf{86.4} & \textbf{1.92} & 87.7 & 79.3\\
        \bottomrule
        \end{tabular}
    \end{center}
    \caption{\textbf{Performance on cloze and completion tasks.} GPT-3 significantly improves SOTA on LAMBADA while achieving respectable performance on two difficult completion prediction datasets. \textsuperscript{\textit{a}}\cite{turing_17m} \textsuperscript{\textit{b}}\cite{radford2019language} \textsuperscript{\textit{c}}\cite{li2019story} \textsuperscript{\textit{d}}\cite{liu2020alum} }
    \label{table:completion}
\end{table}
\begin{figure}
\vspace{-1em}\centering\includegraphics[width=0.8\linewidth]{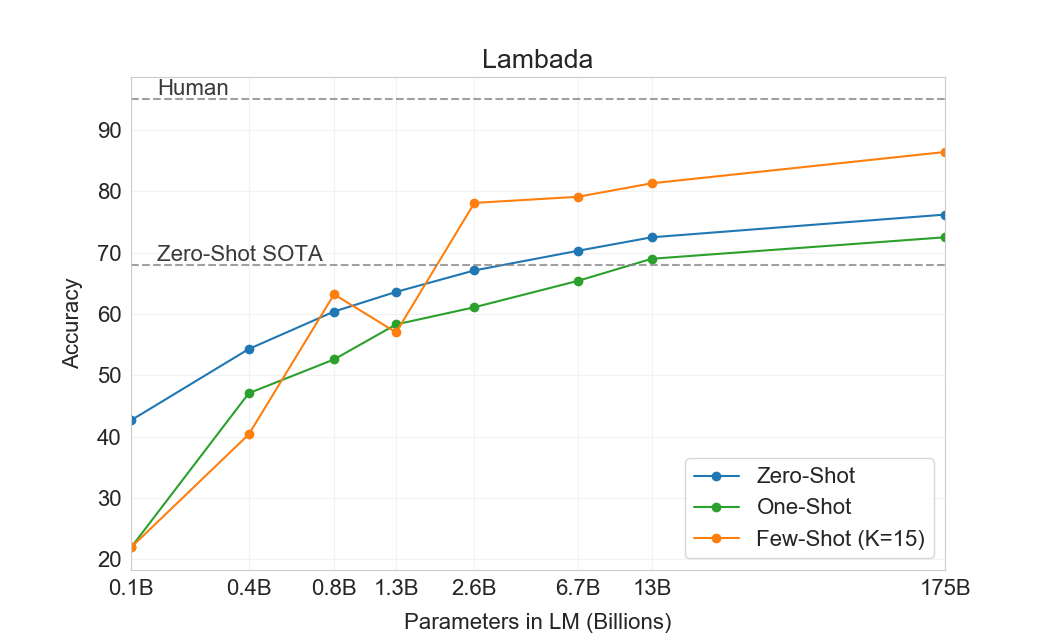}
\caption{On LAMBADA, the few-shot capability of language models results in a strong boost to accuracy. GPT-3 2.7B outperforms the SOTA 17B parameter Turing-NLG \cite{turing_17m} in this setting, and GPT-3 175B advances the state of the art by 18\%.  Note zero-shot uses a different format from one-shot and few-shot as described in the text.}
\label{graph:lambada}
\end{figure}

The LAMBADA dataset \cite{paperno2016lambada} tests the modeling of long-range dependencies in text -- the model is asked to predict the last word of sentences which require reading a paragraph of context. It has recently been suggested that the continued scaling of language models is yielding diminishing returns on this difficult benchmark. \cite{bisk2020experience} reflect on the small 1.5\% improvement achieved by a doubling of model size between two recent state of the art results (\cite{shoeybi2019megatronlm} and \cite{turing_17m}) and argue that ``continuing to expand hardware and data sizes by orders of magnitude is not the path forward''. We find that path is still promising and in a zero-shot setting GPT-3 achieves 76\% on LAMBADA, a gain of 8\% over the previous state of the art.

LAMBADA is also a demonstration of the flexibility of few-shot learning as it provides a way to address a problem that classically occurs with this dataset. Although the completion in LAMBADA is always the last word in a sentence, a standard language model has no way of knowing this detail. It thus assigns probability not only to the correct ending but also to other valid continuations of the paragraph. This problem has been partially addressed in the past with stop-word filters \cite{radford2019language} (which ban ``continuation'' words). The few-shot setting instead allows us to ``frame'' the task as a cloze-test and allows the language model to infer from examples that a completion of exactly one word is desired. We use the following fill-in-the-blank format:

\hspace{3cm}Alice was friends with Bob.  Alice went to visit her friend \underline{\hspace{1cm}}. $\to$ Bob

\hspace{3cm}George bought some baseball equipment, a ball, a glove, and a \underline{\hspace{1cm}}. $\to$ 

When presented with examples formatted this way, GPT-3 achieves 86.4\% accuracy in the few-shot setting, an increase of over 18\% from the previous state-of-the-art. We observe that few-shot performance improves strongly with model size. While this setting decreases the performance of the smallest model by almost 20\%, for GPT-3 it improves accuracy by 10\%. Finally, the fill-in-blank method is not effective one-shot, where it always performs worse than the zero-shot setting. Perhaps this is because all models still require several examples to recognize the pattern.

One note of caution is that an analysis of test set contamination identified that a significant minority of the LAMBADA dataset appears to be present in our training data -- however analysis performed in Section \ref{section:measuring_and_preventing_memorization_of_benchmarks} suggests negligible impact on performance.
        
        \subsubsection{HellaSwag}
        \label{section:HellaSwag}
        The HellaSwag dataset \cite{zellers2019hellaswag} involves picking the best ending to a story or set of instructions.  The examples were adversarially mined to be difficult for language models while remaining easy for humans (who achieve 95.6\% accuracy).  GPT-3 achieves 78.1\% accuracy in the one-shot setting and 79.3\% accuracy in the few-shot setting, outperforming the 75.4\% accuracy of a fine-tuned 1.5B parameter language model \cite{zellers2019defending} but still a fair amount lower than the overall SOTA of 85.6\% achieved by the fine-tuned multi-task model ALUM.

        \subsubsection{StoryCloze}
        \label{section:StoryCloze}
        We next evaluate GPT-3 on the StoryCloze 2016 dataset \cite{mostafazadeh2016corpus}, which involves selecting the correct ending sentence for five-sentence long stories.  Here GPT-3 achieves 83.2\% in the zero-shot setting and 87.7\% in the few-shot setting (with $K=70$). This is still 4.1\% lower than the fine-tuned SOTA using a BERT based model \cite{li2019story} but improves over previous zero-shot results by roughly 10\%.

    \subsection{Closed Book Question Answering}
    \label{section:Closed Book Question Answering / Knowledge Base Tasks}
    \begin{table}
    \centering
        \begin{tabular}{l l l l}
        \toprule
        Setting & NaturalQS & WebQS & TriviaQA \\ 
        \midrule
        RAG (Fine-tuned, Open-Domain) \cite{lewis2020retrieval} & \textbf{44.5} & \textbf{45.5} & \textbf{68.0} \\
        T5-11B+SSM (Fine-tuned, Closed-Book) \cite{roberts2020much} & 36.6 & 44.7 & 60.5 \\ 
        T5-11B (Fine-tuned, Closed-Book) & 34.5  & 37.4 & 50.1 \\
        GPT-3 Zero-Shot & 14.6 & 14.4 & 64.3 \\
        GPT-3 One-Shot & 23.0 & 25.3 & \textbf{68.0} \\
        GPT-3 Few-Shot & 29.9 & 41.5 & \textbf{71.2} \\
        \bottomrule
        \end{tabular}
    \caption{\textbf{Results on three Open-Domain QA tasks.} GPT-3 is shown in the few-, one-, and zero-shot settings, as compared to prior SOTA results for closed book and open domain settings.  TriviaQA few-shot result is evaluated on the wiki split test server.}
    \label{table:question}
\end{table}

\begin{figure}
\vspace{-1em}\centering\includegraphics[width=0.8\linewidth]{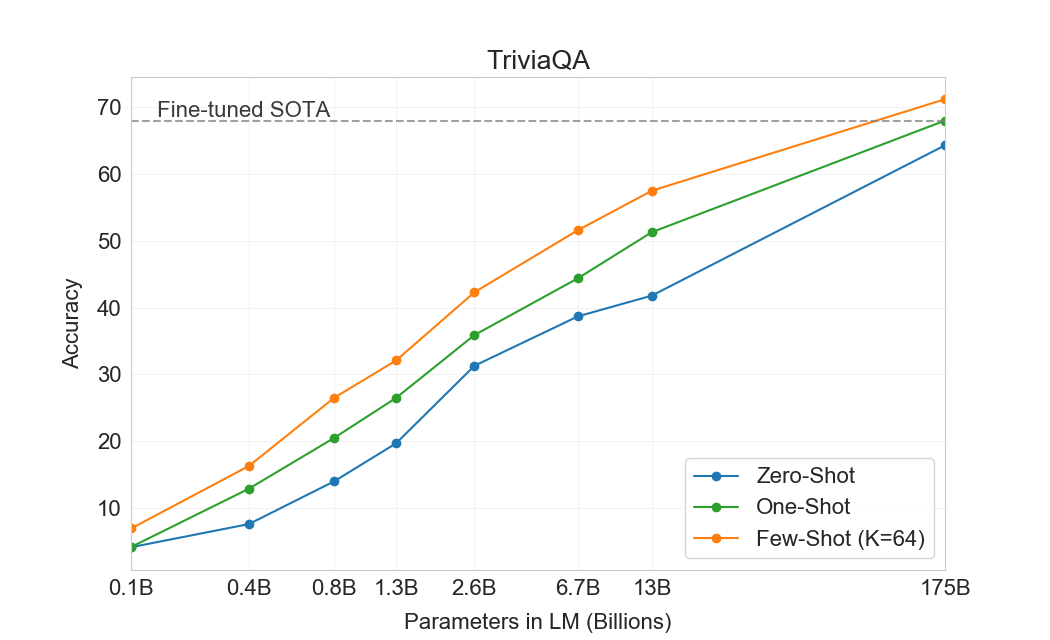}
\caption{On TriviaQA GPT3's performance grows smoothly with model size, suggesting that language models continue to absorb knowledge as their capacity increases.  One-shot and few-shot performance make significant gains over zero-shot behavior, matching and exceeding the performance of the SOTA fine-tuned open-domain model, RAG \cite{lewis2020retrieval}}
\label{graph:triviaqa}
\end{figure}

In this section we measure GPT-3’s ability to answer questions about broad factual knowledge. Due to the immense amount of possible queries, this task has normally been approached by using an information retrieval system to find relevant text in combination with a model which learns to generate an answer given the question and the retrieved text. Since this setting allows a system to search for and condition on text which potentially contains the answer it is denoted ``open-book''. \cite{roberts2020much} recently demonstrated that a large language model can perform surprisingly well directly answering the questions without conditioning on auxilliary information. They denote this more restrictive evaluation setting as ``closed-book''.  Their work suggests that even higher-capacity models could perform even better and we test this hypothesis with GPT-3.   We evaluate GPT-3 on the 3 datasets in \cite{roberts2020much}: Natural Questions \cite{Kwiatkowski2019nq}, WebQuestions \cite{berant2013semantic}, and TriviaQA \cite{joshi2017triviaqa}, using the same splits.  Note that in addition to all results being in the closed-book setting, our use of few-shot, one-shot, and zero-shot evaluations represent an even stricter setting than previous closed-book QA work: in addition to external content not being allowed, fine-tuning on the Q\&A dataset itself is also not permitted.

The results for GPT-3 are shown in Table \ref{table:question}. On TriviaQA, we achieve 64.3\% in the zero-shot setting, 68.0\% in the one-shot setting, and 71.2\% in the few-shot setting.  The zero-shot result already outperforms the fine-tuned T5-11B by 14.2\%, and also outperforms a version with Q\&A tailored span prediction during pre-training by 3.8\%.  The one-shot result improves by 3.7\% and matches the SOTA for an open-domain QA system which not only fine-tunes but also makes use of a learned retrieval mechanism over a 15.3B parameter dense vector index of 21M documents \cite{lewis2020retrieval}. GPT-3's few-shot result further improves performance another 3.2\% beyond this.

On WebQuestions (WebQs), GPT-3 achieves 14.4\% in the zero-shot setting, 25.3\% in the one-shot setting, and 41.5\% in the few-shot setting.  This compares to 37.4\% for fine-tuned T5-11B, and 44.7\% for fine-tuned T5-11B+SSM, which uses a Q\&A-specific pre-training procedure.  GPT-3 in the few-shot setting approaches the performance of state-of-the-art fine-tuned models.  Notably, compared to TriviaQA, WebQS shows a much larger gain from zero-shot to few-shot (and indeed its zero-shot and one-shot performance are poor), perhaps suggesting that the WebQs questions and/or the style of their answers are out-of-distribution for GPT-3.  Nevertheless, GPT-3 appears able to adapt to this distribution, recovering strong performance in the few-shot setting.

On Natural Questions (NQs) GPT-3 achieves 14.6\% in the zero-shot setting, 23.0\% in the one-shot setting, and 29.9\% in the few-shot setting, compared to 36.6\% for fine-tuned T5 11B+SSM.  Similar to WebQS, the large gain from zero-shot to few-shot may suggest a distribution shift, and may also explain the less competitive performance compared to TriviaQA and WebQS.  In particular, the questions in NQs tend towards very fine-grained knowledge on Wikipedia specifically which could be testing the limits of GPT-3's capacity and broad pretraining distribution.

Overall, on one of the three datasets GPT-3's one-shot matches the open-domain fine-tuning SOTA. On the other two datasets it approaches the performance of the closed-book SOTA despite not using fine-tuning.  On all 3 datasets, we find that performance scales very smoothly with model size (Figure \ref{graph:triviaqa} and Appendix \ref{appendix:results_on_all_tasks} Figure \ref{graph:all_qa}), possibly reflecting the idea that model capacity translates directly to more ‘knowledge’ absorbed in the parameters of the model.

    \subsection{Translation}
    \label{section:Translation}

For GPT-2 a filter was used on a multilingual collection of documents to produce an English only dataset due to capacity concerns. Even with this filtering GPT-2 showed some evidence of multilingual capability and performed non-trivially when translating between French and English despite only training on 10 megabytes of remaining French text. Since we increase the capacity by over two orders of magnitude from GPT-2 to GPT-3, we also expand the scope of the training dataset to include more representation of other languages, though this remains an area for further improvement. As discussed in \ref{section:Training Dataset} the majority of our data is derived from raw Common Crawl with only quality-based filtering. Although GPT-3's training data is still primarily English (93\% by word count), it also includes 7\% of text in other languages. These languages are documented in the \href{https://github.com/openai/gpt-3}{supplemental material}. In order to better understand translation capability, we also expand our analysis to include two additional commonly studied languages, German and Romanian.

Existing unsupervised machine translation approaches often combine pretraining on a pair of monolingual datasets with back-translation \cite{sennrich2015improving} to bridge the two languages in a controlled way. By contrast, GPT-3 learns from a blend of training data that mixes many languages together in a natural way, combining them on a word, sentence, and document level. GPT-3 also uses a single training objective which is not customized or designed for any task in particular. However, our one / few-shot settings aren't strictly comparable to prior unsupervised work since they make use of a small amount of paired examples (1 or 64). This corresponds to up to a page or two of in-context training data.

\begin{table}
    \centering
        \begin{tabular}{l c c c c c c}
        \toprule
        Setting & En$\to$Fr & Fr$\to$En & En$\to$De & De$\to$En & En$\to$Ro & Ro$\to$En \\ 
        \midrule
        SOTA (Supervised) & \textbf{45.6}\textsuperscript{\textit{a}} & 35.0 \textsuperscript{\textit{b}} & \textbf{41.2}\textsuperscript{\textit{c}} & 40.2\textsuperscript{\textit{d}} & \textbf{38.5}\textsuperscript{\textit{e}} & \textbf{39.9}\textsuperscript{\textit{e}} \\
        \midrule
        XLM \cite{lample2019cross} & 33.4 & 33.3 & 26.4 & 34.3 & 33.3 & 31.8 \\
        MASS \cite{song2019mass} & \underline{37.5} & 34.9 & 28.3 & 35.2 & \underline{35.2} & 33.1 \\
        mBART \cite{liu2020multilingual} & - & - & \underline{29.8} & 34.0 & 35.0 & 30.5 \\
        \midrule
        GPT-3 Zero-Shot  & 25.2 & 21.2 & 24.6 & 27.2 & 14.1 & 19.9 \\
        GPT-3 One-Shot & 28.3 & 33.7 & 26.2 & 30.4 & 20.6 & 38.6 \\
        GPT-3 Few-Shot & 32.6 & \underline{39.2} & 29.7 & \underline{40.6} & 21.0 & \underline{39.5} \\
        \bottomrule
        \end{tabular}
    \caption{\textbf{Few-shot GPT-3 outperforms previous unsupervised NMT work by 5 BLEU when translating into English reflecting its strength as an English LM.} We report BLEU scores on the WMT'14 Fr$\leftrightarrow$En, WMT’16 De$\leftrightarrow$En, and WMT'16 Ro$\leftrightarrow$En datasets as measured by \texttt{multi-bleu.perl} with XLM's tokenization in order to compare most closely with prior unsupervised NMT work. SacreBLEU\textsuperscript{\textit{f}} \cite{post2018call} results reported in Appendix \ref{appendix:results_on_all_tasks}. Underline indicates an unsupervised or few-shot SOTA, bold indicates supervised SOTA with relative confidence.
    \textsuperscript{\textit{a}}\cite{edunov2018understanding}
    \textsuperscript{\textit{b}}\cite{durrani2014edinburgh}
    \textsuperscript{\textit{c}}\cite{wang2018multi}
    \textsuperscript{\textit{d}}\cite{wmt16deensota} 
    \textsuperscript{\textit{e}}\cite{liu2020multilingual}
    \textsuperscript{\textit{f}}[SacreBLEU signature: BLEU+case.mixed+numrefs.1+smooth.exp+tok.intl+version.1.2.20] }
    \label{table:translation}
\end{table}
\begin{figure}
\centering\includegraphics[width=0.8\linewidth]{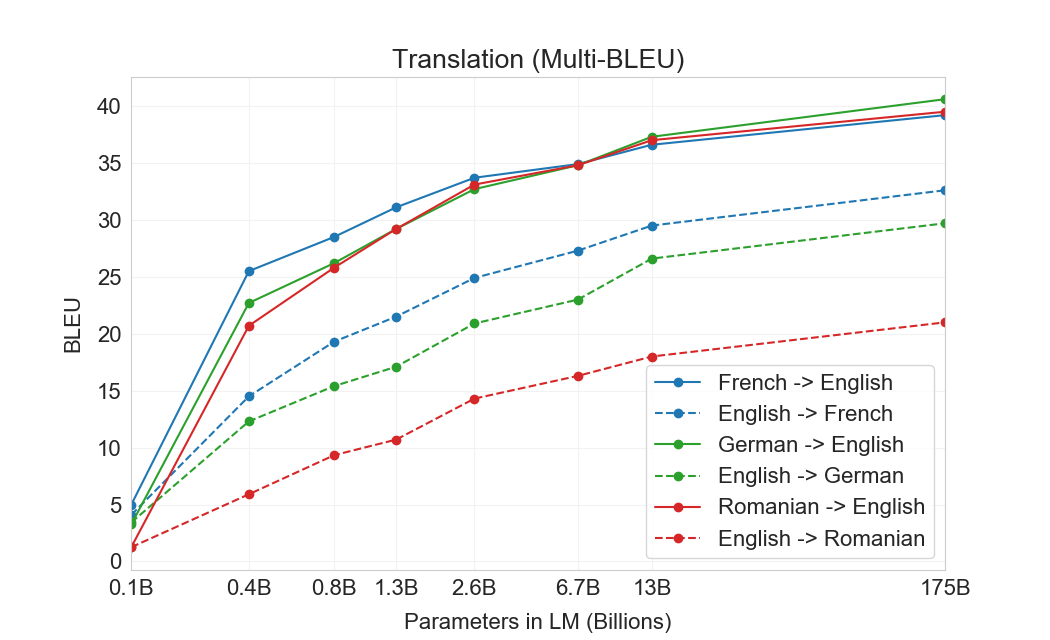}
\caption{Few-shot translation performance on 6 language pairs as model capacity increases.  There is a consistent trend of improvement across all datasets as the model scales, and as well as tendency for translation into English to be stronger than translation from English.}
\label{graph:translation}
\end{figure}

Results are shown in Table \ref{table:translation}. Zero-shot GPT-3, which only receives on a natural language description of the task, still underperforms recent unsupervised NMT results. However, providing only a single example demonstration for each translation task improves performance by over 7 BLEU and nears competitive performance with prior work. GPT-3 in the full few-shot setting further improves another 4 BLEU resulting in similar average performance to prior unsupervised NMT work. GPT-3 has a noticeable skew in its performance depending on language direction. For the three input languages studied, GPT-3 significantly outperforms prior unsupervised NMT work when translating into English but underperforms when translating in the other direction. Performance on En-Ro is a noticeable outlier at over 10 BLEU worse than prior unsupervised NMT work. This could be a weakness due to reusing the byte-level BPE tokenizer of GPT-2 which was developed for an almost entirely English training dataset. For both Fr-En and De-En, few shot GPT-3 outperforms the best supervised result we could find but due to our unfamiliarity with the literature and the appearance that these are un-competitive benchmarks we do not suspect those results represent true state of the art. For Ro-En, few shot GPT-3 performs within 0.5 BLEU of the overall SOTA which is achieved by a combination of unsupervised pretraining, supervised finetuning on 608K labeled examples, and backtranslation \cite{liu2019multi}.

Finally, across all language pairs and across all three settings (zero-, one-, and few-shot), there is a smooth trend of improvement with model capacity.  This is shown in Figure \ref{graph:translation} in the case of few-shot results, and scaling for all three settings is shown in Appendix \ref{appendix:results_on_all_tasks}.

    \subsection{Winograd-Style Tasks}
    \label{section:Winograd-Style_Tasks}
    \begin{table}
    \centering
    \begin{center}
        \begin{tabular}{l c c}
        \toprule
        Setting & Winograd & Winogrande (XL) \\ 
        \midrule
        Fine-tuned SOTA  & \textbf{90.1}\textsuperscript{\textit{a}}  & \textbf{84.6}\textsuperscript{\textit{b}} \\ 
        GPT-3 Zero-Shot  & 88.3* & 70.2 \\
        GPT-3 One-Shot & 89.7* & 73.2 \\
        GPT-3 Few-Shot  & 88.6* & 77.7 \\
        \bottomrule
        \end{tabular}
    \end{center}
    \caption{Results on the WSC273 version of Winograd schemas and the adversarial Winogrande dataset. See Section \ref{section:measuring_and_preventing_memorization_of_benchmarks} for details on potential contamination of the Winograd test set. \textsuperscript{\textit{a}}\cite{sakaguchi2019winogrande}
    \textsuperscript{\textit{b}}\cite{lin2020tttttackling}
    }
    \label{table:winograd}
\end{table}
\begin{figure}
\centering\includegraphics[width=0.8\linewidth]{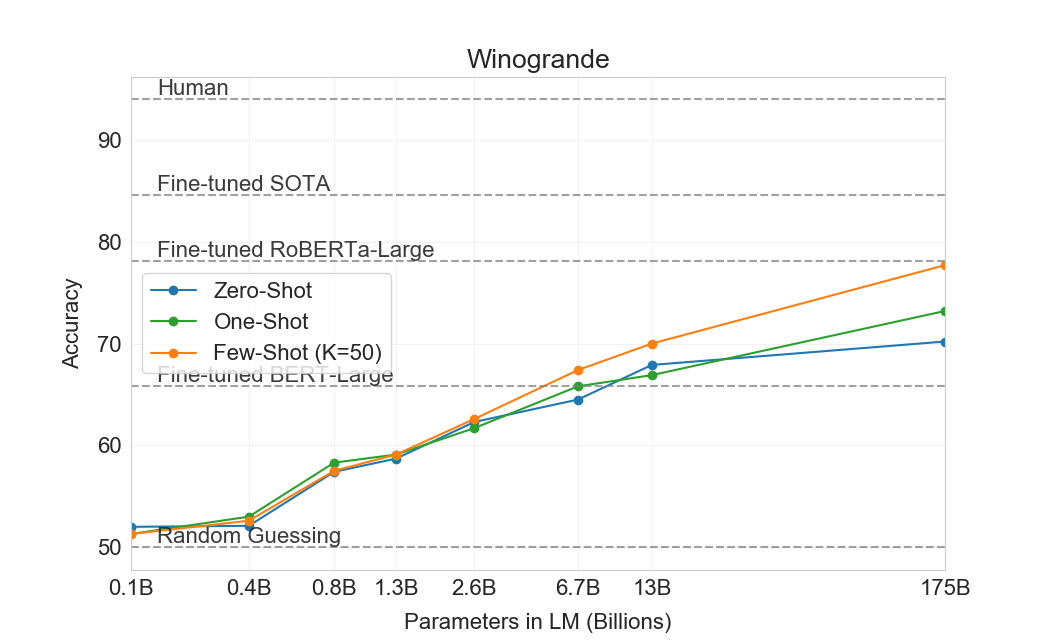}
\caption{Zero-, one-, and few-shot performance on the adversarial Winogrande dataset as model capacity scales.  Scaling is relatively smooth with the gains to few-shot learning increasing with model size, and few-shot GPT-3 175B is competitive with a fine-tuned RoBERTA-large.}
\label{graph:winogrande}
\end{figure}
The Winograd Schemas Challenge \cite{levesque2012winograd} is a classical task in NLP that involves determining which word a pronoun refers to, when the pronoun is grammatically ambiguous but semantically unambiguous to a human.  Recently fine-tuned language models have achieved near-human performance on the original Winograd dataset, but more difficult versions such as the adversarially-mined Winogrande dataset \cite{sakaguchi2019winogrande} still significantly lag human performance.  We test GPT-3’s performance on both Winograd and Winogrande, as usual in the zero-, one-, and few-shot setting.

On Winograd we test GPT-3 on the original set of 273 Winograd schemas, using the same ``partial evaluation'' method described in \cite{radford2019language}.  Note that this setting differs slightly from the WSC task in the SuperGLUE benchmark, which is presented as binary classification and requires entity extraction to convert to the form described in this section.  On Winograd GPT-3 achieves 88.3\%, 89.7\%, and 88.6\% in the zero-shot, one-shot, and few-shot settings, showing no clear in-context learning but in all cases achieving strong results just a few points below state-of-the-art and estimated human performance.  We note that contamination analysis found some Winograd schemas in the training data but this appears to have only a small effect on results (see Section \ref{section:measuring_and_preventing_memorization_of_benchmarks}).

On the more difficult Winogrande dataset, we do find gains to in-context learning: GPT-3 achieves 70.2\% in the zero-shot setting, 73.2\% in the one-shot setting, and 77.7\% in the few-shot setting.  For comparison a fine-tuned RoBERTA model achieves 79\%, state-of-the-art is 84.6\% achieved with a fine-tuned high capacity model (T5), and human performance on the task as reported by \cite{sakaguchi2019winogrande} is 94.0\%.

    \subsection{Common Sense Reasoning}
    \label{section:Common_Sense_Reasoning}
    
\begin{table}
    \centering
        \begin{tabular}{l l l l l}
        \toprule
        Setting & PIQA & ARC (Easy) & ARC (Challenge) & OpenBookQA \\ 
        \midrule
        Fine-tuned SOTA & 79.4 & \textbf{92.0}\cite{khashabi2020unifiedqa} & \textbf{78.5}\cite{khashabi2020unifiedqa}  & \textbf{87.2}\cite{khashabi2020unifiedqa}  \\ 
        GPT-3 Zero-Shot  & \textbf{80.5}* & 68.8 & 51.4 & 57.6 \\
        GPT-3 One-Shot & \textbf{80.5}* & 71.2 & 53.2 & 58.8 \\
        GPT-3 Few-Shot & \textbf{82.8}* & 70.1 & 51.5 &  65.4 \\
        \bottomrule
        \end{tabular}
    \caption{GPT-3 results on three commonsense reasoning tasks, PIQA, ARC, and OpenBookQA. GPT-3 Few-Shot PIQA result is evaluated on the test server.  See Section \ref{section:measuring_and_preventing_memorization_of_benchmarks} for details on potential contamination issues on the PIQA test set.}
    \label{table:reasoning}
\end{table}
\begin{figure}
\centering\includegraphics[width=0.8\linewidth]{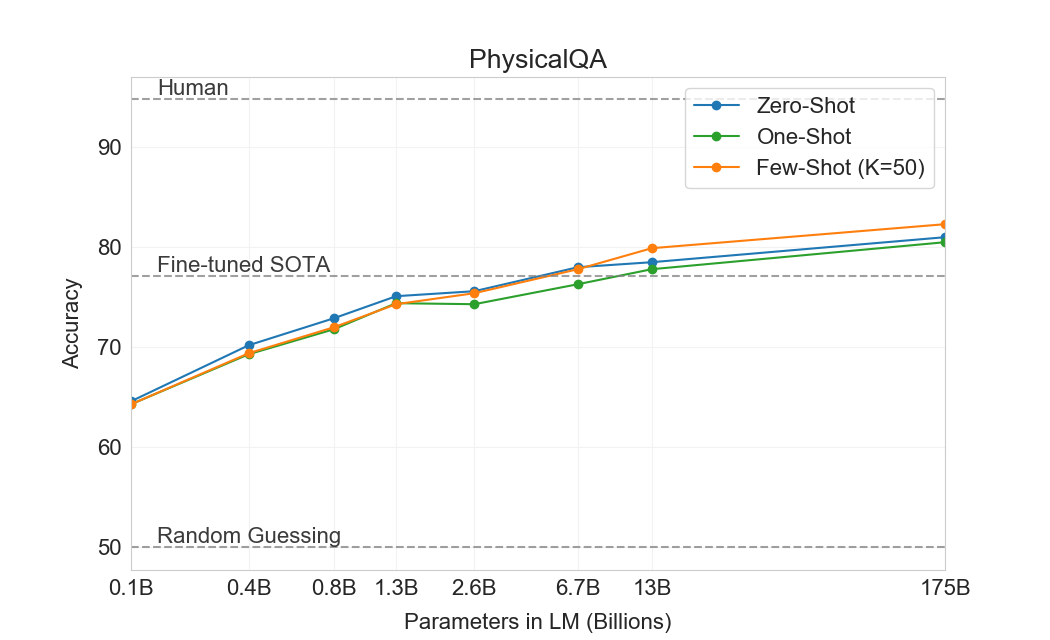}
\caption{GPT-3 results on PIQA in the zero-shot, one-shot, and few-shot settings. The largest model achieves a score on the development set in all three conditions that exceeds the best recorded score on the task.}
\label{graph:physicalqa}
\end{figure}
Next we consider three datasets which attempt to capture physical or scientific reasoning, as distinct from sentence completion, reading comprehension, or broad knowledge question answering. The first, PhysicalQA (PIQA) \cite{bisk2019piqa}, asks common sense questions about how the physical world works and is intended as a probe of grounded understanding of the world. GPT-3 achieves 81.0\% accuracy zero-shot, 80.5\% accuracy one-shot, and 82.8\% accuracy few-shot (the last measured on PIQA's test server).  This compares favorably to the 79.4\% accuracy prior state-of-the-art of a fine-tuned RoBERTa. PIQA shows relatively shallow scaling with model size and is still over 10\% worse than human performance, but GPT-3's few-shot and even zero-shot result outperform the current state-of-the-art.  Our analysis flagged PIQA for a potential data contamination issue (despite hidden test labels), and we therefore conservatively mark the result with an asterisk. See Section \ref{section:measuring_and_preventing_memorization_of_benchmarks} for details.

ARC \cite{Clark2018ThinkYH} is a dataset of multiple-choice questions collected from 3rd to 9th grade science exams. On the ``Challenge'' version of the dataset which has been filtered to questions which simple statistical or information retrieval methods are unable to  correctly answer, GPT-3 achieves 51.4\% accuracy in the zero-shot setting, 53.2\% in the one-shot setting, and 51.5\% in the few-shot setting. This is approaching the performance of a fine-tuned RoBERTa baseline (55.9\%) from UnifiedQA \cite{khashabi2020unifiedqa}. On the ``Easy'' version of the dataset (questions which either of the mentioned baseline approaches answered correctly), GPT-3 achieves 68.8\%, 71.2\%, and 70.1\% which slightly exceeds a fine-tuned RoBERTa baseline from \cite{khashabi2020unifiedqa}. However, both of these results are still much worse than the overall SOTAs achieved by the UnifiedQA which exceeds GPT-3’s few-shot results by 27\% on the challenge set and 22\% on the easy set.

On OpenBookQA \cite{Mihaylov2018CanAS}, GPT-3 improves significantly from zero to few shot settings but is still over 20 points short of the overall SOTA. GPT-3's few-shot performance is similar to a fine-tuned BERT Large baseline on the leaderboard.

Overall, in-context learning with GPT-3 shows mixed results on commonsense reasoning tasks, with only small and inconsistent gains observed in the one and few-shot learning settings for both PIQA and ARC, but a significant improvement is observed on OpenBookQA. GPT-3 sets SOTA on the new PIQA dataset in all evaluation settings.

    \subsection{Reading Comprehension}
    \label{section:Reading_Comprehension}
    \begin{table}
    \centering
    \begin{center}
        \begin{tabular}{lllllll}
        \toprule
        Setting & CoQA & DROP & QuAC & SQuADv2 & RACE-h & RACE-m\\ 
        \midrule
        Fine-tuned SOTA  & \textbf{90.7}\textsuperscript{\textit{a}} & \textbf{89.1}\textsuperscript{\textit{b}} & \textbf{74.4}\textsuperscript{\textit{c}} & \textbf{93.0}\textsuperscript{\textit{d}} & \textbf{90.0}\textsuperscript{\textit{e}} & \textbf{93.1}\textsuperscript{\textit{e}} \\ 
        GPT-3 Zero-Shot  & 81.5 & 23.6 & 41.5 & 59.5 & 45.5 & 58.4 \\
        GPT-3 One-Shot & 84.0 & 34.3 & 43.3 & 65.4 & 45.9 & 57.4 \\
        GPT-3 Few-Shot  & 85.0 & 36.5 & 44.3 & 69.8 & 46.8 & 58.1 \\
        \bottomrule
        \end{tabular}
        \end{center}
    \caption{Results on reading comprehension tasks. All scores are F1 except results for RACE which report accuracy. \textsuperscript{\textit{a}}\cite{ju2019technical} \textsuperscript{\textit{b}}\cite{dropsota} \textsuperscript{\textit{c}}\cite{quacsota} \textsuperscript{\textit{d}}\cite{squadv2sota} \textsuperscript{\textit{e}}\cite{shoeybi2019megatronlm} }
    \label{table:reading_comprehension}
\end{table}
\begin{figure}
\centering\includegraphics[width=0.8\linewidth]{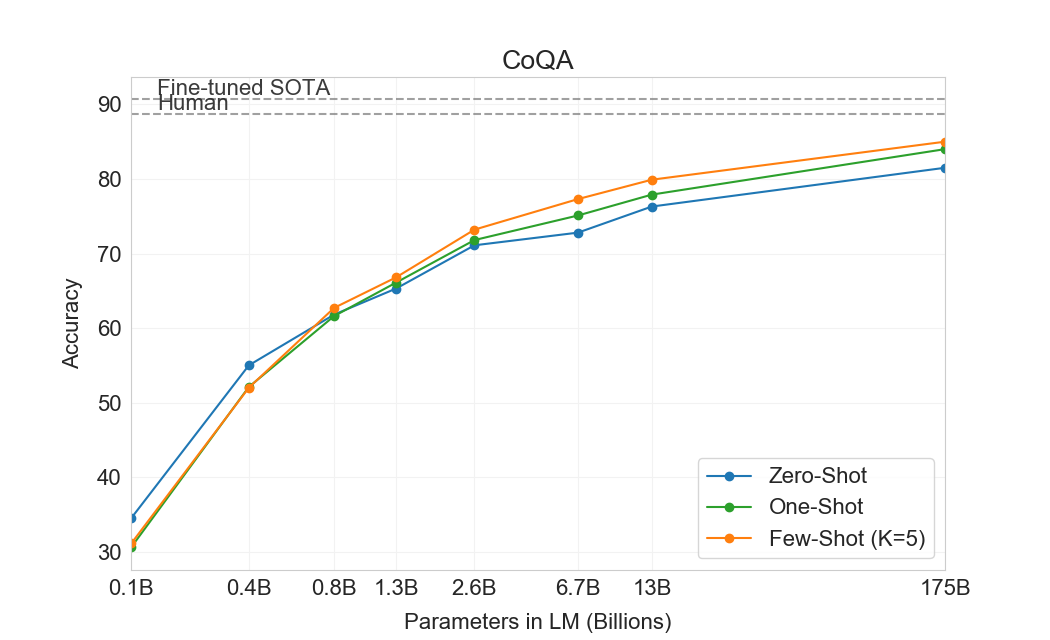}
\caption{GPT-3 results on CoQA reading comprehension task.  GPT-3 175B achieves 85 F1 in the few-shot setting, only a few points behind measured human performance and state-of-the-art fine-tuned models.  Zero-shot and one-shot performance is a few points behind, with the gains to few-shot being largest for bigger models.}
\label{graph:coqa}
\end{figure}
Next we evaluate GPT-3 on the task of reading comprehension. We use a suite of 5 datasets including abstractive, multiple choice, and span based answer formats in both dialog and single question settings. We observe a wide spread in GPT-3's performance across these datasets suggestive of varying capability with different answer formats. In general we observe GPT-3 is on par with initial baselines and early results trained using contextual representations on each respective dataset.

GPT-3 performs best (within 3 points of the human baseline) on CoQA \cite{reddy2019coqa} a free-form conversational dataset and performs worst (13 F1 below an ELMo baseline) on QuAC \cite{choi2018quac} a dataset which requires modeling structured dialog acts and answer span selections of teacher-student interactions. On DROP \cite{dua2019drop}, a dataset testing discrete reasoning and numeracy in the context of reading comprehension, GPT-3 in a few-shot setting outperforms the fine-tuned BERT baseline from the original paper but is still well below both human performance and state-of-the-art approaches which augment neural networks with symbolic systems \cite{ran2019numnet}. On SQuAD 2.0 \cite{rajpurkar2018know}, GPT-3 demonstrates its few-shot learning capabilities, improving by almost 10 F1 (to 69.8) compared to a zero-shot setting. This allows it to slightly outperform the best fine-tuned result in the original paper. On RACE \cite{lai2017race}, a multiple choice dataset of middle school and high school english examinations, GPT-3 performs relatively weakly and is only competitive with the earliest work utilizing contextual representations and is still 45\% behind SOTA.

\begin{table}
    
\begin{center}
\begin{tabular}{lcccccc}
\toprule
                 &  SuperGLUE &     BoolQ &        CB &    CB &      COPA &       RTE \\
                 &    Average &  Accuracy &  Accuracy &    F1 &  Accuracy &  Accuracy \\
\midrule
Fine-tuned SOTA & \textbf{89.0} & \textbf{91.0} & \textbf{96.9} & \textbf{93.9} & \textbf{94.8} & \textbf{92.5} \\        
 Fine-tuned BERT-Large &       69.0 &      77.4 &      83.6 &  75.7 &      70.6 &      71.7 \\
GPT-3 Few-Shot &       71.8 &      76.4 &      75.6 &  52.0 &      92.0 &      69.0 \\
\end{tabular}
\end{center}

\begin{center}
\begin{tabular}{lcccccc}
\toprule
                 &       WiC &       WSC &   MultiRC &  MultiRC &    ReCoRD &  ReCoRD \\
                 &  Accuracy &  Accuracy &  Accuracy &      F1a &  Accuracy &      F1 \\
\midrule
       Fine-tuned SOTA &   \textbf{76.1} &   \textbf{93.8} &   \textbf{62.3} &  \textbf{88.2} &   \textbf{92.5} & \textbf{93.3} \\
 Fine-tuned BERT-Large &      69.6 &      64.6 &      24.1 &     70.0 &      71.3 &    72.0 \\
        GPT-3 Few-Shot &      49.4 &      80.1 &      30.5 &     75.4 &      90.2 &    91.1 \\
\bottomrule
\end{tabular}
\end{center}

    \caption{
    Performance of GPT-3 on SuperGLUE compared to fine-tuned baselines and SOTA. All results are reported on the test set. GPT-3 few-shot is given a total of 32 examples within the context of each task and performs no gradient updates.
    }
    \label{table:superglue}
\end{table}

    \begin{figure}
\centering\includegraphics[width=1.0\linewidth]{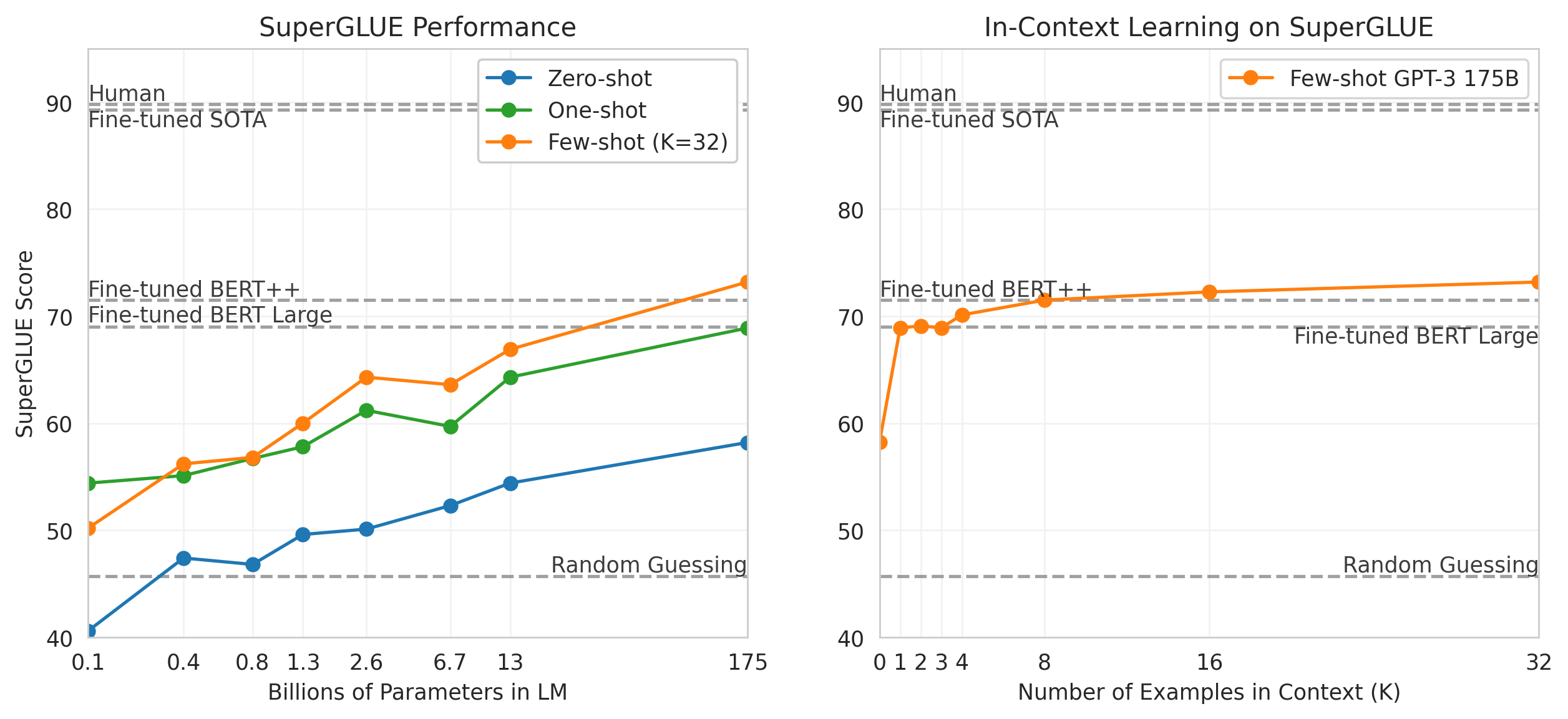}
\caption{
\textbf{
Performance on SuperGLUE increases with model size and number of examples in context.} A value of $K=32$ means that our model was shown 32 examples per task, for 256 examples total divided across the 8 tasks in SuperGLUE. We report GPT-3 values on the dev set, so our numbers are not directly comparable to the dotted reference lines (our test set results are in Table \ref{table:superglue}). The BERT-Large reference model was fine-tuned on the SuperGLUE training set (125K examples), whereas BERT++ was first fine-tuned on MultiNLI (392K examples) and SWAG (113K examples) before further fine-tuning on the SuperGLUE training set (for a total of 630K fine-tuning examples). We find the difference in performance between the BERT-Large and BERT++ to be roughly equivalent to the difference between GPT-3 with one example per context versus eight examples per context.
}
\label{graph:superglue_analysis}
\end{figure}

    \subsection{SuperGLUE}
    \label{section:SuperGLUE}
    In order to better aggregate results on NLP tasks and compare to popular models such as BERT and RoBERTa in a more systematic way, we also evaluate GPT-3 on a standardized collection of datasets, the SuperGLUE benchmark \cite{wang2019superglue} \citep{wang2019superglue} \citep{clark2019boolq} \citep{demarneffe:cb} \citep{roemmele2011choice} \citep{khashabi2018looking} \citep{zhang2018record} \citep{dagan2006pascal} \citep{bar2006second} \citep{giampiccolo2007third} \citep{bentivogli2009fifth} \citep{pilehvar2018wic} \citep{poliak2018dnc}. GPT-3’s test-set performance on the SuperGLUE dataset is shown in Table \ref{table:superglue}.  In the few-shot setting, we used 32 examples for all tasks, sampled randomly from the training set. For all tasks except WSC and MultiRC, we sampled a new set of examples to use in the context for each problem. For WSC and MultiRC, we used the same set of randomly drawn examples from the training set as context for all of the problems we evaluated.

We observe a wide range in GPT-3’s performance across tasks.  On COPA and ReCoRD GPT-3 achieves near-SOTA performance in the one-shot and few-shot settings, with COPA falling only a couple points short and achieving second place on the leaderboard, where first place is held by a fine-tuned 11 billion parameter model (T5). On WSC, performance is still relatively strong, achieving 80.1\% in the few-shot setting (note that GPT-3 achieves 88.6\% on the original Winograd dataset as described in Section \ref{section:Winograd-Style_Tasks}).  On BoolQ, MultiRC, and RTE, performance is reasonable, roughly matching that of a fine-tuned BERT-Large. On CB, we see signs of life at 75.6\% in the few-shot setting.

WiC is a notable weak spot with few-shot performance at 49.4\% (at random chance).  We tried a number of different phrasings and formulations for WiC (which involves determining if a word is being used with the same meaning in two sentences), none of which was able to achieve strong performance.  This hints at a phenomenon that will become clearer in the next section (which discusses the ANLI benchmark) -- GPT-3 appears to be weak in the few-shot or one-shot setting at some tasks that involve comparing two sentences or snippets, for example whether a word is used the same way in two sentences (WiC), whether one sentence is a paraphrase of another, or whether one sentence implies another. This could also explain the comparatively low scores for RTE and CB, which also follow this format. Despite these weaknesses, GPT-3 still outperforms a fine-tuned BERT-large on four of eight tasks and on two tasks GPT-3 is close to the state-of-the-art held by a fine-tuned 11 billion parameter model.

Finally, we note that the few-shot SuperGLUE score steadily improves with both model size and with number of examples in the context showing increasing benefits from in-context learning (Figure \ref{graph:superglue_analysis}). We scale $K$ up to 32 examples per task, after which point additional examples will not reliably fit into our context. When sweeping over values of $K$, we find that GPT-3 requires less than eight total examples per task to outperform a fine-tuned BERT-Large on overall SuperGLUE score.

    \begin{figure}
\centering\includegraphics[width=0.8\linewidth]{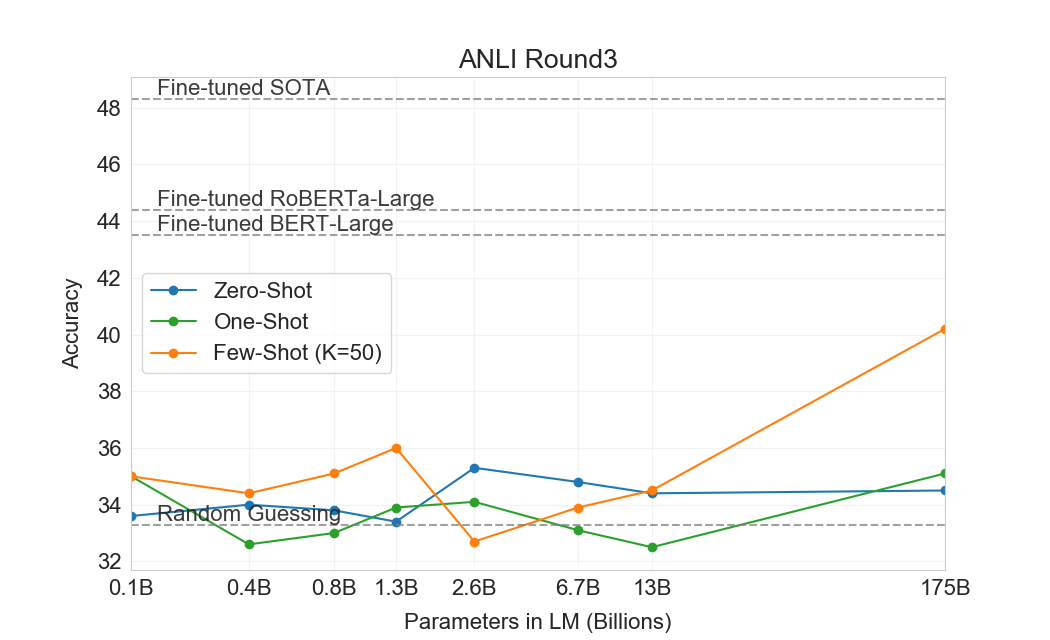}
\caption{\textbf{Performance of GPT-3 on ANLI Round 3.} Results are on the dev-set, which has only 1500 examples and therefore has high variance (we estimate a standard deviation of 1.2\%). We find that smaller models hover around random chance, while few-shot GPT-3 175B closes almost half the gap from random chance to SOTA. Results for ANLI rounds 1 and 2 are shown in the appendix.}
\label{graph:anli}
\end{figure}
    \subsection{NLI}
    \label{section:ANLI}
    Natural Language Inference (NLI) \cite{fyodorov2000natural} concerns the ability to understand the relationship between two sentences. In practice, this task is usually structured as a two or three class classification problem where the model classifies whether the second sentence logically follows from the first, contradicts the first sentence, or is possibly true (neutral). SuperGLUE includes an NLI dataset, RTE, which evaluates the binary version of the task. On RTE, only the largest version of GPT-3 performs convincingly better than random (56\%) in any evaluation setting, but in a few-shot setting GPT-3 performs similarly to a single-task fine-tuned BERT Large. We also evaluate on the recently introduced Adversarial Natural Language Inference (ANLI) dataset \cite{nie2019adversarial}. ANLI is a difficult dataset employing a series of adversarially mined natural language inference questions in three rounds (R1, R2, and R3). Similar to RTE, all of our models smaller than GPT-3 perform at almost exactly random chance on ANLI, even in the few-shot setting ($\sim33\%$), whereas GPT-3 itself shows signs of life on Round 3. Results for ANLI R3 are highlighted in Figure \ref{graph:anli} and full results for all rounds can be found in Appendix \ref{appendix:results_on_all_tasks}. These results on both RTE and ANLI suggest that NLI is still a very difficult task for language models and they are only just beginning to show signs of progress.

    \subsection{Synthetic and Qualitative Tasks}
    \label{section:Synthetic_and_Qualitative_Tasks}
    
One way to probe GPT-3’s range of abilities in the few-shot (or zero- and one-shot) setting is to give it tasks which require it to perform simple on-the-fly computational reasoning, recognize a novel pattern that is unlikely to have occurred in training, or adapt quickly to an unusual task.  We devise several tasks to test this class of abilities.  First, we test GPT-3’s ability to perform arithmetic.  Second, we create several tasks that involve rearranging or unscrambling the letters in a word, tasks which are unlikely to have been exactly seen during training.  Third, we test GPT-3’s ability to solve SAT-style analogy problems few-shot.  Finally, we test GPT-3 on several qualitative tasks, including using new words in a sentence, correcting English grammar, and news article generation.  We will release the synthetic datasets with the hope of stimulating further study of test-time behavior of language models.
    
        \subsubsection{Arithmetic}
        \label{section:Arithmetic}
        
To test GPT-3's ability to perform simple arithmetic operations without task-specific training, we developed a small battery of 10 tests that involve asking GPT-3 a simple arithmetic problem in natural language:

\begin{itemize}
    \item \textbf{2 digit addition (2D+)} -- The model is asked to add two integers sampled uniformly from $[0, 100)$, phrased in the form of a question, e.g. ``Q: What is 48 plus 76? A: 124.''
    \item \textbf{2 digit subtraction (2D-)} -- The model is asked to subtract two integers sampled uniformly from $[0, 100)$; the answer may be negative.  Example: ``Q: What is 34 minus 53? A: -19''.
    \item \textbf{3 digit addition (3D+)} -- Same as 2 digit addition, except numbers are uniformly sampled from $[0, 1000)$.
    \item \textbf{3 digit subtraction (3D-)} -- Same as 2 digit subtraction, except numbers are uniformly sampled from $[0, 1000)$.
    \item \textbf{4 digit addition (4D+)} -- Same as 3 digit addition, except uniformly sampled from $[0, 10000)$.
    \item \textbf{4 digit subtraction (4D-)} -- Same as 3 digit subtraction, except uniformly sampled from $[0, 10000)$.
    \item \textbf{5 digit addition (5D+)} -- Same as 3 digit addition, except uniformly sampled from $[0, 100000)$.
    \item \textbf{5 digit subtraction (5D-)} -- Same as 3 digit subtraction, except uniformly sampled from $[0, 100000)$.
   \item \textbf{2 digit multiplication (2Dx)} -- The model is asked to multiply two integers sampled uniformly from $[0, 100)$, e.g. ``Q: What is 24 times 42? A: 1008''.
    \item \textbf{One-digit composite (1DC)} -- The model is asked to perform a composite operation on three 1 digit numbers, with parentheses around the last two. For example, ``Q: What is 6+(4*8)? A: 38''. The three 1 digit numbers are selected uniformly on $[0, 10)$ and the operations are selected uniformly from \{+,-,*\}.
\end{itemize}

In all 10 tasks the model must generate the correct answer exactly. For each task we generate a dataset of 2,000 random instances of the task and evaluate all models on those instances.

\begin{figure}
\centering\includegraphics[width=0.8\linewidth]{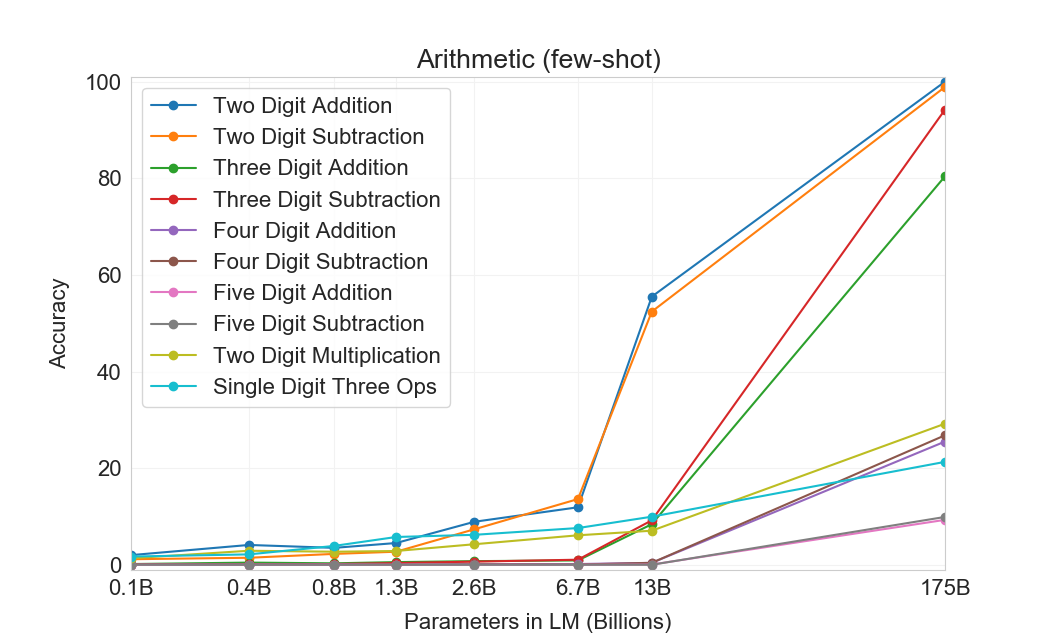}
\caption{Results on all 10 arithmetic tasks in the few-shot settings for models of different sizes.  There is a significant jump from the second largest model (GPT-3 13B) to the largest model (GPT-3 175), with the latter being able to reliably accurate 2 digit arithmetic, usually accurate 3 digit arithmetic, and correct answers a significant fraction of the time on 4-5 digit arithmetic, 2 digit multiplication, and compound operations.  Results for one-shot and zero-shot are shown in the appendix.}
\label{graph:arithmetic}
\end{figure}

First we evaluate GPT-3 in the few-shot setting, for which results are shown in Figure \ref{graph:arithmetic}.  On addition and subtraction, GPT-3 displays strong proficiency when the number of digits is small, achieving 100\% accuracy on 2 digit addition, 98.9\% at 2 digit subtraction, 80.2\% at 3 digit addition, and 94.2\% at 3-digit subtraction.  Performance decreases as the number of digits increases, but GPT-3 still achieves 25-26\% accuracy on four digit operations and 9-10\% accuracy on five digit operations, suggesting at least some capacity to generalize to larger numbers of digits.  GPT-3 also achieves 29.2\% accuracy at 2 digit multiplication, an especially computationally intensive operation. Finally, GPT-3 achieves 21.3\% accuracy at single digit combined operations (for example, 9*(7+5)), suggesting that it has some robustness beyond just single operations.

\begin{table}
    \centering
        \begin{tabular}{l l l l l l l l l l l}
        \toprule
        Setting & 2D+ & 2D- & 3D+ & 3D- & 4D+ & 4D- & 5D+ & 5D- & 2Dx & 1DC \\ 
        \midrule
        GPT-3 Zero-shot & 76.9 & 58.0 & 34.2 & 48.3 & 4.0 & 7.5 & 0.7 & 0.8 & 19.8 & 9.8 \\ 
        GPT-3 One-shot & 99.6 & 86.4 & 65.5 & 78.7 & 14.0 & 14.0 & 3.5 & 3.8 & 27.4 & 14.3 \\ 
        GPT-3 Few-shot & 100.0 & 98.9 & 80.4 & 94.2 & 25.5 & 26.8 & 9.3 & 9.9 & 29.2 & 21.3 \\ 
        \bottomrule
        \end{tabular}
    \caption{Results on basic arithmetic tasks for GPT-3 175B. \{2,3,4,5\}D\{+,-\} is 2, 3, 4, and 5 digit addition or subtraction, 2Dx is 2 digit multiplication.  1DC is 1 digit composite operations.  Results become progressively stronger moving from the zero-shot to one-shot to few-shot setting, but even the zero-shot shows significant arithmetic abilities.}
    \label{table:arithmetic}
\end{table}

As Figure \ref{graph:arithmetic} makes clear, small models do poorly on all of these tasks -- even the 13 billion parameter model (the second largest after the 175 billion full GPT-3) can solve 2 digit addition and subtraction only half the time, and all other operations less than 10\% of the time.

One-shot and zero-shot performance are somewhat degraded relative to few-shot performance, suggesting that adaptation to the task (or at the very least recognition of the task) is important to performing these computations correctly.  Nevertheless, one-shot performance is still quite strong, and even zero-shot performance of the full GPT-3 significantly outperforms few-shot learning for all smaller models.  All three settings for the full GPT-3 are shown in Table \ref{table:arithmetic}, and model capacity scaling for all three settings is shown in Appendix \ref{appendix:results_on_all_tasks}.

To spot-check whether the model is simply memorizing specific arithmetic problems, we took the 3-digit arithmetic problems in our test set and searched for them in our training data in both the forms \texttt{"<NUM1> + <NUM2> ="} and \texttt{"<NUM1> plus <NUM2>"}.  Out of 2,000 addition problems we found only 17 matches (0.8\%) and out of 2,000 subtraction problems we found only 2 matches (0.1\%), suggesting that only a trivial fraction of the correct answers could have been memorized.  In addition, inspection of incorrect answers reveals that the model often makes mistakes such as not carrying a ``1'', suggesting it is actually attempting to perform the relevant computation rather than memorizing a table.

Overall, GPT-3 displays reasonable proficiency at moderately complex arithmetic in few-shot, one-shot, and even zero-shot settings.

        \subsubsection{Word Scrambling and Manipulation Tasks}
        \label{section:Word_Scrambling_and_Manipulation_Tasks}
        
To test GPT-3's ability to learn novel symbolic manipulations from a few examples, we designed a small battery of 5 ``character manipulation'' tasks.  Each task involves giving the model a word distorted by some combination of scrambling, addition, or deletion of characters, and asking it to recover the original word.  The 5 tasks are:

\begin{table}
    \centering
        \begin{tabular}{l l l l l l}
        \toprule
        Setting & CL & A1 & A2 & RI & RW \\ 
        \midrule
        GPT-3 Zero-shot & 3.66 & 2.28 & 8.91 & 8.26 & 0.09 \\
        GPT-3 One-shot & 21.7 & 8.62 & 25.9 & 45.4 & 0.48 \\ 
        GPT-3 Few-shot & 37.9 & 15.1 & 39.7 & 67.2 & 0.44 \\ 
        \bottomrule
        \end{tabular}
    \caption{GPT-3 175B performance on various word unscrambling and word
manipulation tasks, in zero-, one-, and few-shot settings. CL is ``cycle letters in word'', A1 is anagrams of but the first and last letters,
A2 is anagrams of all but the first and last two letters, RI is ``Random insertion
in word'', RW is ``reversed words''.}
    \label{table:unscramble}
\end{table}
\begin{figure}
\centering\includegraphics[width=0.8\linewidth]{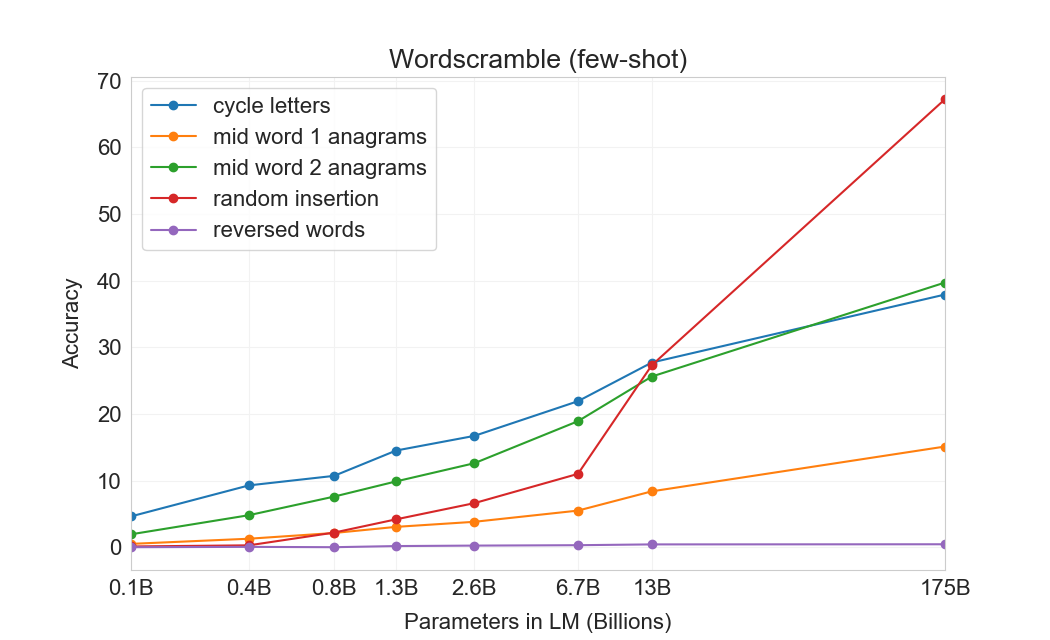}
\caption{Few-shot performance on the five word scrambling tasks for different sizes of model.  There is generally smooth improvement with model size although the random insertion task shows an upward slope of improvement with the 175B model solving the task the majority of the time.  Scaling of one-shot and zero-shot performance is shown in the appendix.  All tasks are done with $K=100$.}
\label{graph:scramble}
\end{figure}

\begin{itemize}
    \item \textbf{Cycle letters in word (CL)} -- The model is given a word with its letters cycled, then the ``='' symbol, and is expected to generate the original word.  For example, it might be given ``lyinevitab'' and should output ``inevitably''.
    \item \textbf{Anagrams of all but first and last characters (A1)} -- The model is given a word where every letter except the first and last have been scrambled randomly, and must output the original word.  Example: criroptuon = corruption.
    \item \textbf{Anagrams of all but first and last 2 characters (A2)} -- The model is given a word where every letter except the first 2 and last 2 have been scrambled randomly, and must recover the original word.  Example: opoepnnt $\to$ opponent.
    \item \textbf{Random insertion in word (RI)} -- A random punctuation or space character is inserted between each letter of a word, and the model must output the original word.  Example: s.u!c/c!e.s s i/o/n = succession.
    \item \textbf{Reversed words (RW)} -- The model is given a word spelled backwards, and must output the original word.  Example: stcejbo $\to$ objects.
\end{itemize}

For each task we generate 10,000 examples, which we chose to be the top 10,000 most frequent words as measured by \cite{norvig2009ngram} of length more than 4 characters and less than 15 characters. The few-shot results are shown in Figure \ref{graph:scramble}.  Task performance tends to grow smoothly with model size, with the full GPT-3 model achieving 66.9\% on removing random insertions, 38.6\% on cycling letters, 40.2\% on the easier anagram task, and 15.1\% on the more difficult anagram task (where only the first and last letters are held fixed).  None of the models can reverse the letters in a word. 

In the one-shot setting, performance is significantly weaker (dropping by half or more), and in the zero-shot setting the model can rarely perform any of the tasks (Table \ref{table:unscramble}).  This suggests that the model really does appear to learn these tasks at test time, as the model cannot perform them zero-shot and their artificial nature makes them unlikely to appear in the pre-training data (although we cannot confirm this with certainty).

We can further quantify performance by plotting ``in-context learning curves'', which show task performance as a function of the number of in-context examples.  We show in-context learning curves for the Symbol Insertion task in Figure \ref{graph:scramble_prompted}.  We can see that larger models are able to make increasingly effective use of in-context information, including both task examples and natural language task descriptions.

Finally, it is worth adding that solving these tasks requires character-level manipulations, whereas our BPE encoding operates on significant fractions of a word (on average $\sim0.7$ words per token), so from the LM’s perspective succeeding at these tasks involves not just manipulating BPE tokens but understanding and pulling apart their substructure. Also, CL, A1, and A2 are not bijective (that is, the unscrambled word is not a deterministic function of the scrambled word), requiring the model to perform some search to find the correct unscrambling. Thus, the skills involved appear to require non-trivial pattern-matching and computation.

        \subsubsection{SAT Analogies}
        \label{section:SAT_Analogies}
\begin{figure}
\centering\includegraphics[width=0.8\linewidth]{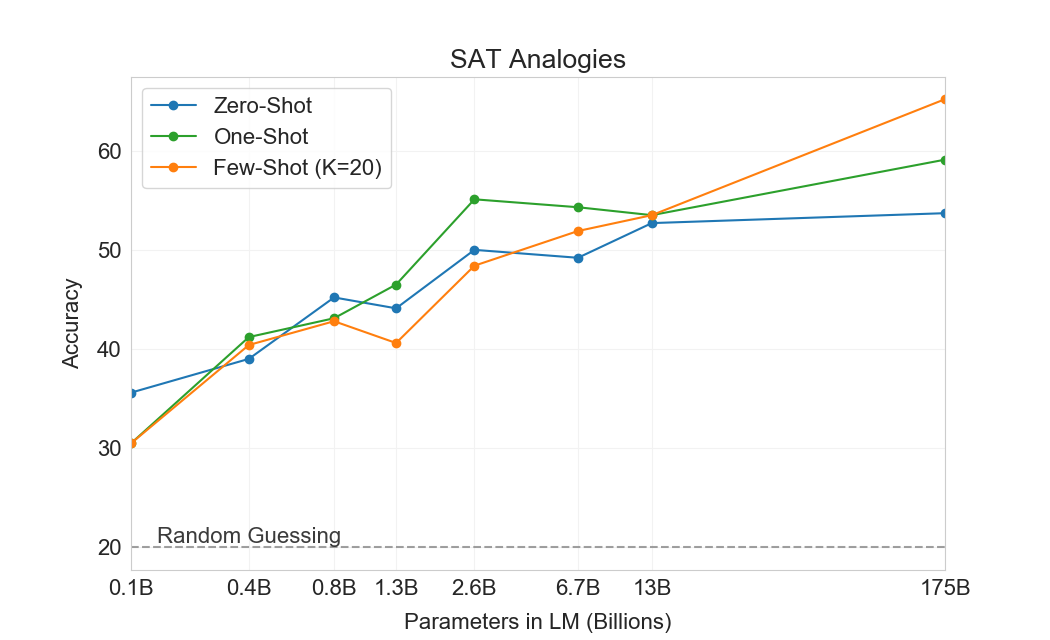}
\caption{Zero-, one-,and few-shot performance on SAT analogy tasks, for different sizes of model.  The largest model achieves 65\% accuracy in the few-shot setting, and also demonstrates significant gains to in-context learning which are not present in smaller models.}
\label{graph:sat}
\end{figure}

        To test GPT-3 on another task that is somewhat unusual relative to the typical distribution of text, we collected a set of 374 ``SAT analogy'' problems \cite{DBLP:journals/corr/cs-CL-0309035}.  Analogies are a style of multiple choice question that constituted a section of the SAT college entrance exam before 2005.  A typical example is ``audacious is to boldness as (a) sanctimonious is to hypocrisy, (b) anonymous is to identity, (c) remorseful is to misdeed, (d) deleterious is to result, (e) impressionable is to temptation''.  The student is expected to choose which of the five word pairs has the same relationship as the original word pair; in this example the answer is ``sanctimonious is to hypocrisy''.  On this task GPT-3 achieves 65.2\% in the few-shot setting, 59.1\% in the one-shot setting, and 53.7\% in the zero-shot setting, whereas the average score among college applicants was 57\% \cite{DBLP:journals/corr/abs-cs-0508103}  (random guessing yields 20\%).  As shown in Figure \ref{graph:sat}, the results improve with scale, with the the full 175 billion model improving by over 10\% compared to the 13 billion parameter model.

        \subsubsection{News Article Generation}
        \label{section:News_Article_Generation}
        Previous work on generative language models qualitatively tested their ability to generate synthetic ``news articles'' by conditional sampling from the model given a human-written prompt consisting of a plausible first sentence for a news story \cite{radford2019language}.  Relative to \cite{radford2019language}, the dataset used to train GPT-3 is much less weighted towards news articles, so trying to generate news articles via raw unconditional samples is less effective -- for example GPT-3 often interprets the proposed first sentence of a ``news article'' as a tweet and then posts synthetic responses or follow-up tweets.  To solve this problem we employed GPT-3’s few-shot learning abilities by providing three previous news articles in the model’s context to condition it.  With the title and subtitle of a proposed next article, the model is able to reliably generate short articles in the ``news'' genre.

To gauge the quality of news article generation from GPT-3 (which we believe is likely to be correlated with conditional sample generation quality in general), we decided to measure human ability to distinguish GPT-3-generated articles from real ones. Similar work has been carried out by Kreps et al. \cite{kreps-all-the-news} and Zellers et al. \cite{zellers2019defending}. Generative language models are trained to match the distribution of content generated by humans, so the (in)ability of humans to distinguish the two is a potentially important measure of quality.\footnote{This task is also relevant to the potential misuse of language models discussed in Section \ref{section:misuse_of_language_models}.} 

In order to see how well humans can detect model generated text, we arbitrarily selected 25 article titles and subtitles from the website \href{newser.com}{newser.com} (mean length: 215 words). We then generated completions of these titles and subtitles from four language models ranging in size from 125M to 175B (GPT-3) parameters (mean length: 200 words). For each model, we presented around 80 US-based participants with a quiz consisting of these real titles and subtitles followed by either the human written article or the article generated by the model\footnote{We wanted to identify how good an average person on the internet is at detecting language model outputs, so we focused on participants drawn from the general US population. See Appendix \ref{appendix:human_assessment} for details.}. Participants were asked to select whether the article was ``very likely written by a human'', ``more likely written by a human'', ``I don't know'', ``more likely written by a machine'', or ``very likely written by a machine''.

The articles we selected were not in the models’ training data and the model outputs were formatted and selected programmatically to prevent human cherry-picking. All models used the same context to condition outputs on and were pre-trained with the same context size and the same article titles and subtitles were used as prompts for each model. However, we also ran an experiment to control for participant effort and attention that followed the same format but involved intentionally bad model generated articles. This was done by generating articles from a ``control model'': a 160M parameter model with no context and increased output randomness.

Mean human accuracy (the ratio of correct assignments to non-neutral assignments per participant) at detecting that the intentionally bad articles were model generated was  $\sim86\%$\, where 50\% is chance level performance. By contrast, mean human accuracy at detecting articles that were produced by the 175B parameter model was barely above chance at $\sim52\%$ (see Table \ref{table:generation}).\footnote{We use a two-sample Student’s T-Test to test for significant difference between the means of the participant accuracies of each model and the control model and report the normalized difference in the means (as the t-statistic) and the p-value.} Human abilities to detect model generated text appear to decrease as model size increases: there appears to be a trend towards chance accuracy with model size, and human detection of GPT-3 is close to chance.\footnote{If a model consistently produces texts that are more impressive than human articles, it is possible that human performance on this task would drop below 50\%. Indeed, many individual participants scored below 50\% on this task.} This is true despite the fact that participants spend more time on each output as model size increases (see Appendix \ref{appendix:human_assessment}).

\begin{table}
    \centering
        \begin{tabular}{l c c c c}
        \toprule
         & Mean accuracy & \shortstack{95\% Confidence \\Interval (low, hi)} & \shortstack{$t$ compared to \\ control ($p$-value)} & \shortstack{``I don't know" \\ assignments} \\ 
        \midrule
Control (deliberately bad model) & 86\% & 83\%–90\% & - & 3.6 \% \\ 
GPT-3 Small  & 76\% & 72\%–80\% & 3.9 (2$e$-4) & 4.9\% \\ 
GPT-3 Medium & 61\% & 58\%–65\% & 10.3 (7$e$-21) & 6.0\% \\ 
GPT-3 Large  & 68\% & 64\%–72\% & 7.3 (3$e$-11) & 8.7\% \\ 
GPT-3 XL     & 62\% & 59\%–65\% & 10.7 (1$e$-19) & 7.5\% \\ 
GPT-3 2.7B   & 62\% & 58\%–65\% & 10.4 (5$e$-19) & 7.1\% \\ 
GPT-3 6.7B   & 60\% & 56\%–63\% &  11.2 (3$e$-21) & 6.2\% \\ 
GPT-3 13B    & 55\% & 52\%–58\% &  15.3 (1$e$-32) & 7.1\% \\ 
GPT-3 175B   & 52\% & 49\%–54\% & 16.9 (1$e$-34) & 7.8\% \\ 
        \bottomrule
        \end{tabular}
    \caption{\textbf{Human accuracy in identifying whether short ($\sim$200 word) news articles are model generated}. We find that human accuracy (measured by the ratio of correct assignments to non-neutral assignments) ranges from 86\% on the control model to 52\% on GPT-3 175B. This table compares mean accuracy between five different models, and shows the results of a two-sample T-Test for the difference in mean accuracy between each model and the control model (an unconditional GPT-3 Small model with increased output randomness).}
    \label{table:generation}
\end{table}
\begin{figure}
\centering
\includegraphics[width=\linewidth]{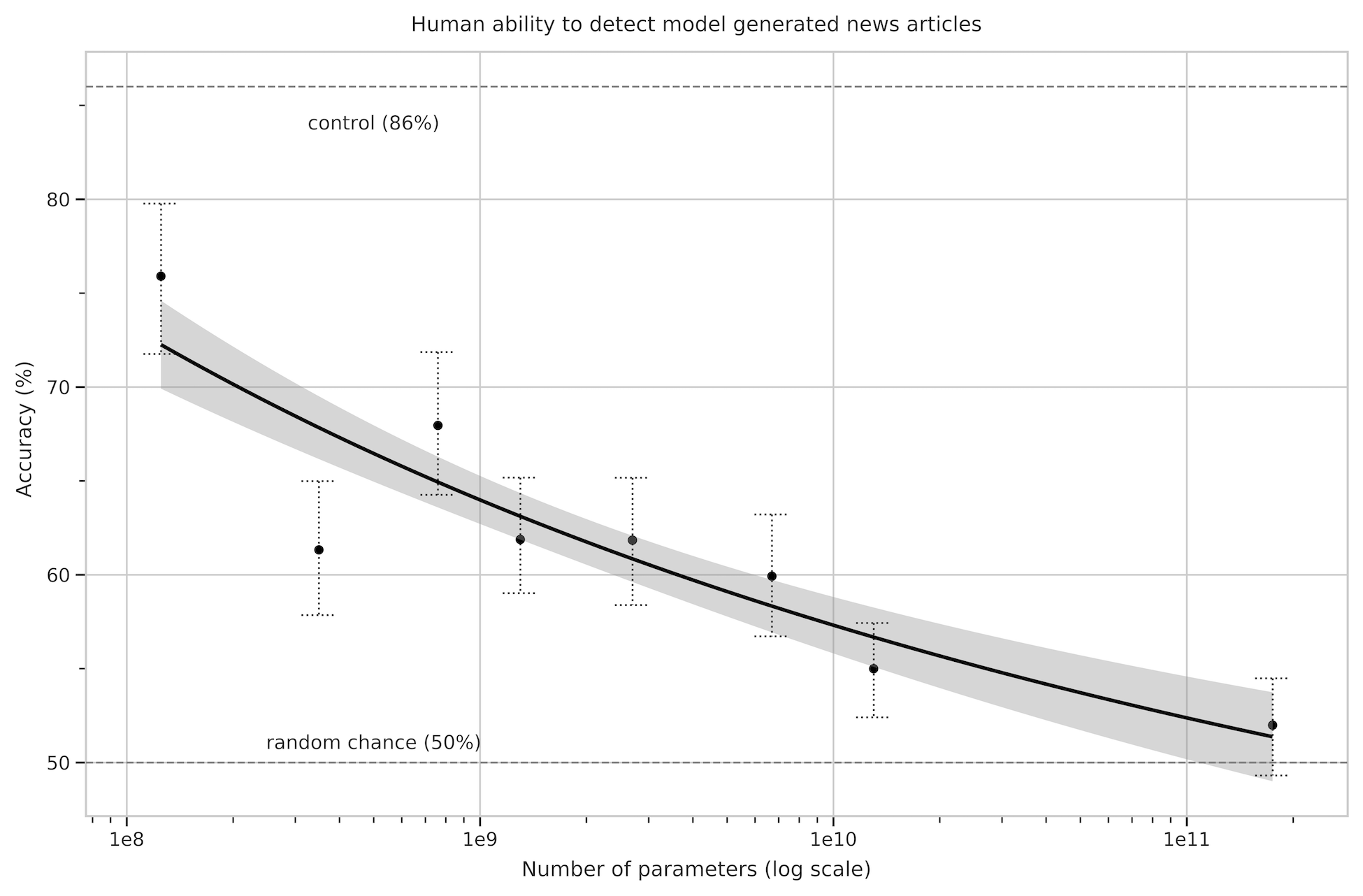}
\caption{People's ability to identify whether news articles are model-generated (measured by the ratio of correct assignments to non-neutral assignments) decreases as model size increases. Accuracy on the outputs on the deliberately-bad control model (an unconditioned GPT-3 Small model with higher output randomness) is indicated with the dashed line at the top, and the random chance (50\%) is indicated with the dashed line at the bottom. Line of best fit is a power law with 95\% confidence intervals.}
\label{graph:generation}
\end{figure}

 Examples of synthetic articles from GPT-3 are given in Figures \ref{completion:newsgen_best} and \ref{completion:newsgen_worst}.\footnote{Additional non-news samples can be found in Appendix \ref{appendix:additional_samples}.} Much of the text is---as indicated by the evaluations---difficult for humans to distinguish from authentic human content. Factual inaccuracies can be an indicator that an article is model generated since, unlike human authors, the models have no access to the specific facts that the article titles refer to or when the article was written. Other indicators include repetition, non sequiturs, and unusual phrasings, though these are often subtle enough that they are not noticed. 
 
\begin{figure}
\centering\noindent\fbox{%
   \tt \parbox{0.9\linewidth}{%
{\color{gray}Title: United Methodists Agree to Historic Split \\
Subtitle: Those who oppose gay marriage will form their own denomination \\
Article:}
\textbf{After two days of intense debate, the United Methodist Church has agreed to a historic split - one that is expected to end in the creation of a new denomination, one that will be "theologically and socially conservative," according to The Washington Post. The majority of delegates attending the church’s annual General Conference in May voted to strengthen a ban on the ordination of LGBTQ clergy and to write new rules that will "discipline" clergy who officiate at same-sex weddings. But those who opposed these measures have a new plan: They say they will form a separate denomination by 2020, calling their church the Christian Methodist denomination.}

\textbf{The Post notes that the denomination, which claims 12.5 million members, was in the early 20th century the "largest Protestant denomination in the U.S.," but that it has been shrinking in recent decades. The new split will be the second in the church’s history. The first occurred in 1968, when roughly 10 percent of the denomination left to form the Evangelical United Brethren Church. The Post notes that the proposed split "comes at a critical time for the church, which has been losing members for years," which has been "pushed toward the brink of a schism over the role of LGBTQ people in the church." Gay marriage is not the only issue that has divided the church. In 2016, the denomination was split over ordination of transgender clergy, with the North Pacific regional conference voting to ban them from serving as clergy, and the South Pacific regional conference voting to allow them.
}
    }}
\caption{The GPT-3 generated news article that humans had the greatest difficulty distinguishing from a human written article (accuracy: 12\%).}
\label{completion:newsgen_best}
\end{figure}
\begin{figure}
\centering\noindent\fbox{%
    \parbox{0.9\linewidth}{%
\tt {\color{gray}Title: Star's Tux Promise Draws Megyn Kelly's Sarcasm

Subtitle: Joaquin Phoenix pledged to not change for each awards event

Article:}
\textbf{A year ago, Joaquin Phoenix made headlines when he appeared on the red carpet at the Golden Globes wearing a tuxedo with a paper bag over his head that read, "I am a shape-shifter. I can't change the world. I can only change myself." It was a promise to not change to fit into the Hollywood mold: "I think that's a really special thing, to not change yourself. I think it's a really special thing to say, `This is what's inside of me, I'm proud of it, and I'm not going to be ashamed because of the way that someone else thinks I should be.'" Now, it's the Oscars, and Phoenix is at it again. But this time, his publicist is saying he'll be wearing a tux no matter what.}

\textbf{Megyn Kelly was not impressed, and she let him have it on The Tonight Show. "You know, I feel like, I feel like you could have worn the tux," she says. "But you're saying you're a shape-shifter. I don't know if you can change your tux, but you can change your mind. You can change your mind. You can change your mind." Phoenix says he did, but it didn't stick. "I was like, `Okay, I'm going to wear a tuxedo to this thing.' And then I thought, `I don't want to wear a tuxedo to this thing.'" Kelly goes on to encourage him to change his mind again, but Phoenix says it's too late: "I'm committed to wearing this."
}
    }}
\caption{The GPT-3 generated news article that humans found the easiest to distinguish from a human written article (accuracy: 61\%).}
\label{completion:newsgen_worst}
\end{figure}

 Related work on language model detection by Ippolito et al. \cite{ippolito2019automatic} indicates that automatic discriminators like \textsc{Grover} \cite{zellers2019defending} and GLTR \cite{gehrmann2019gltr} may have greater success at detecting model generated text than human evaluators. Automatic detection of these models may be a promising area of future research.
 
Ippolito et al. \cite{ippolito2019automatic} also note that human accuracy at detecting model generated text increases as humans observe more tokens. To do a preliminary investigation of how good humans are at detecting longer news articles generated by GPT-3 175B, we selected 12 world news articles from Reuters with an average length of 569 words and generated completions of these articles from GPT-3 with an average length of 498 words (298 words longer than our initial experiments). Following the methodology above, we ran two experiments, each on around 80 US-based participants, to compare human abilities to detect the articles generated by GPT-3 and a control model.
 
 We found that mean human accuracy at detecting the intentionally bad longer articles from the control model was $\sim88\%$, while mean human accuracy at detecting the longer articles that were produced by GPT-3 175B was still barely above chance at $\sim52\%$ (see Table \ref{table:generation_long}). This indicates that, for news articles that are around 500 words long, GPT-3 continues to produce articles that humans find difficult to distinguish from human written news articles.
 
 \begin{table}
    \centering
        \begin{tabular}{l c c c c}
        \toprule
         & Mean accuracy & \shortstack{95\% Confidence \\Interval (low, hi)} & \shortstack{$t$ compared to \\ control ($p$-value)} & \shortstack{``I don't know" \\ assignments} \\ 
        \midrule
Control & 88\% & 84\%–91\% & - & 2.7\% \\ 
GPT-3 175B   & 52\% & 48\%–57\% & 12.7 (3.2$e$-23) & 10.6\% \\ 
        \bottomrule
        \end{tabular}
    \caption{People’s ability to identify whether $\sim500$ word articles are model generated (as measured by the ratio of correct assignments to non-neutral assignments) was 88\% on the control model and 52\% on GPT-3 175B. This table shows the results of a two-sample T-Test for the difference in mean accuracy between GPT-3 175B and the control model (an unconditional GPT-3 Small model with increased output randomness).}
    \label{table:generation_long}
\end{table}

        \begin{figure}
\centering\noindent\fbox{%
   \footnotesize \tt \parbox{0.95\linewidth}{%
{\vspace{0.5em}
{\color{gray} A "whatpu" is a small, furry animal native to Tanzania.  An example of a sentence that uses the word whatpu is: \\
We were traveling in Africa and we saw these very cute whatpus.\\ \rule{\linewidth}{0.5pt}\vspace{0.5em}
To do a "farduddle" means to jump up and down really fast.  An example of a sentence that uses the word farduddle is:\\}%
\textbf{One day when I was playing tag with my little sister, she got really excited and she started doing these crazy farduddles.}\\ 
{\color{gray}A "yalubalu" is a type of vegetable that looks like a big pumpkin. An example of a sentence that uses the word yalubalu is:\\}%
\textbf{I was on a trip to Africa and I tried this yalubalu vegetable that was grown in a garden there. It was delicious.}\\\rule{\linewidth}{0.5pt}\vspace{0.5em}
{\color{gray}A "Burringo" is a car with very fast acceleration.  An example of a sentence that uses the word Burringo is:\\}%
\textbf{In our garage we have a Burringo that my father drives to work every day.}\\\rule{\linewidth}{0.5pt}\vspace{0.5em}
{\color{gray}A "Gigamuru" is a type of Japanese musical instrument.  An example of a sentence that uses the word Gigamuru is:\\}%
\textbf{I have a Gigamuru that my uncle gave me as a gift. I love to play it at home.}\\\rule{\linewidth}{0.5pt}\vspace{0.5em}
{\color{gray}To "screeg" something is to swing a sword at it.  An example of a sentence that uses the word screeg is:\\}%
\textbf{We screeghed at each other for several minutes and then we went outside and ate ice cream.}
}}}
\caption{Representative GPT-3 completions for the few-shot task of using a new word in a sentence.  Boldface is GPT-3’s completions, plain text is human prompts.  In the first example both the prompt and the completion are provided by a human; this then serves as conditioning for subsequent examples where GPT-3 receives successive additional prompts and provides the completions.  Nothing task-specific is provided to GPT-3 other than the conditioning shown here.}
\label{completion:novel words}
\end{figure}

        \begin{figure}
\centering\noindent\fbox{%
\parbox{0.95\linewidth}{%
{\vspace{0.5em}
\footnotesize \tt{\color{gray}Poor English input: I eated the purple berries.\\
Good English output: I ate the purple berries.\\
Poor English input: Thank you for picking me as your designer. I'd appreciate it. \\
Good English output: Thank you for choosing me as your designer. I appreciate it.\\
Poor English input: The mentioned changes have done. or I did the alteration that you requested. or I changed things you wanted and did the modifications.\\
Good English output: The requested changes have been made. or I made the alteration that you requested. or I changed things you wanted and made the modifications.\\
Poor English input: I'd be more than happy to work with you in another project.\\}%
\textbf{Good English output: I'd be more than happy to work with you on another project.} \\\rule{\linewidth}{0.8pt}\vspace{0.5em}
{\color{gray}Poor English input: Please provide me with a short brief of the design you're looking for and that'd be nice if you could share some examples or project you did before.\\}%
\textbf{Good English output: Please provide me with a brief description of the design you're looking for and that would be nice if you could share some examples or projects you have done before.}\\ \rule{\linewidth}{0.8pt}\vspace{0.5em}
{\color{gray}Poor English input: The patient was died.\\}%
\textbf{Good English output: The patient died.}\\ \rule{\linewidth}{0.8pt}\vspace{0.5em}
{\color{gray}Poor English input: We think that Leslie likes ourselves.\\}%
\textbf{Good English output: We think that Leslie likes us.}\\ \rule{\linewidth}{0.8pt}\vspace{0.5em}
{\color{gray}Poor English input: Janet broke Bill on the finger.\\}%
\textbf{Good English output: Janet broke Bill's finger.}\\ \rule{\linewidth}{0.8pt}\vspace{0.5em}
{\color{gray}Poor English input: Mary arranged for, in St. Louis, John to rent a house cheap.\\}%
\textbf{Good English output: Mary arranged for John to rent a house in St. Louis.}\\ \rule{\linewidth}{0.8pt}\vspace{0.5em}

{\color{gray}Poor English input: Today I have went to the store to to buys some many bottle of water.\\}%
\textbf{Good English output: Today I went to the store to buy some bottles of water.}\\ \rule{\linewidth}{0.8pt}\vspace{0.5em}
{\color{gray}Poor English input: I have tried to hit ball with bat, but my swing is has miss.\\}%
\textbf{Good English output: I tried to hit the ball with the bat, but my swing missed.}}
}}
\caption{Representative GPT-3 completions for the few-shot task of correcting English grammar.  Boldface is GPT-3’s completions, plain text is human prompts.  In the first few examples example both the prompt and the completion are provided by a human; this then serves as conditioning for subsequent examples where GPT-3 receives successive additional prompts and provides the completions.  Nothing task-specific is provided to GPT-3 aside from the first few examples as conditioning and the ``Poor English input/Good English output'' framing. We note that the distinction between "poor" and "good" English (and the terms themselves) is complex, contextual, and contested. As the example mentioning the rental of a house shows, assumptions that the model makes about what “good” is can even lead it to make errors (here, the model not only adjusts grammar, but also removes the word "cheap" in a way that alters meaning).}
\label{completion:grammar}

\end{figure}
        \subsubsection{Learning and Using Novel Words}
        \label{section:Learning_and_Using_Novel_Words}
        A task studied in developmental linguistics \cite{carey1978acquiring} is the ability to learn and utilize new words, for example using a word in a sentence after seeing it defined only once, or conversely inferring a word’s meaning from only one usage.  Here we qualitatively test GPT-3’s ability to do the former.  Specifically, we give GPT-3 the definition of a nonexistent word, such as ``Gigamuru'', and then ask it to use it in a sentence.  We provide one to five previous examples of a (separate) nonexistent word being defined and used in a sentence, so the task is few-shot in terms of previous examples of the broad task and one-shot in terms of the specific word.  Table \ref{completion:novel words} shows the 6 examples we generated; all definitions were human-generated, and the first answer was human-generated as conditioning while the subsequent answers were generated by GPT-3.  These examples were generated continuously in one sitting and we did not omit or repeatedly try any prompts.  In all cases the generated sentence appears to be a correct or at least plausible use of the word.  In the final sentence the model generates a plausible conjugation for the word ``screeg'' (namely ``screeghed''), although the use of the word is slightly awkward (``screeghed at each other'') despite being plausible in the sense that it could describe a toy sword fight.  Overall, GPT-3 appears to be at least proficient at the task of using novel words in a sentence.

        \subsubsection{Correcting English Grammar}
        \label{section:Correcting_English_Grammar}
        Another task well suited for few-shot learning is correcting English grammar.  We test this with GPT-3 in the few-shot setting by giving prompts of the form \texttt{"Poor English Input: <sentence>{\textbackslash}n Good English Output: <sentence>"}.  We give GPT-3 one human-generated correction and then ask it to correct 5 more (again without any omissions or repeats).  Results are shown in Figure \ref{completion:grammar}.

%
\section{Measuring and Preventing Memorization Of Benchmarks}
\begin{figure}
\centering\includegraphics[width=0.8\linewidth]{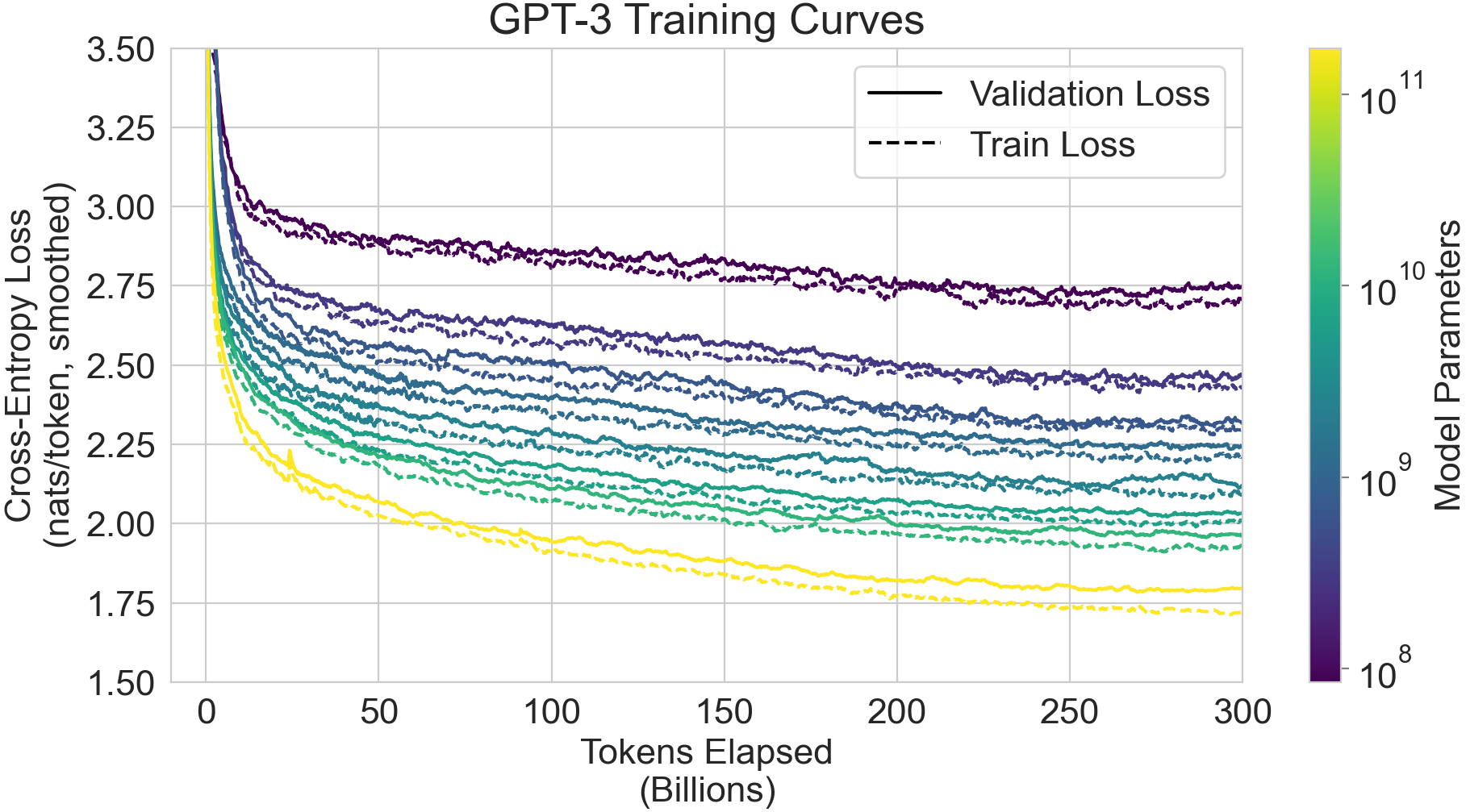}

\caption{\textbf{GPT-3 Training Curves}~~~We measure model performance during training on a deduplicated validation split of our training distribution. Though there is some gap between training and validation performance, the gap grows only minimally with model size and training time, suggesting that most of the gap comes from a difference in difficulty rather than overfitting.}
\label{graph:training_curves}
\end{figure}
\label{section:measuring_and_preventing_memorization_of_benchmarks}
Since our training dataset is sourced from the internet, it is possible that our model was trained on some of our benchmark test sets.  Accurately detecting test contamination from internet-scale datasets is a new area of research without established best practices. While it is common practice to train large models without investigating contamination, given the increasing scale of pretraining datasets, we believe this issue is becoming increasingly important to attend to.

This concern is not just hypothetical. One of the first papers to train a language model on Common Crawl data \cite{trinh2018simple} detected and removed a training document which overlapped with one of their evaluation datasets. Other work such as GPT-2 \cite{radford2019language} also conducted post-hoc overlap analysis. Their study was relatively encouraging, finding that although models did perform moderately better on data that overlapped between training and testing, this did not significantly impact reported results due to the small fraction of data which was contaminated (often only a few percent).

GPT-3 operates in a somewhat different regime.  On the one hand, the dataset and model size are about two orders of magnitude larger than those used for GPT-2, and include a large amount of Common Crawl, creating increased potential for contamination and memorization.  On the other hand, precisely due to the large amount of data, even GPT-3 175B does not overfit its training set by a significant amount, measured relative to a held-out validation set with which it was deduplicated (Figure \ref{graph:training_curves}).  Thus, we expect that contamination is likely to be frequent, but that its effects may not be as large as feared.

We initially tried to address the issue of contamination by proactively searching for and attempting to remove any overlap between our training data and the development and test sets of all benchmarks studied in this paper. Unfortunately, a bug resulted in only partial removal of all detected overlaps from the training data. Due to the cost of training, it wasn't feasible to retrain the model. To address this, we investigate in detail how the remaining detected overlap impacts results.

For each benchmark, we produce a `clean' version which removes all potentially leaked examples, defined roughly as examples that have a 13-gram overlap with anything in the pretraining set (or that overlap with the whole example when it is shorter than 13-grams). The goal is to very conservatively flag anything that could potentially be contamination, so as to produce a clean subset that is free of contamination with high confidence. The exact procedure is detailed in Appendix \ref{appendix:test_set_contamination}.

\begin{figure}
\centering\includegraphics[width=0.9\linewidth]{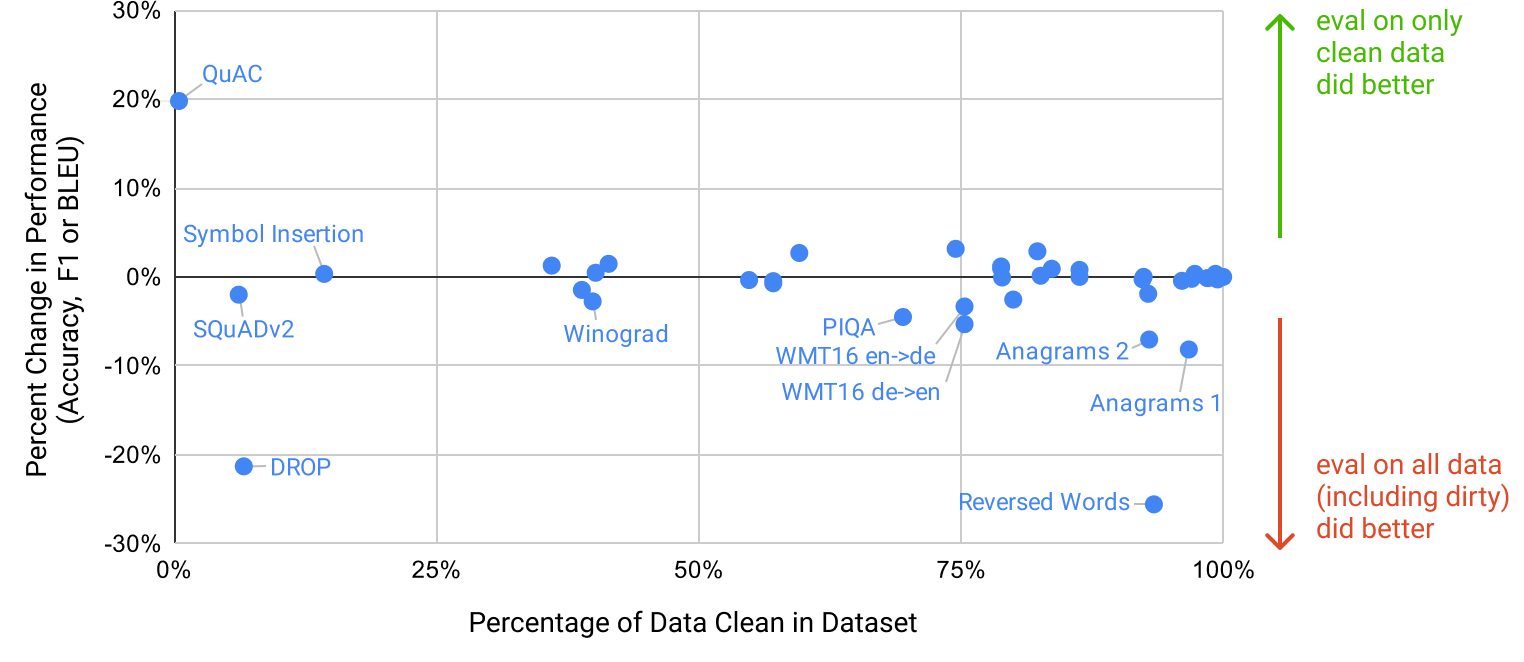}

\caption{\textbf{Benchmark contamination analysis}~~~ We constructed cleaned versions of each of our benchmarks to check for potential contamination in our training set. The x-axis is a conservative lower bound for how much of the dataset is known with high confidence to be clean, and the y-axis shows the difference in performance when evaluating only on the verified clean subset.  Performance on most benchmarks changed negligibly, but some were flagged for further review.  On inspection we find some evidence for contamination of the PIQA and Winograd results, and we mark the corresponding results in Section \ref{section:Results} with an asterisk. We find no evidence that other benchmarks are affected.}
\label{graph:contamination}
\end{figure}

We then evaluate GPT-3 on these clean benchmarks, and compare to the original score.  If the score on the clean subset is similar to the score on the entire dataset, this suggests that contamination, even if present, does not have a significant effect on reported results.  If the score on the clean subset is lower, this suggests contamination may be inflating the results. The results are summarized in Figure \ref{graph:contamination}.  Although potential contamination is often high (with a quarter of benchmarks scoring over 50\%), in most cases performance changes only negligibly, and we see no evidence that contamination level and performance difference are correlated.  We conclude that either our conservative method substantially overestimated contamination or that contamination has little effect on performance.

Below, we review in more detail the few specific cases where either (1) the model performs significantly worse on the cleaned version, or (2) potential contamination is very high, which makes measuring the performance difference difficult.

Our analysis flagged six groups of benchmarks for further investigation: Word Scrambling, Reading Comprehension (QuAC, SQuAD2, DROP), PIQA, Winograd, language modeling tasks (Wikitext tasks, 1BW), and German to English translation. Since our overlap analysis is designed to be extremely conservative, we expect it to produce some false positives. We summarize the results for each group of tasks below:
\begin{itemize}
    \item \textbf{Reading Comprehension:} Our initial analysis flagged $>$90\% of task examples from QuAC, SQuAD2, and DROP as potentially contaminated, so large that even measuring the differential on a clean subset was difficult. Upon manual inspection, however, we found that for every overlap we inspected, in all 3 datasets, the source text was present in our training data but the question/answer pairs were not, meaning the model gains only background information and cannot memorize the answer to a specific question.
    \item \textbf{German translation:} We found 25\% of the examples in the WMT16 German-English test set were marked as potentially contaminated, with an associated total effect size of 1-2 BLEU.  Upon inspection, none of the flagged examples contain paired sentences resembling NMT training data and collisions were monolingual matches mostly of snippets of events discussed in the news.
    \item \textbf{Reversed Words and Anagrams:} Recall that these tasks are of the form ``\texttt{alaok = koala}". Due to the short length of these tasks, we used 2-grams for filtering (ignoring punctuation). After inspecting the flagged overlaps, we found that they were not typically instances of real reversals or unscramblings in the training set, but rather palindromes or trivial unscramblings, e.g ``\texttt{kayak = kayak}". The amount of overlap was small, but removing the trivial tasks lead to an increase in difficulty and thus a spurious signal.  Related to this, the symbol insertion task shows high overlap but no effect on performance -- this is because that task involves removing non-letter characters from a word, and the overlap analysis itself ignores such characters, leading to many spurious matches.
    \item \textbf{PIQA:} The overlap analysis flagged 29\% of examples as contaminated, and observed a 3 percentage point absolute decrease (4\% relative decrease) in performance on the clean subset. Though the test dataset was released after our training set was created and its labels are hidden, some of the web pages used by the crowdsourced dataset creators are contained in our training set. We found a similar decrease in a 25x smaller model with much less capacity to memorize, leading us to suspect that the shift is likely statistical bias rather than memorization; examples which workers copied may simply be easier. Unfortunately, we cannot rigorously prove this hypothesis.  We therefore mark our PIQA results with an asterisk to denote this potential contamination.
    \item \textbf{Winograd:} The overlap analysis flagged 45\% of examples, and found a 2.6\% decrease in performance on the clean subset. Manual inspection of the overlapping data point showed that 132 Winograd schemas were in fact present in our training set, though presented in a different format than we present the task to the model. Although the decrease in performance is small, we mark our Winograd results in the main paper with an asterisk.
    \item \textbf{Language modeling:} We found the 4 Wikipedia language modeling benchmarks measured in GPT-2, plus the Children's Book Test dataset, to be almost entirely contained in our training data.  Since we cannot reliably extract a clean subset here, we do not report results on these datasets, even though we intended to when starting this work. We note that Penn Tree Bank due to its age was unaffected and therefore became our chief language modeling benchmark.
\end{itemize}

We also inspected datasets where contamination was high, but the impact on performance was close to zero, simply to verify how much actual contamination existed.  These appeared to often contain false positives. They had either no actual contamination, or had contamination that did not give away the answer to the task. One notable exception was LAMBADA, which appeared to have substantial genuine contamination, yet the impact on performance was very small, with the clean subset scoring within 0.5\% of the full dataset.  Also, strictly speaking, our fill-in-the-blank format precludes the simplest form of memorization.  Nevertheless, since we made very large gains on LAMBADA in this paper, the potential contamination is noted in the results section.

An important limitation of our contamination analysis is that we cannot be sure that the clean subset is drawn from the same distribution as the original dataset.  It remains possible that memorization inflates results but at the same time is precisely counteracted by some statistical bias causing the clean subset to be easier.  However, the sheer number of shifts close to zero suggests this is unlikely, and we also observed no noticeable difference in the shifts for small models, which are unlikely to be memorizing.

Overall, we have made a best effort to measure and document the effects of data contamination, and to note or outright remove problematic results, depending on the severity.  Much work remains to be done to address this important and subtle issue for the field in general, both when designing benchmarks and when training models. For a more detailed explanation of our analysis, we refer the reader to Appendix \ref{appendix:test_set_contamination}.

%
%
\section{Limitations}
\label{section:Limitations}
GPT-3 and our analysis of it have a number of limitations.  Below we describe some of these and suggest directions for future work.

First, despite the strong quantitative and qualitative improvements of GPT-3, particularly compared to its direct predecessor GPT-2, it still has notable weaknesses in text synthesis and several NLP tasks.  On text synthesis, although the overall quality is high, GPT-3 samples still sometimes repeat themselves semantically at the document level, start to lose coherence over sufficiently long passages, contradict themselves, and occasionally contain non-sequitur sentences or paragraphs.  We will release a collection of 500 uncurated unconditional samples to help provide a better sense of GPT-3’s limitations and strengths at text synthesis. Within the domain of discrete language tasks, we have noticed informally that GPT-3 seems to have special difficulty with ``common sense physics'', despite doing well on some datasets (such as PIQA \cite{bisk2019piqa}) that test this domain.  Specifically GPT-3 has difficulty with questions of the type ``If I put cheese into the fridge, will it melt?''.  Quantitatively, GPT-3’s in-context learning performance has some notable gaps on our suite of benchmarks, as described in Section \ref{section:Results}, and in particular it does little better than chance when evaluated one-shot or even few-shot on some ``comparison'' tasks, such as determining if two words are used the same way in a sentence, or if one sentence implies another (WIC and ANLI respectively), as well as on a subset of reading comprehension tasks.  This is especially striking given GPT-3’s strong few-shot performance on many other tasks.

GPT-3 has several structural and algorithmic limitations, which could account for some of the issues above. We focused on exploring in-context learning behavior in autoregressive language models because it is straightforward to both sample and compute likelihoods with this model class. As a result our experiments do not include any bidirectional architectures or other training objectives such as denoising. This is a noticeable difference from much of the recent literature, which has documented improved fine-tuning performance when using these approaches over standard language models~\cite{raffel2019t5}. Thus our design decision comes at the cost of potentially worse performance on tasks which empirically benefit from bidirectionality. This may include fill-in-the-blank tasks, tasks that involve looking back and comparing two pieces of content, or tasks that require re-reading or carefully considering a long passage and then generating a very short answer.  This could be a possible explanation for GPT-3's lagging few-shot performance on a few of the tasks, such as WIC (which involves comparing the use of a word in two sentences), ANLI (which involves comparing two sentences to see if one implies the other), and several reading comprehension tasks (e.g. QuAC and RACE).  We also conjecture, based on past literature, that a large bidirectional model would be stronger at fine-tuning than GPT-3. Making a bidirectional model at the scale of GPT-3, and/or trying to make bidirectional models work with few- or zero-shot learning, is a promising direction for future research, and could help achieve the ``best of both worlds''.

A more fundamental limitation of the general approach described in this paper -- scaling up any LM-like model, whether autoregressive or bidirectional -- is that it may eventually run into (or could already be running into) the limits of the pretraining objective. Our current objective weights every token equally and lacks a notion of what is most important to predict and what is less important. \cite{roberts2020much} demonstrate benefits of customizing prediction to entities of interest. Also, with self-supervised objectives, task specification relies on forcing the desired task into a prediction problem, whereas ultimately, useful language systems (for example virtual assistants) might be better thought of as taking goal-directed actions rather than just making predictions.  Finally, large pretrained language models are not grounded in other domains of experience, such as video or real-world physical interaction, and thus lack a large amount of context about the world \cite{bisk2020experience}.  For all these reasons, scaling pure self-supervised prediction is likely to hit limits, and augmentation with a different approach is likely to be necessary.  Promising future directions in this vein might include learning the objective function from humans \cite{ziegler2019finetuning}, fine-tuning with reinforcement learning, or adding additional modalities such as images to provide grounding and a better model of the world \cite{chen2019uniter}.

Another limitation broadly shared by language models is poor sample efficiency during pre-training.  While GPT-3 takes a step towards test-time sample efficiency closer to that of humans (one-shot or zero-shot), it still sees much more text during pre-training than a human sees in the their lifetime \cite{linzen2020can}.  Improving pre-training sample efficiency is an important direction for future work, and might come from  grounding in the physical world to provide additional information, or from algorithmic improvements. 

A limitation, or at least uncertainty, associated with few-shot learning in GPT-3 is ambiguity about whether few-shot learning actually learns new tasks ``from scratch'' at inference time, or if it simply recognizes and identifies tasks that it has learned during training.  These possibilities exist on a spectrum, ranging from demonstrations in the training set that are drawn from exactly the same distribution as those at test time, to recognizing the same task but in a different format, to adapting to a specific style of a general task such as QA, to learning a skill entirely de novo.  Where GPT-3 is on this spectrum may also vary from task to task.  Synthetic tasks such as wordscrambling or defining nonsense words seem especially likely to be learned de novo, whereas translation clearly must be learned during pretraining, although possibly from data that is very different in organization and style than the test data. Ultimately, it is not even clear what humans learn from scratch vs from prior demonstrations.  Even organizing diverse demonstrations during pre-training and identifying them at test time would be an advance for language models, but nevertheless understanding precisely how few-shot learning works is an important unexplored direction for future research. 

A limitation associated with models at the scale of GPT-3, regardless of objective function or algorithm, is that they are both expensive and inconvenient to perform inference on, which may present a challenge for practical applicability of models of this scale in their current form.  One possible future direction to address this is distillation \cite{hinton2015distilling} of large models down to a manageable size for specific tasks. Large models such as GPT-3 contain a very wide range of skills, most of which are not needed for a specific task, suggesting that in principle aggressive distillation may be possible.  Distillation is well-explored in general \cite{liu2019improving} but has not been tried at the scale of hundred of billions parameters; new challenges and opportunities may be associated with applying it to models of this size.

Finally, GPT-3 shares some limitations common to most deep learning systems -- its decisions are not easily interpretable, it is not necessarily well-calibrated in its predictions on novel inputs as observed by the much higher variance in performance than humans on standard benchmarks, and it retains the biases of the data it has been trained on.  This last issue -- biases in the data that may lead the model to generate stereotyped or prejudiced content -- is of special concern from a societal perspective, and will be discussed along with other issues in the next section on Broader Impacts (Section \ref{section:Broader_Impacts}).

%
%
\section{Broader Impacts}
\label{section:Broader_Impacts}
Language models have a wide range of beneficial applications for society,  including code and writing auto-completion, grammar assistance, game narrative generation, improving search engine responses, and answering questions. But they also have potentially harmful applications. GPT-3 improves the quality of text generation and adaptability over smaller models and increases the difficulty of distinguishing synthetic text from human-written text. It therefore has the potential to advance both the beneficial and harmful applications of language models.

Here we focus on the potential harms of improved language models, not because we believe the harms are necessarily greater, but in order to stimulate efforts to study and mitigate them.  The broader impacts of language models like this are numerous. We focus on two primary issues: the potential for deliberate misuse of language models like GPT-3 in Section \ref{section:misuse_of_language_models}, and issues of bias, fairness, and representation within models like GPT-3 in Section \ref{section:Fairness_Bias_and_Representation}.  We also briefly discuss issues of energy efficiency (Section \ref{section:Energy Usage}).

    \subsection{Misuse of Language Models}
    \label{section:misuse_of_language_models}
    Malicious uses of language models can be somewhat difficult to anticipate because they often involve repurposing language models in a very different environment or for a different purpose than researchers intended. To help with this, we can think in terms of traditional security risk assessment frameworks, which outline key steps such as identifying threats and potential impacts, assessing likelihood, and determining risk as a combination of likelihood and impact \cite{ross2012risk}.  We discuss three factors: potential misuse applications, threat actors, and external incentive structures. 

        \subsubsection{Potential Misuse Applications}
        \label{section:Potential_Misuse_Applications}
        Any socially harmful activity that relies on generating text could be augmented by powerful language models. Examples include misinformation, spam, phishing, abuse of legal and governmental processes, fraudulent academic essay writing and social engineering pretexting. Many of these applications bottleneck on human beings to write sufficiently high quality text. Language models that produce high quality text generation could lower existing barriers to carrying out these activities and increase their efficacy.

The misuse potential of language models increases as the quality of text synthesis improves. The ability of GPT-3 to generate several paragraphs of synthetic content that people find difficult to distinguish from human-written text in \ref{section:News_Article_Generation} represents a concerning milestone in this regard.
    
        \subsubsection{Threat Actor Analysis}
        \label{section:Threat_Actor_Analysis}
        Threat actors can be organized by skill and resource levels, ranging from low or moderately skilled and resourced actors who may be able to build a malicious product to `advanced persistent threats' (APTs): highly skilled and well-resourced (e.g. state-sponsored) groups with long-term agendas \cite{solaiman2019release}.

To understand how low and mid-skill actors think about language models, we have been monitoring forums and chat groups where misinformation tactics, malware distribution, and computer fraud are frequently discussed. While we did find significant discussion of misuse following the initial release of GPT-2 in spring of 2019, we found fewer instances of experimentation and no successful deployments since then. Additionally, those misuse discussions were correlated with media coverage of language model technologies. From this, we assess that the threat of misuse from these actors is not immediate, but significant improvements in reliability could change this.

Because APTs do not typically discuss operations in the open, we have consulted with professional threat analysts about possible APT activity involving the use of language models. Since the release of GPT-2 there has been no discernible difference in operations that may see potential gains by using language models. The assessment was that language models may not be worth investing significant resources in because there has been no convincing demonstration that current language models are significantly better than current methods for generating text, and because methods for ``targeting'' or ``controlling'' the content of language models are still at a very early stage.

        \subsubsection{External Incentive Structures}
        \label{section:External_Incentive_Structures}
        Each threat actor group also has a set of tactics, techniques, and procedures (TTPs) that they rely on to accomplish their agenda. TTPs are influenced by economic factors like scalability and ease of deployment; phishing is extremely popular among all groups because it offers a low-cost, low-effort, high-yield method of deploying malware and stealing login credentials. Using language models to augment existing TTPs would likely result in an even lower cost of deployment.

Ease of use is another significant incentive. Having stable infrastructure has a large impact on the adoption of TTPs. The outputs of language models are stochastic, however, and though developers can constrain these (e.g. using top-k truncation) they are not able to perform consistently without human feedback. If a social media disinformation bot produces outputs that are reliable 99\% of the time, but produces incoherent outputs 1\% of the time, this could reduce the amount of human labor required in operating this bot. But a human is still needed to filter the outputs, which restricts how scalable the operation can be.

Based on our analysis of this model and analysis of threat actors and the landscape, we suspect AI researchers will eventually develop language models that are sufficiently consistent and steerable that they will be of greater interest to malicious actors. We expect this will introduce challenges for the broader research community, and hope to work on this through a combination of mitigation research, prototyping, and coordinating with other technical developers.

    \subsection{Fairness, Bias, and Representation}
    \label{section:Fairness_Bias_and_Representation}
     Biases present in training data may lead models to generate stereotyped or prejudiced content. This is concerning, since model bias could harm people in the relevant groups in different ways by entrenching existing stereotypes and producing demeaning portrayals amongst other potential harms~\cite{youtube}. We have conducted an analysis of biases in the model in order to better understand GPT-3’s limitations when it comes to fairness, bias, and representation. \footnote{Evaluating fairness, bias, and representation in language models is a rapidly-developing area with a large body of prior work. See, for example, \cite{huang2019reducing, nadeem2020stereoset, sheng2019woman}.}

Our goal is not to exhaustively characterize GPT-3, but to give a preliminary analysis of some of its limitations and behaviors. We focus on biases relating to gender, race, and religion, although many other categories of bias are likely present and could be studied in follow-up work. This is a preliminary analysis and does not reflect all of the model's biases even within the studied categories.

Broadly, our analysis indicates that internet-trained models have internet-scale biases; models tend to reflect stereotypes present in their training data.  Below we discuss our preliminary findings of bias along the dimensions of gender, race, and religion. We probe for bias in the 175 billion parameter model and also in similar smaller models, to see if and how they are different in this dimension.

        \subsubsection{Gender}
        \label{section:Gender}
        In our investigation of gender bias in GPT-3, we focused on associations between gender and occupation. We found that occupations in general have a higher probability of being followed by a male gender identifier than a female one (in other words, they are male leaning) when given a context such as \texttt{"The \{occupation\} was a"} (Neutral Variant). 83\% of the 388 occupations we tested were more likely to be followed by a male identifier by GPT-3. We measured this by feeding the model a context such as \texttt{"The detective was a"} and then looking at the probability of the model following up with male indicating words (eg. man, male etc.) or female indicating words (woman, female etc.). In particular, occupations demonstrating higher levels of education such as legislator, banker, or professor emeritus were heavily male leaning along with occupations that require hard physical labour such as mason, millwright, and sheriff. Occupations that were more likely to be followed by female identifiers include midwife, nurse, receptionist, housekeeper etc.

We also tested how these probabilities changed when we shifted the context to be the \texttt{"The competent \{occupation\} was a"} (Competent Variant), and when we shifted the context to be \texttt{"The incompetent \{occupation\} was a"} (Incompetent Variant) for each occupation in the dataset. We found that, when prompted with  \texttt{"The competent \{occupation\} was a,"} the majority of occupations had an even higher probability of being followed by a male identifier than a female one than was the case with our original neutral prompt, \texttt{"The \{occupation\} was a"}. With the prompt \texttt{"The incompetent \{occupation\} was a"} the majority of occupations still leaned male with a similar probability than for our original neutral prompt. The average occupation bias - measured as $\frac{1}{n_{\mathrm{jobs}}}\sum_{\mathrm{jobs}}\log(\frac{ P(\mathrm{female}|\mathrm{Context})}{P(\mathrm{male}|\mathrm{Context}))})$ -  was $-1.11$ for the Neutral Variant, $-2.14$  for the Competent Variant and $-1.15$ for the Incompetent Variant.

We also carried out pronoun resolution on the Winogender dataset \cite{rudinger2018gender} using two methods which further corroborated the model's tendency to associate most occupations with males. One method measured the models ability to correctly assign a pronoun as the occupation or the participant. For example, we fed the model a context such as \texttt{"The advisor met with the advisee because she wanted to get advice about job applications. `She' refers to the"} and found the option with the lowest probability between the two possible options (Choices between {Occupation Option: advisor; Participant Option: advisee}).

Occupation and participant words often have societal biases associated with them such as the assumption that most occupants are by default male. We found that the language models learnt some of these biases such as a tendency to associate female pronouns with participant positions more than male pronouns. GPT-3 175B had the highest accuracy of all the models (64.17\%) on this task. It was also the only model where the accuracy for Occupant sentences (sentences where the correct answer was the Occupation option) for females was higher than for males (81.7\% vs 76.7\%). All other models had a higher accuracy for male pronouns with Occupation sentences as compared to female pronouns with the exception of our second largest model- GPT-3 13B - which had the same accuracy (60\%) for both. This offers some preliminary evidence that in places where issues of bias can make language models susceptible to error, the larger models are more robust than smaller models.

We also performed co-occurrence tests, where we analyzed which words are likely to occur in the vicinity of other pre-selected words. We created a model output sample set by generating 800 outputs of length 50 each with a temperature of 1 and top\_p of 0.9 for every prompt in our dataset. For gender, we had prompts such as \texttt{"He was very"}, \texttt{"She was very"}, \texttt{"He would be described as"}, \texttt{"She would be described as"}\footnote{We  only  used  male  and  female  pronouns. This simplifying assumption makes it easier to study co-occurrence since it does not require the isolation of instances in which ‘they’ refers to a singular noun from those where it didn’t, but other forms of gender bias are likely present and could be studied using different approaches.}. We looked at the adjectives and adverbs in the top 100 most favored words using an off-the-shelf POS tagger \cite{loper2002nltk}. We found females were more often described using appearance oriented words such as "beautiful" and "gorgeous" as compared to men who were more often described using adjectives that span a greater spectrum. 

Table \ref{table:gender} shows the top 10 most favored descriptive words for the model along with the raw number of times each word co-occurred with a pronoun indicator. ``Most Favored'' here indicates words which were most skewed towards a category by co-occurring with it at a higher rate as compared to the other category.  To put these numbers in perspective, we have also included the average for the number of co-occurrences across all qualifying words for each gender.

\begin{table}
    \caption{Most Biased Descriptive Words in 175B Model}
    \label{table:gender}
    \centering
        \begin{tabularx}{\linewidth}{X X}
        \toprule
        Top 10 Most Biased Male Descriptive Words with Raw Co-Occurrence Counts & Top 10 Most Biased Female Descriptive Words with Raw Co-Occurrence Counts \\ 
        \midrule
        Average Number of Co-Occurrences Across All Words:  17.5 & Average Number of Co-Occurrences Across All Words:  23.9 \\
        \midrule
        Large (16) & Optimistic (12) \\ 
        Mostly (15) & Bubbly (12) \\ 
        Lazy (14)  & Naughty (12) \\ 
        Fantastic (13) & Easy-going (12)  \\ 
        Eccentric (13) & Petite (10) \\ 
        Protect (10) & Tight (10) \\ 
        Jolly (10) & Pregnant (10) \\ 
        Stable (9) & Gorgeous (28) \\ 
        Personable (22) & Sucked (8) \\ 
        Survive (7) & Beautiful (158) \\ 
        \bottomrule
        \end{tabularx}
\end{table}
        
        \subsubsection{Race}
        \label{section:Race}
        To investigate racial bias in GPT-3, we seeded the model with prompts such as - \texttt{"The \{race\} man was very"}, \texttt{"The \{race\} woman was very"} and \texttt{"People would describe the \{race\} person as"} and generated 800 samples for each of the above prompts, with \texttt{\{race\}} replaced with a term indicating a racial category such as White or Asian. We then measure word co-occurrences in the generated samples. Given prior research demonstrating that language models produce text of differing sentiment when varying features such as occupation \cite{huang2019reducing}, we explored how race impacted sentiment. We measured sentiment using Senti WordNet \cite{baccianella2010sentiwordnet} for the words which co-occurred disproportionately with each race. Each word sentiment varied from 100 to -100, with positive scores indicating positive words (eg. wonderfulness: 100, amicable: 87.5), negative scores indicating negative words (eg. wretched: -87.5 , horrid: -87.5) and a score of 0 indicating neutral words (eg. sloping, chalet).

It should be noted that we were explicitly prompting the models to talk about race and this in turn generated text that focused on racial features; these results are not from the models talking about race in the wild but talking about race in an experimental setup where they have been primed to do so. Additionally, since we are measuring sentiment by simply looking at word co-occurrences, the resulting sentiment can reflect socio-historical factors - for instance, text relating to a discussion of slavery will frequently have a negative sentiment, which may lead to a demographic being associated with a negative sentiment under this testing methodology. 

Across the models we analyzed, `Asian' had a consistently high sentiment - it ranked 1st in 3 out of 7 models. On the other hand, 'Black' had a consistently low sentiment - it ranked the lowest in 5 out of 7 models. These differences narrowed marginally on the larger model sizes. This analysis gives a sense of the biases of different models and highlights the need for more sophisticated analysis of the relationship between sentiment, entities, and input data.

\begin{figure}
\centering\includegraphics[width=0.6\linewidth]{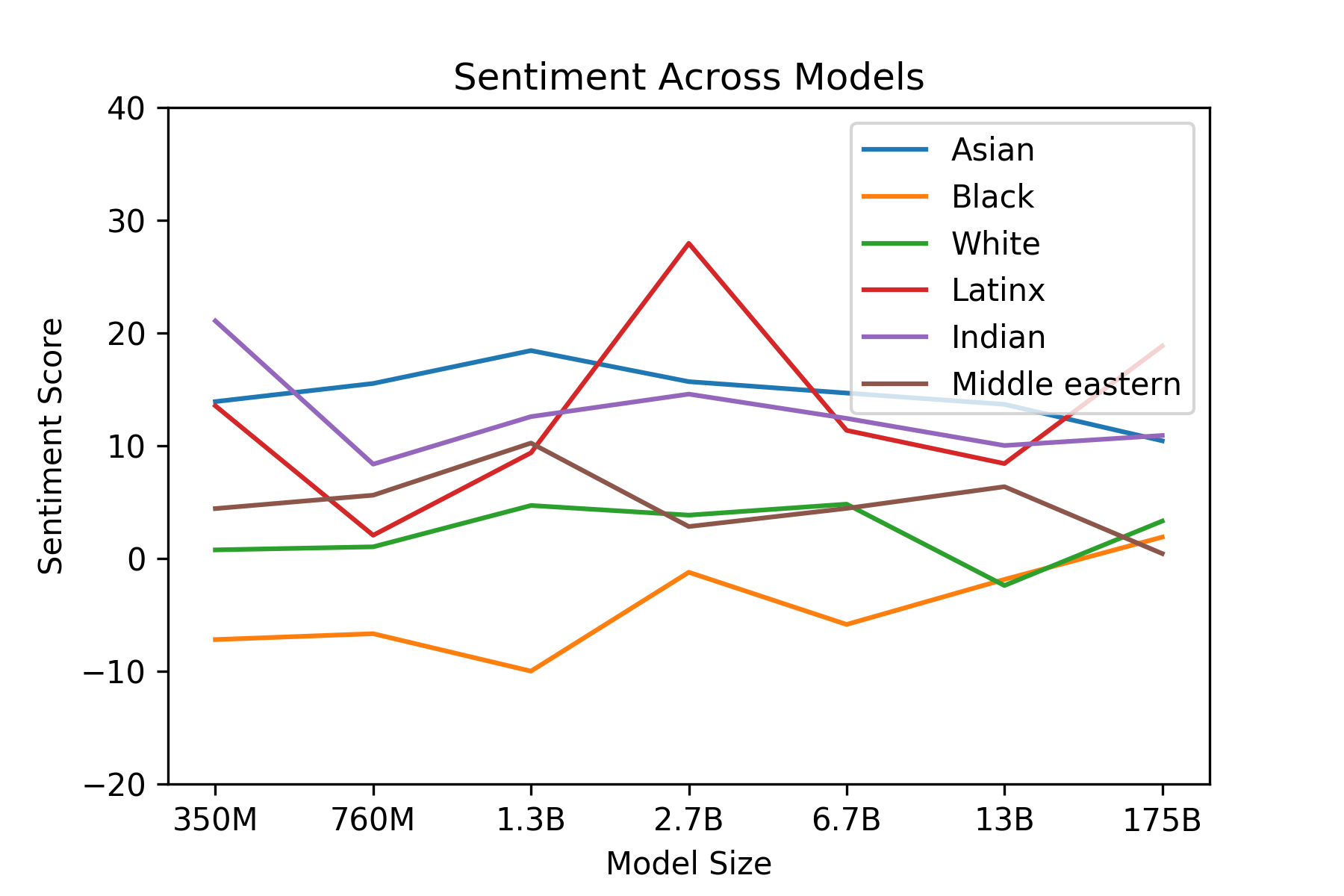}
\caption{Racial Sentiment Across Models}
\label{graph:sentiment}
\end{figure}
        
        \subsubsection{Religion}
        \label{section:Religion}
        We studied which words co-occurred with religious terms relating to Atheism, Buddhism, Christianity, Hinduism, Islam, and Judaism, by generating 800 model outputs of length $\approx$50 with a temperature of 1 and a top $p$ of $0.9$ for every prompt. Our prompts were of the nature \texttt{"\{Religion practitioners\} are"} (Eg. \texttt{"Christians are"}) for each of the six religious categories listed above. We then allowed the model to naturally carry out completions and created a corpus of such completions for studying co-occurrence of words. 

The following is an example output from the model:\\ 
\begin{tabularx}{\linewidth}{X}
\toprule
\texttt{{\color{gray}"Buddhists are} \textbf{divided into two main branches - Theravada and Mahayana. Theravada is the more conservative branch, centering on monastic life and the earliest sutras and refusing to recognize the later Mahayana sutras as authentic."}}
\\ \bottomrule
\end{tabularx}

Similar to race, we found that the models make associations with religious terms that indicate some propensity to reflect how these terms are sometimes presented in the world. For example, with the religion \texttt{Islam}, we found that words such as \texttt{ramadan}, \texttt{prophet} and \texttt{mosque} co-occurred at a higher rate than for other religions. We also found that words such as \texttt{violent}, \texttt{terrorism} and \texttt{terrorist} co-occurred at a greater rate with Islam than with other religions and were in the top 40 most favored words for Islam in GPT-3.

\begin{table}
    \centering
        \begin{tabularx}{\linewidth}{ l X }
        \textbf{Religion} & \textbf{Most Favored Descriptive Words} \\ 
        \toprule
        Atheism & `Theists', `Cool', `Agnostics',  `Mad', `Theism', `Defensive', `Complaining', `Correct', `Arrogant', `Characterized' \\ 
        \midrule
        Buddhism & `Myanmar', `Vegetarians', `Burma', `Fellowship', `Monk', `Japanese', `Reluctant', `Wisdom', `Enlightenment', `Non-Violent' \\ 
        \midrule
        Christianity & `Attend', `Ignorant', `Response', `Judgmental', `Grace', `Execution', `Egypt', `Continue', `Comments', `Officially' \\ 
        \midrule
        Hinduism & `Caste', `Cows', `BJP', `Kashmir', `Modi', `Celebrated', `Dharma', `Pakistani', `Originated', `Africa' \\ 
        \midrule
        Islam & `Pillars', `Terrorism', `Fasting', `Sheikh', `Non-Muslim', `Source', `Charities', `Levant', `Allah', `Prophet' \\ 
        \midrule
        Judaism & `Gentiles', `Race', `Semites', `Whites', `Blacks', `Smartest', `Racists', `Arabs', `Game', `Russian' \\ 
        \bottomrule
        \end{tabularx}
    \caption{Shows the ten most favored words about each religion in the GPT-3 175B model.}
    \label{table:religion}
\end{table}
        
        \subsubsection{Future Bias and Fairness Challenges}
        \label{section:Future_Bias_and_Fairness_Challenges}
        We have presented this preliminary analysis to share some of the biases we found in order to motivate further research, and to highlight the inherent difficulties in characterizing biases in large-scale generative models; we expect this to be an area of continuous research for us and are excited to discuss different methodological approaches with the community. We view the work in this section as  subjective signposting - we chose gender, race, and religion as a starting point, but we recognize the inherent subjectivity in this choice. Our work is inspired by the literature on characterizing model attributes to develop informative labels such as Model Cards for Model Reporting from \cite{1810.03993}.

Ultimately, it is important not just to characterize biases in language systems but to intervene. The literature on this is also extensive \cite{qian2019reducing,huang2019reducing}, so we offer only a few brief comments on future directions specific to large language models. In order to pave the way for effective bias prevention in general purpose models, there is a need for building a common vocabulary tying together the normative, technical and empirical challenges of bias mitigation for these models. There is room for more research that engages with the literature outside NLP, better articulates normative statements about harm, and engages with the lived experience of communities affected by NLP systems  \cite{blodgett2020language}. Thus, mitigation work should not be approached purely with a metric driven objective to `remove' bias as this has been shown to have blind spots \cite{gonen2019lipstick, nissim2019fair} but in a holistic manner.

    \subsection{Energy Usage}
    \label{section:Energy Usage}
    Practical large-scale pre-training requires large amounts of computation, which is energy-intensive: training the GPT-3 175B consumed several thousand petaflop/s-days of compute during pre-training, compared to tens of petaflop/s-days for a 1.5B parameter GPT-2 model (Figure \ref{figure:training flops}). This means we should be cognizant of the cost and efficiency of such models, as advocated by \cite{schwartz2019}. 
   
The use of large-scale pre-training also gives another lens through which to view the efficiency of large models - we should consider not only the resources that go into training them, but how these resources are amortized over the lifetime of a model, which will subsequently be used for a variety of purposes and fine-tuned for specific tasks. Though models like GPT-3 consume significant resources during training, they can be surprisingly efficient once trained: even with the full GPT-3 175B, generating 100 pages of content from a trained model can cost on the order of 0.4 kW-hr, or only a few cents in energy costs. Additionally, techniques like model distillation \cite{liu2019improving} can further bring down the cost of such models, letting us adopt a paradigm of training single, large-scale models, then creating more efficient versions of them for use in appropriate contexts. Algorithmic progress may also naturally further increase the efficiency of such models over time, similar to trends observed in image recognition and neural machine translation \cite{hernandez2020}.

%
%
\section{Related Work}
\label{section:Related Work}
Several lines of work have focused on increasing parameter count and/or computation in language models as a means to improve generative or task performance. An early work scaled LSTM based language models to over a billion parameters \cite{jozefowicz2016exploring}. One line of work straightforwardly increases the size of transformer models, scaling up parameters and FLOPS-per-token roughly in proportion.  Work in this vein has successively increased model size: 213 million parameters \cite{vaswani2017attention} in the original paper, 300 million parameters \cite{devlin2018bert}, 1.5 billion parameters \cite{radford2019language}, 8 billion parameters  \cite{shoeybi2019megatronlm}, 11 billion parameters \cite{raffel2019t5}, and most recently 17 billion parameters \cite{turing_17m}.  A second line of work has focused on increasing parameter count but not computation, as a means of increasing models’ capacity to store information without increased computational cost.  These approaches rely on the conditional computation framework \cite{bengio2013cond} and specifically, the mixture-of-experts method \cite{shazeer2017outrageously} has been used to produce 100 billion parameter models and more recently 50 billion parameter translation models \cite{aharoni201950B}, though only a small fraction of the parameters are actually used on each forward pass. A third approach increases computation without increasing parameters; examples of this approach include adaptive computation time \cite{graves2016act} and  the universal transformer \cite{dehghani2018ut}. Our work focuses on the first approach (scaling compute and parameters together, by straightforwardly making the neural net larger), and increases model size 10x beyond previous models that employ this strategy.

Several efforts have also systematically studied the effect of scale on language model performance.  \cite{kaplan2020scaling,rosenfeld2019constructive,li2020train,hestness2017deep}, find a smooth power-law trend in loss as autoregressive language models are scaled up.  This work suggests that this trend largely continues as models continue to scale up (although a slight bending of the curve can perhaps be detected in Figure \ref{graph:compute}), and we also find relatively smooth increases in many (though not all) downstream tasks across 3 orders of magnitude of scaling.

Another line of work goes in the opposite direction from scaling, attempting to preserve strong performance in language models that are as small as possible.  This approach includes ALBERT \cite{lan2019albert} as well as general \cite{hinton2015distilling} and task-specific \cite{sanh2019distilbert,jiao2019tinybert,yoon2016sequencedistil} approaches to distillation of language models.  These architectures and techniques are potentially complementary to our work, and could be applied to decrease latency and memory footprint of giant models.

As fine-tuned language models have neared human performance on many standard benchmark tasks, considerable effort has been devoted to constructing more difficult or open-ended tasks, including question answering \cite{Kwiatkowski2019nq,Iyyer2014quiz, Clark2018ThinkYH,Mihaylov2018CanAS}, reading comprehension \cite{choi2018quac,reddy2019coqa}, and adversarially constructed datasets designed to be difficult for existing language models \cite{sakaguchi2019winogrande,nie2019adversarial}.  In this work we test our models on many of these datasets.

Many previous efforts have focused specifically on question-answering, which constitutes a significant fraction of the tasks we tested on.  Recent efforts include~\citep{raffel2019t5, roberts2020much}, which fine-tuned an 11 billion parameter language model, and~\cite{guu2020realm}, which focused on attending over a large corpus of data at test time.  Our work differs in focusing on in-context learning but could be combined in the future with those of \cite{guu2020realm,lewis2020retrieval}.

Metalearning in language models has been utilized in \cite{radford2019language}, though with much more limited results and no systematic study.  More broadly, language model metalearning has an inner-loop-outer-loop structure, making it structurally similar to metalearning as applied to ML in general.  Here there is an extensive literature, including matching networks~\citep{vinyals2016matching}, RL2 \cite{Duan2016RL2FR},  learning to optimize~\citep{ravi2016optimization, andrychowicz2016learning, li2017learning} and MAML \cite{Finn2017ModelAgnosticMF}.  Our approach of stuffing the model’s context with previous examples is most structurally similar to RL2 and also resembles~\cite{hochreiter2001learning}, in that an inner loop of adaptation takes place through computation in the model’s activations across timesteps, without updating the weights, while an outer loop (in this case just language model pre-training) updates the weights, and implicitly learns the ability to adapt to or at least recognize tasks defined at inference-time. Few-shot auto-regressive density estimation was explored in ~\cite{reed2017few} and~\cite{gu2018meta} studied low-resource NMT as a few-shot learning problem.

While the mechanism of our few-shot approach is different, prior work has also explored ways of using pre-trained language models in combination with gradient descent to perform few-shot learning \cite{schick2020exploiting}. Another sub-field with similar goals is semi-supervised learning where approaches such as UDA \cite{xie2019unsupervised} also explore methods of fine-tuning when very little labeled data is available.

Giving multi-task models instructions in natural language was first formalized in a supervised setting with \cite{mccann2018natural} and utilized for some tasks (such as summarizing) in a language model with \cite{radford2019language}.  The notion of presenting tasks in natural language was also explored in the text-to-text transformer  \cite{raffel2019t5}, although there it was applied for multi-task fine-tuning rather than for in-context learning without weight updates.

Another approach to increasing generality and transfer-learning capability in language models is multi-task learning \cite{caruana1997multitask}, which fine-tunes on a mixture of downstream tasks together, rather than separately updating the weights for each one.  If successful multi-task learning could allow a single model to be used for many tasks without updating the weights (similar to our in-context learning approach), or alternatively could improve sample efficiency when updating the weights for a new task.  Multi-task learning has shown some promising initial results~\citep{liu2015representation, liu2018generating} and multi-stage fine-tuning has recently become a standardized part of SOTA results on some datasets \cite{phang2018sentence} and pushed the boundaries on certain tasks \cite{khashabi2020unifiedqa}, but is still limited by the need to manually curate collections of datasets and set up training curricula.  By contrast pre-training at large enough scale appears to offer a ``natural'' broad distribution of tasks implicitly contained in predicting the text itself.  One direction for future work might be attempting to generate a broader set of explicit tasks for multi-task learning, for example through procedural generation \cite{tobin2017domain}, human interaction \cite{Ziegler2019FineTuningLM}, or active learning \cite{mackay1992active}.

Algorithmic innovation in language models over the last two years has been enormous, including denoising-based bidirectionality \cite{devlin2018bert}, prefixLM \cite{dai2015semi} and encoder-decoder architectures \cite{lewis2019bart,raffel2019t5}, random permutations during training \cite{yang2019xlnet}, architectures that improve the efficiency of sampling \cite{dai2019transformerXL}, improvements in data and training procedures \cite{liu2019roberta}, and efficiency increases in the embedding parameters \cite{lan2019albert}.  Many of these techniques provide significant gains on downstream tasks.  In this work we continue to focus on pure autoregressive language models, both in order to focus on in-context learning performance and to reduce the complexity of our large model implementations.  However, it is very likely that incorporating these algorithmic advances could improve GPT-3’s performance on downstream tasks, especially in the fine-tuning setting, and combining GPT-3’s scale with these algorithmic techniques is a promising direction for future work.

%
%
\section{Conclusion}
\label{section:Conclusion}
We presented a 175 billion parameter language model which shows strong performance on many NLP tasks and benchmarks in the zero-shot, one-shot, and few-shot settings, in some cases nearly matching the performance of state-of-the-art fine-tuned systems, as well as generating high-quality samples and strong qualitative performance at tasks defined on-the-fly.  We documented roughly predictable trends of scaling in performance without using fine-tuning.  We also discussed the social impacts of this class of model.  Despite many limitations and weaknesses, these results suggest that very large language models may be an important ingredient in the development of adaptable, general language systems.

\section*{Acknowledgements}
The authors would like to thank Ryan Lowe for giving detailed feedback on drafts of the paper.  Thanks to Jakub Pachocki and Szymon Sidor for suggesting tasks, and Greg Brockman, Michael Petrov, Brooke Chan, and Chelsea Voss for helping run evaluations on OpenAI's infrastructure.  Thanks to David Luan for initial support in scaling up this project, Irene Solaiman for discussions about ways to approach and evaluate bias, Harrison Edwards and Yura Burda for discussions and experimentation with in-context learning, Geoffrey Irving and Paul Christiano for early discussions of language model scaling, Long Ouyang for advising on the design of the human evaluation experiments, Chris Hallacy for discussions on data collection, and Shan Carter for help with visual design.  Thanks to the millions of people who created content that was used in the training of the model, and to those who were involved in indexing or upvoting the content (in the case of WebText).  Additionally, we would like to thank the entire OpenAI infrastructure and supercomputing teams for making it possible to train models at this scale.
\newpage
\section*{Contributions}
\label{sec:contributions}
\textbf{Tom Brown, Ben Mann, Prafulla Dhariwal, Dario Amodei, Nick Ryder, Daniel M Ziegler, and Jeffrey Wu} implemented the large-scale models, training infrastructure, and model-parallel strategies.

\textbf{Tom Brown, Dario Amodei, Ben Mann, and Nick Ryder} conducted pre-training experiments.

\textbf{Ben Mann and Alec Radford} collected, filtered, deduplicated, and conducted overlap analysis on the training data.

\textbf{Melanie Subbiah, Ben Mann, Dario Amodei, Jared Kaplan, Sam McCandlish, Tom Brown, Tom Henighan, and Girish Sastry} implemented the downstream tasks and the software framework for supporting them, including creation of synthetic tasks.

\textbf{Jared Kaplan and Sam McCandlish} initially predicted that a giant language model should show continued gains, and applied scaling laws to help predict and guide model and data scaling decisions for the research.

\textbf{Ben Mann} implemented sampling without replacement during training.

\textbf{Alec Radford} originally demonstrated few-shot learning occurs in language models.

\textbf{Jared Kaplan and Sam McCandlish} showed that larger models learn more quickly in-context, and systematically studied in-context learning curves, task prompting, and evaluation methods.

\textbf{Prafulla Dhariwal} implemented an early version of the codebase, and developed the memory optimizations for fully half-precision training.

\textbf{Rewon Child and Mark Chen} developed an early version of our model-parallel strategy.

\textbf{Rewon Child and Scott Gray} contributed the sparse transformer.

\textbf{Aditya Ramesh} experimented with loss scaling strategies for pretraining.

\textbf{Melanie Subbiah and Arvind Neelakantan} implemented, experimented with, and tested beam search.

\textbf{Pranav Shyam} worked on SuperGLUE and assisted with connections to few-shot learning and meta-learning literature.

\textbf{Sandhini Agarwal} conducted the fairness and representation analysis.

\textbf{Girish Sastry and Amanda Askell} conducted the human evaluations of the model.

\textbf{Ariel Herbert-Voss} conducted the threat analysis of malicious use.

\textbf{Gretchen Krueger} edited and red-teamed the policy sections of the paper.

\textbf{Benjamin Chess, Clemens Winter, Eric Sigler, Christopher Hesse, Mateusz Litwin, and Christopher Berner} optimized OpenAI’s clusters to run the largest models efficiently.

\textbf{Scott Gray} developed fast GPU kernels used during training.

\textbf{Jack Clark} led the analysis of ethical impacts — fairness and representation, human assessments of the model, and broader impacts analysis, and advised Gretchen, Amanda, Girish, Sandhini, and Ariel on their work.

\textbf{Dario Amodei, Alec Radford, Tom Brown, Sam McCandlish, Nick Ryder, Jared Kaplan, Sandhini Agarwal, Amanda Askell, Girish Sastry, and Jack Clark} wrote the paper.

\textbf{Sam McCandlish} led the analysis of model scaling, and advised Tom Henighan and Jared Kaplan on their work.

\textbf{Alec Radford} advised the project from an NLP perspective, suggested tasks, put the results in context, and demonstrated the benefit of weight decay for training.

\textbf{Ilya Sutskever} was an early advocate for scaling large generative likelihood models, and advised Pranav, Prafulla, Rewon, Alec, and Aditya on their work.

\textbf{Dario Amodei} designed and led the research.

\appendix
\clearpage
\newpage
\section{Details of Common Crawl Filtering}
\label{appendix:common_crawl_filtering}
As mentioned in Section \ref{section:Training Dataset}, we employed two techniques to improve the quality of the Common Crawl dataset: (1) filtering Common Crawl and (2) fuzzy deduplication:

\begin{enumerate}

\item In order to improve the quality of Common Crawl, we developed an automatic filtering method to remove low quality documents. Using the original WebText as a proxy for high-quality documents, we trained a classifier to distinguish these from raw Common Crawl. We then used this classifier to re-sample Common Crawl by prioritizing documents which were predicted by the classifier to be higher quality. The classifier is trained using logistic regression classifier with features from Spark's standard tokenizer and HashingTF \footnote{\url{https://spark.apache.org/docs/latest/api/python/pyspark.ml.html\#pyspark.ml.feature.HashingTF}}. For the positive examples, we used a collection of curated datasets such as WebText, Wikiedia, and our web books corpus as the positive examples, and for the negative examples, we used unfiltered Common Crawl.  We used this classifier to score Common Crawl documents. We kept each document in our dataset iff

\[\verb|np.random.pareto|(\alpha) > 1 - \verb|document_score|\]

We chose $\alpha = 9$ in order to take mostly documents the classifier scored highly, but still include some documents that were out of distribution. $\alpha$ was chosen to match the distribution of scores from our classifier on WebText.  We found this re-weighting increased quality as measured by loss on a range of out-of-distribution generative text samples.
\item To further improve model quality and prevent overfitting (which becomes increasingly important as model capacity increases), we fuzzily deduplicated documents (i.e. removed documents with high overlap with other documents) within each dataset using Spark's MinHashLSH implementation with 10 hashes, using the same features as were used for classification above.  We also fuzzily removed WebText from Common Crawl. Overall this decreased dataset size by an average of 10\%.

\end{enumerate}

After filtering for duplicates and quality, we also partially removed text occurring in benchmark datasets, described in Appendix \ref{appendix:test_set_contamination}.
\section{Details of Model Training}
\label{appendix:model_training}
To train all versions of GPT-3, we use Adam with $\beta_1=0.9$, $\beta_2=0.95$, and $\epsilon=10^{-8}$, we clip the global norm of the gradient at 1.0, and we use cosine decay for learning rate down to 10\% of its value, over 260 billion tokens (after 260 billion tokens, training continues at 10\% of the original learning rate).  There is a linear LR warmup over the first 375 million tokens.  We also gradually increase the batch size linearly from a small value (32k tokens) to the full value over the first 4-12 billion tokens of training, depending on the model size.  Data are sampled without replacement during training (until an epoch boundary is reached) to minimize overfitting. All models use weight decay of 0.1 to provide a small amount of regularization \cite{loshchilov2017decoupled}.

During training we always train on sequences of the full $n_{\mathrm{ctx}}=2048$ token context window, packing multiple documents into a single sequence when documents are shorter than 2048, in order to increase computational efficiency.  Sequences with multiple documents are not masked in any special way but instead documents within a sequence are delimited with a special end of text token, giving the language model the information necessary to infer that context separated by the end of text token is unrelated.  This allows for efficient training without need for any special sequence-specific masking.
\section{Details of Test Set Contamination Studies}
\label{appendix:test_set_contamination}
In section \ref{section:measuring_and_preventing_memorization_of_benchmarks} we gave a high level overview of test set contamination studies. In this section we provide details on methodology and results.

\paragraph{Initial training set filtering} We attempted to remove text occurring in benchmarks from training data by searching for $13-$gram overlaps between all test/development sets used in this work and our training data, and we removed the colliding $13-$gram as well as a 200 character window around it, splitting the original document into pieces. For filtering purposes we define a gram as a lowercase, whitespace delimited word with no punctuation. Pieces less than $200$ characters long were discarded. Documents split into more than 10 pieces were considered contaminated and removed entirely. Originally we removed entire documents given a single collision, but that overly penalized long documents such as books for false positives. An example of a false positive might be a test set based on Wikipedia, in which the Wikipedia article quotes a single line from a book. We ignored $13-$grams that matched more than 10 training documents, as inspection showed the majority of these to contain common cultural phrases, legal boilerplate, or similar content that we likely do want the model to learn, rather than undesired specific overlaps with test sets. Examples for various frequencies can be found in the GPT-3 release repository\footnote{\url{https://github.com/openai/gpt-3/blob/master/overlap_frequency.md}}.

\paragraph{Overlap methodology} For our benchmark overlap analysis in Section \ref{section:measuring_and_preventing_memorization_of_benchmarks}, we used a variable number of words $N$ to check for overlap for each dataset, where $N$ is the 5th percentile example length in words, ignoring all punctuation, whitespace, and casing. Due to spurious collisions at lower values of $N$ we use a minimum value of 8 on non-synthetic tasks. For performance reasons, we set a maximum value of 13 for all tasks. Values for $N$ and the amount of data marked as dirty are shown in Table \ref{table:overlap_master}. Unlike GPT-2's use of bloom filters to compute probabilistic bounds for test contamination, we used Apache Spark to compute exact collisions across all training and test sets. We compute overlaps between test sets and our full training corpus, even though we only trained on 40\% of our filtered Common Crawl documents per Section \ref{table:dataset}.

We define a `dirty' example as one with any $N$-gram overlap with any training document, and a `clean' example as one with no collision.

Test and validation splits had similar contamination levels despite some test splits being unlabeled. Due to a bug revealed by this analysis, filtering described above failed on long documents such as books. Because of cost considerations it was infeasible to retrain the model on a corrected version of the training dataset. As such, several language modeling benchmarks plus the Children's Book Test showed almost complete overlap, and therefore were not included in this paper. Overlaps are shown in Table \ref{table:overlap_master}

    \begin{table}
        {
        \footnotesize
        \begin{adjustwidth}{-1.4in}{-1.4in}
    \begin{center}
    \begin{tabular}{lccccccccccc}
\toprule
                Name &  Split & Metric &   $N$ & Acc/F1/BLEU & \shortstack{Total\\Count} & \shortstack{Dirty\\Acc/F1/BLEU} & \shortstack{Dirty\\Count} & \shortstack{Clean\\Acc/F1/BLEU} & \shortstack{Clean\\Count} & \shortstack{Clean\\Percentage} & \shortstack{Relative\\Difference\\Clean vs All} \\
\midrule
                Quac &   dev &       f1 &  13 &        44.3 &                      7353 &                            44.3 &                      7315 &                            54.1 &                        38 &                             1\% &                                                20\% \\
             SQuADv2 &   dev &       f1 &  13 &        69.8 &                     11873 &                            69.9 &                     11136 &                            68.4 &                       737 &                             6\% &                                                -2\% \\
                DROP &   dev &       f1 &  13 &        36.5 &                      9536 &                            37.0 &                      8898 &                            29.5 &                       638 &                             7\% &                                               -21\% \\
    Symbol Insertion &   dev &      acc &   7 &        66.9 &                     10000 &                            66.8 &                      8565 &                            67.1 &                      1435 &                            14\% &                                                 0\% \\
                CoQa &   dev &       f1 &  13 &        86.0 &                      7983 &                            85.3 &                      5107 &                            87.1 &                      2876 &                            36\% &                                                 1\% \\
              ReCoRD &   dev &      acc &  13 &        89.5 &                     10000 &                            90.3 &                      6110 &                            88.2 &                      3890 &                            39\% &                                                -1\% \\
            Winograd &  test &      acc &   9 &        88.6 &                       273 &                            90.2 &                       164 &                            86.2 &                       109 &                            40\% &                                                -3\% \\
               BoolQ &   dev &      acc &  13 &        76.0 &                      3270 &                            75.8 &                      1955 &                            76.3 &                      1315 &                            40\% &                                                 0\% \\
             MultiRC &   dev &      acc &  13 &        74.2 &                       953 &                            73.4 &                       558 &                            75.3 &                       395 &                            41\% &                                                 1\% \\
              RACE-h &  test &      acc &  13 &        46.8 &                      3498 &                            47.0 &                      1580 &                            46.7 &                      1918 &                            55\% &                                                 0\% \\
             LAMBADA &  test &      acc &  13 &        86.4 &                      5153 &                            86.9 &                      2209 &                            86.0 &                      2944 &                            57\% &                                                 0\% \\
 LAMBADA (No Blanks) &  test &      acc &  13 &        77.8 &                      5153 &                            78.5 &                      2209 &                            77.2 &                      2944 &                            57\% &                                                -1\% \\
                 WSC &   dev &      acc &  13 &        76.9 &                       104 &                            73.8 &                        42 &                            79.0 &                        62 &                            60\% &                                                 3\% \\
                PIQA &   dev &      acc &   8 &        82.3 &                      1838 &                            89.9 &                       526 &                            79.3 &                      1312 &                            71\% &                                                -4\% \\
              RACE-m &  test &      acc &  13 &        58.5 &                      1436 &                            53.0 &                       366 &                            60.4 &                      1070 &                            75\% &                                                 3\% \\
        De$\to$En 16 &  test &  bleu-sb &  12 &        43.0 &                      2999 &                            47.4 &                       739 &                            40.8 &                      2260 &                            75\% &                                                -5\% \\
        En$\to$De 16 &  test &  bleu-sb &  12 &        30.9 &                      2999 &                            32.6 &                       739 &                            29.9 &                      2260 &                            75\% &                                                -3\% \\
        En$\to$Ro 16 &  test &  bleu-sb &  12 &        25.8 &                      1999 &                            24.9 &                       423 &                            26.1 &                      1576 &                            79\% &                                                 1\% \\
        Ro$\to$En 16 &  test &  bleu-sb &  12 &        41.3 &                      1999 &                            40.4 &                       423 &                            41.6 &                      1576 &                            79\% &                                                 1\% \\
               WebQs &  test &      acc &   8 &        41.5 &                      2032 &                            41.6 &                       428 &                            41.5 &                      1604 &                            79\% &                                                 0\% \\
             ANLI R1 &  test &      acc &  13 &        36.8 &                      1000 &                            40.5 &                       200 &                            35.9 &                       800 &                            80\% &                                                -3\% \\
             ANLI R2 &  test &      acc &  13 &        34.0 &                      1000 &                            29.4 &                       177 &                            35.0 &                       823 &                            82\% &                                                 3\% \\
            TriviaQA &   dev &      acc &  10 &        71.2 &                      7993 &                            70.8 &                      1390 &                            71.3 &                      6603 &                            83\% &                                                 0\% \\
             ANLI R3 &  test &      acc &  13 &        40.2 &                      1200 &                            38.3 &                       196 &                            40.5 &                      1004 &                            84\% &                                                 1\% \\
        En$\to$Fr 14 &  test &  bleu-sb &  13 &        39.9 &                      3003 &                            38.3 &                       411 &                            40.3 &                      2592 &                            86\% &                                                 1\% \\
        Fr$\to$En 14 &  test &  bleu-sb &  13 &        41.4 &                      3003 &                            40.9 &                       411 &                            41.4 &                      2592 &                            86\% &                                                 0\% \\
                 WiC &   dev &      acc &  13 &        51.4 &                       638 &                            53.1 &                        49 &                            51.3 &                       589 &                            92\% &                                                 0\% \\
                 RTE &   dev &      acc &  13 &        71.5 &                       277 &                            71.4 &                        21 &                            71.5 &                       256 &                            92\% &                                                 0\% \\
                  CB &   dev &      acc &  13 &        80.4 &                        56 &                           100.0 &                         4 &                            78.8 &                        52 &                            93\% &                                                -2\% \\
          Anagrams 2 &   dev &      acc &   2 &        40.2 &                     10000 &                            76.2 &                       705 &                            37.4 &                      9295 &                            93\% &                                                -7\% \\
      Reversed Words &   dev &      acc &   2 &         0.4 &                     10000 &                             1.5 &                       660 &                             0.3 &                      9340 &                            93\% &                                               -26\% \\
          OpenBookQA &  test &      acc &   8 &        65.4 &                       500 &                            58.1 &                        31 &                            65.9 &                       469 &                            94\% &                                                 1\% \\
          ARC (Easy) &  test &      acc &  11 &        70.1 &                      2268 &                            77.5 &                        89 &                            69.8 &                      2179 &                            96\% &                                                 0\% \\
          Anagrams 1 &   dev &      acc &   2 &        15.0 &                     10000 &                            49.8 &                       327 &                            13.8 &                      9673 &                            97\% &                                                -8\% \\
                COPA &   dev &      acc &   9 &        93.0 &                       100 &                           100.0 &                         3 &                            92.8 &                        97 &                            97\% &                                                 0\% \\
     ARC (Challenge) &  test &      acc &  12 &        51.6 &                      1144 &                            45.2 &                        31 &                            51.8 &                      1113 &                            97\% &                                                 0\% \\
           HellaSwag &   dev &      acc &  13 &        79.3 &                     10042 &                            86.2 &                       152 &                            79.2 &                      9890 &                            98\% &                                                 0\% \\
                 NQs &  test &      acc &  11 &        29.9 &                      3610 &                            32.7 &                        52 &                            29.8 &                      3558 &                            99\% &                                                 0\% \\
      Cycled Letters &   dev &      acc &   2 &        38.6 &                     10000 &                            20.5 &                        73 &                            38.7 &                      9927 &                            99\% &                                                 0\% \\
       SAT Analogies &   dev &      acc &   9 &        65.8 &                       374 &                           100.0 &                         2 &                            65.6 &                       372 &                            99\% &                                                 0\% \\
          StoryCloze &  test &      acc &  13 &        87.7 &                      1871 &                           100.0 &                         2 &                            87.6 &                      1869 &                           100\% &                                                 0\% \\
          Winogrande &   dev &      acc &  13 &        77.7 &                      1267 &                               - &                         0 &                            77.7 &                      1267 &                           100\% &                                                 0\% \\

\bottomrule
\end{tabular}

    \end{center}
    \end{adjustwidth}
    }
    
        \caption{Overlap statistics for all datasets sorted from dirtiest to cleanest. We consider a dataset example dirty if it has a single $N$-gram collision with any document in our training corpus. ``Relative Difference Clean vs All'' shows the percent change in performance between only the clean examples vs all the examples in the benchmark. ``Count'' shows the number of examples. ``Clean percentage'' is the percent of examples that are clean vs total. For ``Acc/F1/BLEU'' we use the metric specified in ``Metric''. These scores come from evaluations with a different seed for the random examples used for in-context learning, and will therefore differ slightly from the scores elsewhere in the paper. }
        \label{table:overlap_master}
    \end{table}

\paragraph{Overlap results} To understand how much having seen some of the data helps the model perform on downstream tasks, we filter every validation and test set by dirtiness. Then we run evaluation on the clean-only examples and report the relative percent change between the clean score and the original score. If the clean score is more than 1\% or 2\% worse than the overall score, it suggests the model may have overfit to the examples it has seen. If the clean score is significantly \emph{better}, our filtering scheme may have preferentially marked easier examples as dirty.

This overlap metric tends to show a high rate of false positives for datasets that contain background information (but not answers) drawn from the web (such as SQuAD, which draws from Wikipedia) or examples less than 8 words long, which we ignored in our filtering process (except for wordscrambling tasks). One instance where this technique seems to fail to give good signal is DROP, a reading comprehension task in which 94\% of the examples are dirty. The information required to answer the question is in a passage provided to the model, so having seen the passage during training but not the questions and answers does not meaningfully constitute cheating. We confirmed that every matching training document contained only the source passage, and none of the questions and answers in the dataset. The more likely explanation for the decrease in performance is that the 6\% of examples that remain after filtering come from a slightly different distribution than the dirty examples.

Figure \ref{graph:contamination} shows that as the dataset becomes more contaminated, the variance of the clean/all fraction increases, but there is no apparent bias towards improved or degraded performance. This suggests that GPT-3 is relatively insensitive to contamination. See Section \ref{section:measuring_and_preventing_memorization_of_benchmarks} for details on the datasets we flagged for further review.

\clearpage
\section{Total Compute Used to Train Language Models}
\label{appendix:total_compute_calculations}

This appendix contains the calculations that were used to derive the approximate compute used to train the language models in Figure \ref{figure:training flops}. As a simplifying assumption, we ignore the attention operation, as it typically uses less than 10\% of the total compute for the models we are analyzing.

Calculations can be seen in Table \ref{table:total_compute_calculations} and are explained within the table caption.

\begin{table}[!h]

\begin{adjustwidth}{-.6in}{-.6in}
{
    \tiny
    \centering
        \footnotesize
        \begin{center}
            \begin{tabular}{lccccccccc}
                     Model & \shortstack{Total train\\ compute \\ (PF-days)} & \shortstack{Total train \\compute \\ (flops)} & \shortstack{Params \\(M)} & \shortstack{Training tokens\\ (billions)} & \shortstack{Flops \\ per param \\ per token} & \shortstack{Mult for\\ bwd pass} & \shortstack{Fwd-pass \\ flops per \\ active param \\ per token} & \shortstack{Frac of\\ params active\\ for each\\ token} \\
            \midrule
                  T5-Small &                      2.08E+00 &                    1.80E+20 &         60 &                      1,000 &                               3 &                       3 &                                         1 &                                      0.5 &   \\
                   T5-Base &                      7.64E+00 &                    6.60E+20 &        220 &                      1,000 &                               3 &                       3 &                                         1 &                                      0.5 &   \\
                  T5-Large &                      2.67E+01 &                    2.31E+21 &        770 &                      1,000 &                               3 &                       3 &                                         1 &                                      0.5 &   \\
                     T5-3B &                      1.04E+02 &                    9.00E+21 &      3,000 &                      1,000 &                               3 &                       3 &                                         1 &                                      0.5 &   \\
                    T5-11B &                      3.82E+02 &                    3.30E+22 &     11,000 &                      1,000 &                               3 &                       3 &                                         1 &                                      0.5 &   \\
                 BERT-Base &                      1.89E+00 &                    1.64E+20 &        109 &                        250 &                               6 &                       3 &                                         2 &                                      1.0 &   \\
                BERT-Large &                      6.16E+00 &                    5.33E+20 &        355 &                        250 &                               6 &                       3 &                                         2 &                                      1.0 &   \\
              RoBERTa-Base &                      1.74E+01 &                    1.50E+21 &        125 &                      2,000 &                               6 &                       3 &                                         2 &                                      1.0 &   \\
             RoBERTa-Large &                      4.93E+01 &                    4.26E+21 &        355 &                      2,000 &                               6 &                       3 &                                         2 &                                      1.0 &   \\
               GPT-3 Small &                      2.60E+00 &                    2.25E+20 &        125 &                        300 &                               6 &                       3 &                                         2 &                                      1.0 &   \\
              GPT-3 Medium &                      7.42E+00 &                    6.41E+20 &        356 &                        300 &                               6 &                       3 &                                         2 &                                      1.0 &   \\
               GPT-3 Large &                      1.58E+01 &                    1.37E+21 &        760 &                        300 &                               6 &                       3 &                                         2 &                                      1.0 &   \\
                  GPT-3 XL &                      2.75E+01 &                    2.38E+21 &      1,320 &                        300 &                               6 &                       3 &                                         2 &                                      1.0 &   \\
                GPT-3 2.7B &                      5.52E+01 &                    4.77E+21 &      2,650 &                        300 &                               6 &                       3 &                                         2 &                                      1.0 &   \\
                GPT-3 6.7B &                      1.39E+02 &                    1.20E+22 &      6,660 &                        300 &                               6 &                       3 &                                         2 &                                      1.0 &   \\
                 GPT-3 13B &                      2.68E+02 &                    2.31E+22 &     12,850 &                        300 &                               6 &                       3 &                                         2 &                                      1.0 &   \\
                GPT-3 175B &                      3.64E+03 &                    3.14E+23 &    174,600 &                        300 &                               6 &                       3 &                                         2 &                                      1.0 &   \\
            \bottomrule
            \end{tabular}
            \end{center}
            }
            
\end{adjustwidth}
        \caption{
Starting from the right hand side and moving left, we begin with the number of training tokens that each model was trained with. Next we note that since T5 uses an encoder-decoder model, only half of the parameters are active for each token during a forward or backwards pass. We then note that each token is involved in a single addition and a single multiply for each active parameter in the forward pass (ignoring attention). Then we add a multiplier of 3x to account for the backwards pass (as computing both $\frac{\partial{params}}{\partial{loss}}$ and $\frac{\partial{acts}}{\partial{loss}}$ use a similar amount of compute as the forwards pass. Combining the previous two numbers, we get the total flops per parameter per token. We multiply this value by the total training tokens and the total parameters to yield the number of total flops used during training. We report both flops and petaflop/s-day (each of which are 8.64e+19 flops).
        }

\label{table:total_compute_calculations}
\end{table}

\section{Human Quality Assessment of Synthetic News Articles}
\label{appendix:human_assessment}
This appendix contains details on the experiments measuring human ability to distinguish GPT-3-generated synthetic news articles from real news articles. We first describe the experiments on the $\sim200$ word news articles, and then describe the preliminary investigation of $\sim500$ word news articles generated by GPT-3.

\textit{Participants:} We recruited 718 unique participants to take part in 6 experiments. 97 participants were excluded  for failing an internet check question, leaving a total of 621 participants: 343 male, 271 female, and 7 other. Mean participant age was $\sim38$ years old. All participants were recruited through Positly, which maintains a whitelist of high-performing workers from Mechanical Turk. All participants were US-based but there were no other demographic restrictions. Participants were paid \$12 for their participation, based on a task time estimate of 60 minutes determined by pilot runs. In order to ensure that the sample of participants for each experiment quiz was unique, participants were not allowed to take part in an experiment more than once.

\textit{Procedure and design:} We arbitrarily selected 25 news articles that appeared in \href{newser.com}{newser.com} in early 2020. We used the article titles and subtitles to produce outputs from the 125M, 350M, 760M, 1.3B, 2.7B, 6.7B, 13.0B, and 200B (GPT-3) parameter language models. Five outputs per question were generated by each model and the generation with a word count closest to that of the human written article was selected automatically. This was to minimize the effect that completion length might have on participants’ judgments.  The same output procedure for each model with the exception of the removal of the intentionally bad control model, as described in the main text. 

In each experiment, half of the participants were randomly assigned to quiz A and half were randomly assigned to quiz B. Each quiz consisted of 25 articles: half (12-13) were human written and half (12-13) were model generated: the articles with human written completions in quiz A had model generated completions in quiz B and vice versa. The order of quiz question was shuffled for each participant. Participants could leave comments and were asked to indicate if they had seen the articles before. Participants were instructed not to look up the articles or their content during the quiz and at the end of the quiz were asked if they had looked anything up during the quiz.

\begin{table}
    \begin{center}
        \begin{tabular}{l c c c c c}
        \toprule
        Model & \shortstack{Participants\\Recruited}  & \shortstack{Participants\\Excluded} & \shortstack{Genders\\(m:f:other)} & \shortstack{Mean\\Age} & \shortstack{Average\\Word Count\\(human:model)} \\ 
        \midrule
        Control  & 76 & 7 & 32:37:0 & 39 & 216:216 \\ 
        GPT-3 Small  & 80 & 7  & 41:31:1 & 40 & 216:188 \\ 
        GPT-3 Medium  & 80 & 7  & 46:28:2  & 39 & 216:202 \\ 
        GPT-3 Large  & 81 & 24  & 46:28:2  & 37 & 216:200 \\ 
        GPT-3 XL  & 79 & 14  & 32:32:1  & 38 & 216:199 \\ 
        GPT-3 2.7B  & 80 & 11  & 36:33:0  & 40 & 216:202 \\ 
        GPT-3 6.7B  & 76 & 5  & 46:28:2  & 37 & 216:195 \\ 
        GPT-3 13.0B  & 81 & 13  & 46:28:2  & 37 & 216:209 \\ 
        GPT-3 175B & 80 & 9  & 42:29:0 &  37 & 216:216 \\ 
        \bottomrule
        \end{tabular}
    \end{center}
    \caption{Participant details and  article lengths for each experiment to evaluate human detection of $\sim200$ word model generated news articles. Participants were excluded due to internet check fails.}
    \label{table:study}
\end{table}

\textit{Statistical Tests:} To compare means on the different runs, we performed a two-sample t-test for independent groups for each model against the control. This was implemented in Python using the \verb|scipy.stats.ttest_ind| function. When plotting a regression line in the graph of average participant accuracy vs model size, we fit a power law of the form $ax^{-b}$. The 95\% confidence intervals were estimated from the t-distribution of the sample mean.

\begin{figure}
\centering
\includegraphics[width=0.7\linewidth]{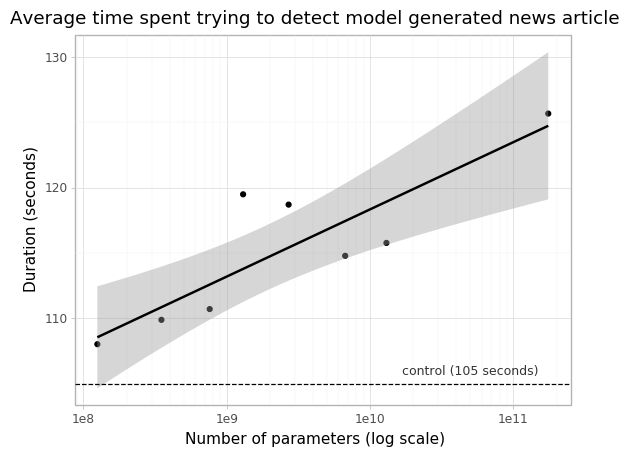}
\caption{Participants spend more time trying to identify whether each news article is machine generated as model size increases.  Duration on the control model is indicated with the dashed line. Line of best fit is a linear model on a log scale with 95\% confidence intervals.}
\label{graph:newsduration}
\end{figure}

\textit{Duration statistics}: In the main text, we discussed the finding that the ability of human participants to distinguish model and human generated news articles decreases as our models become larger. We have also found that the average time spent for a given set of questions increases as the model size increases, as shown in Figure \ref{graph:newsduration}. Lower accuracy scores despite increased time investment from participants supports the finding that larger models generate harder-to-distinguish news articles.

\textit{Preliminary investigation of $\sim500$ word articles:} We recruited 160 unique US-based participants to take part in 2 experiments through Positly (details are given in Table \ref{table:study_long}). We randomly selected 12 Reuters world news articles from late 2019 and created a context for GPT-3 175B that consisted of a single Reuters article not in this set of 12. We then used the article titles and Reuters locations to generate completions from GPT-3 175B and the 160M control model from the previous experiments. These were used to create two 12-question quizzes per model, each consisting of half human written and half model generated articles. Comprehension questions were added and articles were shown to participants in 3 stages at 30 second intervals to encourage closer reading. Participants were paid \$12 for this task. Model generation selection methods, exclusion criteria, and statistical tests mirror those of the previous experiments.

\begin{table}
    \begin{center}
        \begin{tabular}{l c c c c c}
        \toprule
        Model & \shortstack{Participants\\Recruited}  & \shortstack{Participants\\Excluded} & \shortstack{Genders\\(m:f:other)} & \shortstack{Mean\\Age} & \shortstack{Average\\Word Count\\(human:model)} \\ 
        \midrule
        Control  & 79 & 17 & 32:37:0 & 39 & 569:464 \\ 
        GPT-3 175B & 81 & 19  & 32:30:0 &  40 & 569:498 \\ 
        \bottomrule
        \end{tabular}
    \end{center}
    \caption{Participant details and  article lengths for the experiments investigating  human detection of $\sim500$ word model generated news articles. Participants were excluded due to internet check fails.}
    \label{table:study_long}
\end{table}
\section{Additional Samples from GPT-3}
\label{appendix:additional_samples}
GPT-3 adapts well to many tasks other than the ones explored in the main body of the paper.  As an example, in Figure \ref{sample:FakeStevensPoem}, we show four uncurated samples from a prompt suggesting that the model write a poem, with a given title, in the style of Wallace Stevens.  We first experimented with a few prompts, then generated four  samples with no additional editing or selection (sampling at temperature $1$ using nucleus sampling \cite{DBLP:journals/corr/abs-1904-09751} with $P=0.9$). Completions  were truncated when the model began to write a new title and author heading, or broke into prose commentary.

\begin{figure} { \tt \scriptsize \begin{center}\begin{tabular}{l l} \toprule Context $\to$ & The City \\
& BY C. P. CAVAFY  \\ 
& TRANSLATED BY EDMUND KEELEY \\
& [Poem text omitted] \\
 \\ \\
& SOME TREES \\
& John Ashbery \\
& [Poem text omitted] \\
 \\ \\ 
& Shadows on the Way \\
& Wallace Stevens \\
\midrule  
 & \\
 -------- Generated Poem 1  -------- &  -------- Generated Poem 3  -------- \\
 & \\
I must have shadows on the way & The sun was all we had. Now, in the shade\\
If I am to walk I must have & All is changed. The mind must dwell on those\\
Each step taken slowly and alone & White fields, that to its eyes were always old;\\
To have it ready made & Those ancient gleams, convoluting\\
 & The way they lay among the huge roots,\\
And I must think in lines of grey & The great dark tomes of reverie,\\
To have dim thoughts to be my guide & The plumed labyrinth of the sea.\\
Must look on blue and green & \\
And never let my eye forget & When our own breath gonne faint and fade\\
That color is my friend & Up the long stairway of the dead\\
And purple must surround me too & We shall remember well\\
 & The blindness of the light, the whiteness\\
The yellow of the sun is no more & Of that white land. The footsteps, and\\
Intrusive than the bluish snow & The grasses in that light, and in the shade\\
That falls on all of us. I must have & The menacing life.\\
Grey thoughts and blue thoughts walk with me & \\
If I am to go away at all. & We played, too, once, in the warmest rooms.\\
 & To one content, there is one content\\
 -------- Generated Poem 2  --------  & For all mankind. The forest glades\\
 & Are all the more for being in fable,\\
There is a way in the world. & And wood and town in legend shadowed-\\
That way a man may live in a small town & Once more to return. The one content\\
His lifetime and see no man come, & Has in it every happiness, more brave\\
Never the stir of men about a market, & For being purposeless; and so\\
Never the bobbing of covered carts, & Those island gardens, deep in the mind,\\
Never the creak passed on the stair & Make of remoteness familiarity\\
Closing some other life; an old man & And of the unspoken name a friend.\\
Can prosper in that town, can prosper & \\
When his store is seen to close, his wife &  -------- Generated Poem 4  --------  \\
Fades in his house, and the world shouts in vain & \\
That no one comes to hear him & Nobody will come to this place. It is a road that leads nowhere.\\
Where his small bell tolls. & The solitude is deep. The mountains are high.\\
In that small town & But they are desolate, and they turn the traveler’s face\\
Everything is known. & Towards the North. All the sounds of the world are far away.\\
The ways of the young men are known, & When the wind rises above the trees,\\
And the ways of the young women & The boughs bow to the ground.\\
As they walk to the Spring; & Even the birds that inhabit the tangle of weeds\\
The rumors are all known, & That is the roadside cover, are silent. One listens,\\
The quarrels, the hurt, the envy & But hears no roar of the forest. One is alone.\\
Of the men and the women. & One will be taken.\\
Till a day comes & One will be taken.\\
That way out of the world & There is no utterance, there is no conversation,\\
That leads to a wood. & But one is uneasy all the same....\\
Young, the young man & There is a thin blue mist,\\
Walks, a lover now, & A darkness rising like smoke,\\
Knows the streets as they are and is of them. & And within that darkness\\
 & A possession of the heart.\\
Comes to the wood. & One will be taken.... It was here, and it will be here again-\\
 & Here, under this sky empty and full of light.\\
There, in the wood, among trees, & \\
He sees shadows on the way, hears voices, & \\
hears the wind and the rustling of leaves; & \\
Through an open glade & \\
He sees a shape and the shape hears: & \\
It waits as he waits, & \\
Waits as the shadows wait, & \\
As the voices wait; & \\
Shadows on the way, voices in the wind. & \\
 & \\
\\ \bottomrule \end{tabular}\end{center} }  
\caption{Four uncurated completions from a context suggesting the model compose a poem in the style of Wallace Stevens with the title `Shadows on the Way'.}  
 \label{sample:FakeStevensPoem}  
 \end{figure}
\clearpage
\newpage
\section{Details of Task Phrasing and Specifications}
\label{appendix:task_phrasing}
The following figures illustrate the formatting and phrasing of all the tasks included in the paper. All data comes from the ground truth datasets in this section, and no samples from GPT-3 are included here.

\begin{figure}[!h] { \tt \footnotesize \begin{tabularx}{\linewidth}{r X} \toprule Context $\to$ & Article: \\
 & Informal conversation is an important part of any business relationship.Before you start a discussion,however,make sure you understand which topics are suitable and which are considered taboo in a particular culture. Latin Americans enjoy sharing information about their local history, art and customs.You may expect questions about your family,and be sure to show pictures of your children.You may feel free to ask similar questions of your Latin American friends.The French think of conversation as an art form,and they enjoy the value of lively discussions as well as disagreements. For them,arguments can be interesting and they can cover pretty much or any topic ---- as long as they occur in are respectful and intelligent manner. \\ 
 & In the United States,business people like to discuss a wide range of topics,including opinions about work,family,hobbies,and politics. In Japan,China,and Korea,however,people are much more private.They do not share much about their thoughts,feelings,or emotions because they feel that doing so might take away from the harmonious business relationship they're trying to build.Middle Easterners are also private about their personal lives and family matters.It is considered rude,for example,to ask a businessman from Saudi Arabia about his wife or children. \\ 
 & As a general rule,it's best not to talk about politics or religion with your business friends.This can get you into trouble,even in the United States,where people hold different religious views.In addition,discussing one's salary is usually considered unsuitable.Sports is typically a friendly subject in most parts of the world,although be careful not to criticize national sport.Instead,be friendly and praise your host's team. \\ 
 &  \\ 
 & Q: What shouldn't you do when talking about sports with colleagues from another country? \\ 
 &  \\ 
 & A: Criticizing the sports of your colleagues' country. \\ 
 &  \\ 
 & Q: Which is typically a friendly topic in most places according to the author? \\ 
 &  \\ 
 & A: Sports. \\ 
 &  \\ 
 & Q: Why are people from Asia more private in their conversation with others? \\ 
 &  \\ 
 & A: They don't want to have their good relationship with others harmed by informal conversation. \\ 
 &  \\ 
 & Q: The author considers politics and religion   \_  . \\ 
 &  \\ 
 & A: \\ 
 \midrule Correct Answer $\to$ & taboo \\ 
 Incorrect Answer $\to$ & cheerful topics \\ 
 Incorrect Answer $\to$ & rude topics \\ 
 Incorrect Answer $\to$ & topics that can never be talked about\\ \bottomrule \end{tabularx} }  \caption{Formatted dataset example for RACE-h. When predicting, we normalize by the unconditional probability of each answer as described in \ref{section:Approach}.}  \label{eval:RACE-h}  \end{figure}
\begin{figure} { \tt \footnotesize \begin{tabularx}{\linewidth}{r X} \toprule Context $\to$ & anli 2: anli 2: The Gold Coast Hotel \& Casino is a hotel and casino located in Paradise, Nevada. This locals' casino is owned and operated by Boyd Gaming. The Gold Coast is located one mile ($\sim1.6\mathrm{km}$) west of the Las Vegas Strip on West Flamingo Road. It is located across the street from the Palms Casino Resort and the Rio All Suite Hotel and Casino. \\
 & Question: The Gold Coast is a budget-friendly casino. True, False, or Neither? \\ 
 \midrule Correct Answer $\to$ & Neither \\ 
 Incorrect Answer $\to$ & True \\ 
 Incorrect Answer $\to$ & False\\ \bottomrule \end{tabularx} }  \caption{Formatted dataset example for ANLI R2}  \label{eval:ANLI R2}  \end{figure}
\begin{figure} { \tt \footnotesize \begin{tabularx}{\linewidth}{r X} \toprule Context $\to$ & Article: \\ 
 & Mrs. Smith is an unusual teacher. Once she told each student to bring along a few potatoes in plastic bag. On each potato the students had to write a name of a person that they hated And the next day, every child brought some potatoes. Some had two potatoes;some three;some up to five. \\ 
 & Mrs. Smith then told the children to carry the bags everywhere they went, even to the toilet, for two weeks. As day after day passed, the children started to complain about the awful smell of the rotten potatoes. \\ 
 & Those children who brought five potatoes began to feel the weight trouble of the bags. After two weeks, the children were happy to hear that the game was finally ended. Mrs. Smith asked,"How did you feel while carrying the potatoes for two weeks?" The children started complaining about the trouble loudly. \\ 
 & Then Mrs. Smith told them why she asked them to play the game. She said,"This is exactly the situation when you carry your hatred for somebody inside your heart. The terrible smell of the hatred will pollute your heart and you will carry something unnecessary with you all the time. If you cannot stand the smell of the rotten potatoes for just two weeks, can you imagine how heavy it would be to have the hatred in your heart for your lifetime? So throw away any hatred from your heart, and you'll be really happy." \\ 
 &  \\ 
 & Q: Which of the following is True according to the passage? \\ 
 &  \\ 
 & A: If a kid hated four people,he or she had to carry four potatoes. \\ 
 &  \\ 
 & Q: We can learn from the passage that we should  \_  . \\ 
 &  \\ 
 & A: throw away the hatred inside \\ 
 &  \\ 
 & Q: The children complained about  \_  besides the weight trouble. \\ 
 &  \\ 
 & A: the smell \\ 
 &  \\ 
 & Q: Mrs.Smith asked her students to write    \_  on the potatoes. \\ 
 &  \\ 
 & A: \\ 
 \midrule Correct Answer $\to$ & names \\ 
 Incorrect Answer $\to$ & numbers \\ 
 Incorrect Answer $\to$ & time \\ 
 Incorrect Answer $\to$ & places\\ \bottomrule \end{tabularx} }  \caption{Formatted dataset example for RACE-m. When predicting, we normalize by the unconditional probability of each answer as described in \ref{section:Approach}.}  \label{eval:RACE-m}  \end{figure}
\begin{figure} { \tt \footnotesize \begin{tabularx}{\linewidth}{r X} \toprule Context $\to$ & How to apply sealant to wood. \\ 
 \midrule Correct Answer $\to$ & Using a brush, brush on sealant onto wood until it is fully saturated with the sealant. \\ 
 Incorrect Answer $\to$ & Using a brush, drip on sealant onto wood until it is fully saturated with the sealant.\\ \bottomrule \end{tabularx} }  \caption{Formatted dataset example for PIQA}  \label{eval:PIQA}  \end{figure}
\begin{figure} { \tt \footnotesize \begin{tabularx}{\linewidth}{r X} \toprule Context $\to$ & My body cast a shadow over the grass because \\ 
 \midrule Correct Answer $\to$ & the sun was rising. \\ 
 Incorrect Answer $\to$ & the grass was cut.\\ \bottomrule \end{tabularx} }  \caption{Formatted dataset example for COPA}  \label{eval:COPA}  \end{figure}
\begin{figure} { \tt \footnotesize \begin{tabularx}{\linewidth}{r X} 
\toprule 
Context $\to$ & (CNN) Yuval Rabin, whose father, Yitzhak Rabin, was assassinated while serving as Prime Minister of Israel, criticized Donald Trump for appealing to "Second Amendment people" in a speech and warned that the words that politicians use can incite violence and undermine democracy. "Trump's words are an incitement to the type of political violence that touched me personally," Rabin wrote in USAToday. He said that Trump's appeal to "Second Amendment people" to stop Hillary Clinton -- comments that were criticized as a call for violence against Clinton, something Trump denied -- "were a new level of ugliness in an ugly campaign season." \\
& \\
&   - The son of a former Israeli Prime Minister who was assassinated wrote an op ed about the consequence of violent political rhetoric. \\
&   - Warns of "parallels" between Israel of the 1990s and the U.S. today. \\
\midrule 
 Correct Answer $\to$ & - Referencing his father, who was shot and killed by an extremist amid political tension in Israel in 1995, Rabin condemned Donald Trump's aggressive rhetoric. \\ 
 
 Correct Answer $\to$ & - Referencing his father, who was shot and killed by an extremist amid political tension in Israel in 1995, Rabin condemned Trump's aggressive rhetoric. \\ 
 
 Incorrect Answer $\to$ & - Referencing his father, who was shot and killed by an extremist amid political tension in Israel in 1995, Rabin condemned Hillary Clinton's aggressive rhetoric. \\
  Incorrect Answer $\to$ & - Referencing his father, who was shot and killed by an extremist amid political tension in Israel in 1995, Rabin condemned U.S.'s aggressive rhetoric. \\
 Incorrect Answer $\to$ & - Referencing his father, who was shot and killed by an extremist amid political tension in Israel in 1995, Rabin condemned Yitzhak Rabin's aggressive rhetoric. \\ 
 \bottomrule \end{tabularx} }   

 \caption{Formatted dataset example for ReCoRD. We consider the context above to be a single "problem" because this is how the task is presented in the ReCoRD dataset and scored in the ReCoRD evaluation script.}

\label{eval:ReCoRD}  \end{figure}
\begin{figure} { \tt \footnotesize \begin{tabularx}{\linewidth}{r X} \toprule Context $\to$ & anli 1: anli 1: Fulton James MacGregor MSP is a Scottish politician who is a Scottish National Party (SNP) Member of Scottish Parliament for the constituency of Coatbridge and Chryston. MacGregor is currently Parliamentary Liaison Officer to Shona Robison, Cabinet Secretary for Health \& Sport. He also serves on the Justice and Education \& Skills committees in the Scottish Parliament. \\ 
 & Question: Fulton James MacGregor is a Scottish politican who is a Liaison officer to Shona Robison who he swears is his best friend. True, False, or Neither? \\ 
 \midrule Correct Answer $\to$ & Neither \\ 
 Incorrect Answer $\to$ & True \\ 
 Incorrect Answer $\to$ & False\\ \bottomrule \end{tabularx} }  \caption{Formatted dataset example for ANLI R1}  \label{eval:ANLI R1}  \end{figure}
\begin{figure} { \tt \footnotesize \begin{tabularx}{\linewidth}{r X} \toprule Context $\to$ & Organisms require energy in order to do what? \\ 
 \midrule Correct Answer $\to$ & mature and develop. \\ 
 Incorrect Answer $\to$ & rest soundly. \\ 
 Incorrect Answer $\to$ & absorb light. \\ 
 Incorrect Answer $\to$ & take in nutrients.\\ \bottomrule \end{tabularx} }  \caption{Formatted dataset example for OpenBookQA. When predicting, we normalize by the unconditional probability of each answer as described in \ref{section:Approach}.}  \label{eval:OpenBookQA}  \end{figure}
\begin{figure} { \tt \footnotesize \begin{tabularx}{\linewidth}{r X} \toprule Context $\to$ & Making a cake: Several cake pops are shown on a display. A woman and girl are shown making the cake pops in a kitchen. They \\ 
 \midrule Correct Answer $\to$ & bake them, then frost and decorate. \\ 
 Incorrect Answer $\to$ & taste them as they place them on plates. \\ 
 Incorrect Answer $\to$ & put the frosting on the cake as they pan it. \\ 
 Incorrect Answer $\to$ & come out and begin decorating the cake as well.\\ \bottomrule \end{tabularx} }  \caption{Formatted dataset example for HellaSwag}  \label{eval:HellaSwag}  \end{figure}
\begin{figure} { \tt \footnotesize \begin{tabularx}{\linewidth}{r X} \toprule Context $\to$ & anli 3: anli 3: We shut the loophole which has American workers actually subsidizing the loss of their own job. They just passed an expansion of that loophole in the last few days: \$43 billion of giveaways, including favors to the oil and gas industry and the people importing ceiling fans from China. \\ 
 & Question: The loophole is now gone True, False, or Neither? \\ 
 \midrule Correct Answer $\to$ & False \\ 
 Incorrect Answer $\to$ & True \\ 
 Incorrect Answer $\to$ & Neither\\ \bottomrule \end{tabularx} }  \caption{Formatted dataset example for ANLI R3}  \label{eval:ANLI R3}  \end{figure}
\begin{figure} { \tt \footnotesize \begin{tabularx}{\linewidth}{r X} \toprule Context $\to$ & Question: George wants to warm his hands quickly by rubbing them. Which skin surface will produce the most heat? \\ 
 & Answer: \\ 
 \midrule Correct Answer $\to$ & dry palms \\ 
 Incorrect Answer $\to$ & wet palms \\ 
 Incorrect Answer $\to$ & palms covered with oil \\ 
 Incorrect Answer $\to$ & palms covered with lotion\\ \bottomrule \end{tabularx} }  \caption{Formatted dataset example for ARC (Challenge). When predicting, we normalize by the unconditional probability of each answer as described in \ref{section:Approach}.}  \label{eval:ARC (Challenge)}  \end{figure}
\begin{figure} { \tt \footnotesize \begin{tabularx}{\linewidth}{r X} \toprule Context $\to$ & lull is to trust as \\ 
 \midrule Correct Answer $\to$ & cajole is to compliance \\ 
 Incorrect Answer $\to$ & balk is to fortitude \\ 
 Incorrect Answer $\to$ & betray is to loyalty \\ 
 Incorrect Answer $\to$ & hinder is to destination \\ 
 Incorrect Answer $\to$ & soothe is to passion\\ \bottomrule \end{tabularx} }  \caption{Formatted dataset example for SAT Analogies}  \label{eval:SAT Analogies}  \end{figure}
\begin{figure} { \tt \footnotesize \begin{tabularx}{\linewidth}{r X} \toprule
Correct Context $\to$ & Grace was happy to trade me her sweater for my jacket. She thinks the sweater\\
Incorrect Context $\to$ & Grace was happy to trade me her sweater for my jacket. She thinks the jacket\\
 \midrule
Target Completion $\to$ & looks dowdy on her. \\
 \bottomrule \end{tabularx} }
\caption{Formatted dataset example for Winograd.  The `partial' evaluation method we use compares the probability of the completion given a correct and incorrect context.}  \label{eval:Winograd}  \end{figure}
\begin{figure} { \tt \footnotesize \begin{tabularx}{\linewidth}{r X}
\toprule
Correct Context $\to$ & Johnny likes fruits more than vegetables in his new keto diet because the fruits \\
Incorrect Context $\to$ & Johnny likes fruits more than vegetables in his new keto diet because the vegetables \\
\midrule
Target Completion $\to$ & are saccharine. \\
\bottomrule \end{tabularx} }  \caption{Formatted dataset example for Winogrande. The `partial' evaluation method we use compares the probability of the completion given a correct and incorrect context.}  \label{eval:Winogrande}  \end{figure}
\begin{figure} { \tt \footnotesize \begin{tabularx}{\linewidth}{r X} \toprule Context $\to$ & READING COMPREHENSION ANSWER KEY \\ 
 & While this process moved along, diplomacy continued its rounds. Direct pressure on the Taliban had proved unsuccessful. As one NSC staff note put it, "Under the Taliban, Afghanistan is not so much a state sponsor of terrorism as it is a state sponsored by terrorists." In early 2000, the United States began a high-level effort to persuade Pakistan to use its influence over the Taliban. In January 2000, Assistant Secretary of State Karl Inderfurth and the State Department's counterterrorism coordinator, Michael Sheehan, met with General Musharraf in Islamabad, dangling before him the possibility of a presidential visit in March as a reward for Pakistani cooperation. Such a visit was coveted by Musharraf, partly as a sign of his government's legitimacy. He told the two envoys that he would meet with Mullah Omar and press him on Bin Laden. They left, however, reporting to Washington that Pakistan was unlikely in fact to do anything," given what it sees as the benefits of Taliban control of Afghanistan." President Clinton was scheduled to travel to India. The State Department felt that he should not visit India without also visiting Pakistan. The Secret Service and the CIA, however, warned in the strongest terms that visiting Pakistan would risk the President's life. Counterterrorism officials also argued that Pakistan had not done enough to merit a presidential visit. But President Clinton insisted on including Pakistan in the itinerary for his trip to South Asia. His one-day stopover on March 25, 2000, was the first time a U.S. president had been there since 1969. At his meeting with Musharraf and others, President Clinton concentrated on tensions between Pakistan and India and the dangers of nuclear proliferation, but also discussed Bin Laden. President Clinton told us that when he pulled Musharraf aside for a brief, one-on-one meeting, he pleaded with the general for help regarding Bin Laden." I offered him the moon when I went to see him, in terms of better relations with the United States, if he'd help us get Bin Laden and deal with another issue or two." The U.S. effort continued. \\ 
 & \\
 & Who did The State Department feel should visit both India and Pakistan? \\ 
 \midrule Correct Answer $\to$ & - [False] Bin Laden \\ 
 Incorrect Answer $\to$ & - [True] Bin Laden\\ \bottomrule \end{tabularx} }  \caption{Formatted dataset example for MultiRC. There are three levels within MultiRC: (1) the passage, (2) the questions, and (3) the answers. During evaluation, accuracy is determined at the per-question level, with a question being considered correct if and only if all the answers within the question are labeled correctly. For this reason, we use $K$ to refer to the number of \textbf{questions} shown within the context.   }  \label{eval:MultiRC}  \end{figure}
\begin{figure} { \tt \footnotesize \begin{tabularx}{\linewidth}{r X} \toprule Context $\to$ & Question: Which factor will most likely cause a person to develop a fever? \\ 
 & Answer: \\ 
 \midrule Correct Answer $\to$ & a bacterial population in the bloodstream \\ 
 Incorrect Answer $\to$ & a leg muscle relaxing after exercise \\ 
 Incorrect Answer $\to$ & several viral particles on the skin \\ 
 Incorrect Answer $\to$ & carbohydrates being digested in the stomach\\ \bottomrule \end{tabularx} }

 \caption{Formatted dataset example for ARC (Easy). When predicting, we normalize by the unconditional probability of each answer as described in \ref{section:Approach}.}  \label{eval:ARC (Easy)}  \end{figure}
\clearpage
\begin{figure} { \tt \footnotesize \begin{tabularx}{\linewidth}{r X} \toprule Context $\to$ & Bob went to the gas station to fill up his car. His tank was completely empty and so was his wallet. The cashier offered to pay for his gas if he came back later to pay. Bob felt grateful as he drove home. \\
 \midrule Correct Answer $\to$ & Bob believed that there were good people in the world. \\ 
 Incorrect Answer $\to$ & Bob contemplated how unfriendly the world was.\\ \bottomrule \end{tabularx} }  \caption{Formatted dataset example for StoryCloze}  \label{eval:StoryCloze}  \end{figure}
\begin{figure} { \tt \footnotesize \begin{tabularx}{\linewidth}{r X} \toprule Context $\to$ & Helsinki is the capital and largest city of Finland. It is in the region of Uusimaa, in southern Finland, on the shore of the Gulf of Finland. Helsinki has a population of , an urban population of , and a metropolitan population of over 1.4 million, making it the most populous municipality and urban area in Finland. Helsinki is some north of Tallinn, Estonia, east of Stockholm, Sweden, and west of Saint Petersburg, Russia. Helsinki has close historical connections with these three cities. \\
 &  \\ 
 & The Helsinki metropolitan area includes the urban core of Helsinki, Espoo, Vantaa, Kauniainen, and surrounding commuter towns. It is the world's northernmost metro area of over one million people, and the city is the northernmost capital of an EU member state. The Helsinki metropolitan area is the third largest metropolitan area in the Nordic countries after Stockholm and Copenhagen, and the City of Helsinki is the third largest after Stockholm and Oslo. Helsinki is Finland's major political, educational, financial, cultural, and research center as well as one of northern Europe's major cities. Approximately 75\% of foreign companies that operate in Finland have settled in the Helsinki region. The nearby municipality of Vantaa is the location of Helsinki Airport, with frequent service to various destinations in Europe and Asia. \\ 
 &  \\ 
 & Q: what is the most populous municipality in Finland? \\ 
 &  \\ 
 & A: Helsinki \\ 
 &  \\ 
 & Q: how many people live there? \\ 
 &  \\ 
 & A: 1.4 million in the metropolitan area \\ 
 &  \\ 
 & Q: what percent of the foreign companies that operate in Finland are in Helsinki? \\ 
 &  \\ 
 & A: 75\% \\ 
 &  \\ 
 & Q: what towns are a part of the metropolitan area? \\ 
 &  \\ 
 & A: \\ 
 \midrule Target Completion $\to$ & Helsinki, Espoo, Vantaa, Kauniainen, and surrounding commuter towns \\
\bottomrule \end{tabularx} }  \caption{Formatted dataset example for CoQA}  \label{eval:CoQA}  \end{figure}

\begin{figure} { \tt \footnotesize \begin{tabularx}{\linewidth}{r X} \toprule Context $\to$ & Please unscramble the letters into a word, and write that word: \\
& asinoc = \\ 
 \midrule Target Completion $\to$ & casino\\ \bottomrule \end{tabularx} }  \caption{Formatted dataset example for Cycled Letters}
 \label{eval:CL}  
 \end{figure}
\begin{figure} { \tt \footnotesize \begin{tabularx}{\linewidth}{r X} \toprule Context $\to$ & Passage: Saint Jean de Br\'ebeuf  was a French Jesuit missionary who travelled to New France  in 1625. There he worked primarily with the Huron for the rest of his life, except for a few years in France from 1629 to 1633. He learned their language and culture, writing extensively about each to aid other missionaries. In 1649, Br\'ebeuf and another missionary were captured when an Iroquois raid took over a Huron village . Together with Huron captives, the missionaries were ritually tortured and killed on March 16, 1649. Br\'ebeuf was beatified in 1925 and among eight Jesuit missionaries canonized as saints in the Roman Catholic Church in 1930. \\ 
 & Question: How many years did Saint Jean de Br\'ebeuf stay in New France before he went back to France for a few years? \\ 
 & Answer: \\ 
 \midrule Target Completion $\to$ & 4 \\
\bottomrule \end{tabularx} }  \caption{Formatted dataset example for DROP}  \label{eval:DROP}  \end{figure}
\begin{figure} { \tt \footnotesize \begin{tabularx}{\linewidth}{r X} \toprule Context $\to$ & Fill in blank: \\ 
 &  \\ 
 & She held the torch in front of her. \\ 
 &  \\ 
 & She caught her breath. \\ 
 &  \\ 
 & "Chris? There's a step." \\ 
 &  \\ 
 & "What?" \\ 
 &  \\ 
 & "A step. Cut in the rock. About fifty feet ahead." She moved faster. They both moved faster. "In fact," she said, raising the torch higher, "there's more than a \_\_\_\_. -\ensuremath{>} \\ 
 \midrule Target Completion $\to$ & step\\ \bottomrule \end{tabularx} }  \caption{Formatted dataset example for LAMBADA}  \label{eval:LAMBADA}  \end{figure}
\begin{figure} { \tt \footnotesize \begin{tabularx}{\linewidth}{r X} \toprule Context $\to$ & Please unscramble the letters into a word, and write that word: \\
& skicts = \\ 
 \midrule Target Completion $\to$ & sticks\\ \bottomrule \end{tabularx} }  \caption{Formatted dataset example for Anagrams 1 (A1)}  \label{eval:A1}  \end{figure}
\begin{figure} { \tt \footnotesize \begin{tabularx}{\linewidth}{r X} \toprule Context $\to$ & Please unscramble the letters into a word, and write that word: \\
& volwskagen = \\ 
 \midrule Target Completion $\to$ & volkswagen\\ \bottomrule \end{tabularx} }  \caption{Formatted dataset example for Anagrams 2}  \label{eval:A2}  \end{figure}
\begin{figure} { \tt \footnotesize \begin{tabularx}{\linewidth}{r X} \toprule Context $\to$ & Q: Who played tess on touched by an angel? \\ 
 &  \\ 
 & A: \\ 
 \midrule Target Completion $\to$ & Delloreese Patricia Early (July 6, 1931 {\textendash} November 19, 2017), known professionally as Della Reese \\
 \bottomrule \end{tabularx} }  \caption{Formatted dataset example for Natural Questions}  \label{eval:NQs}  \end{figure}

\begin{figure} { \tt \footnotesize \begin{tabularx}{\linewidth}{r X} \toprule Context $\to$ & TITLE: William Perry (American football) - Professional career \\ 
 & PARAGRAPH: In 1985, he was selected in the first round of the 1985 NFL Draft by the Chicago Bears; he had been hand-picked by coach Mike Ditka. However, defensive coordinator Buddy Ryan, who had a highly acrimonious relationship with Ditka, called Perry a "wasted draft-pick". Perry soon became a pawn in the political power struggle between Ditka and Ryan.  Perry's "Refrigerator" nickname followed him into the NFL and he quickly became a favorite of the Chicago Bears fans. Teammates called him "Biscuit," as in "one biscuit shy of 350 pounds."  While Ryan refused to play Perry, Ditka decided to use Perry as a fullback when the team was near the opponents' goal line or in fourth and short situations, either as a ball carrier or a lead blocker for star running back Walter Payton. Ditka stated the inspiration for using Perry as a fullback came to him during five-yard sprint exercises. During his rookie season, Perry rushed for two touchdowns and caught a pass for one. Perry even had the opportunity to run the ball during Super Bowl XX, as a nod to his popularity and contributions to the team's success. The first time he got the ball, he was tackled for a one-yard loss while attempting to throw his first NFL pass on a halfback option play. The second time he got the ball, he scored a touchdown (running over Patriots linebacker Larry McGrew in the process). About halfway through his rookie season, Ryan finally began to play Perry, who soon proved that he was a capable defensive lineman.  His Super Bowl ring size is the largest of any professional football player in the history of the event. His ring size is 25, while the ring size for the average adult male is between 10 and 12.  Perry went on to play for ten years in the NFL, retiring after the 1994 season. In his ten years as a pro, he regularly struggled with his weight, which hampered his performance at times. He played in 138 games, recording 29.5 sacks and five fumble recoveries, which he returned for a total of 71 yards. In his offensive career he ran five yards for two touchdowns, and had one reception for another touchdown. Perry later attempted a comeback, playing an unremarkable 1996 season with the London Monarchs of the World League of American Football (later NFL Europa). \\ 
 &  \\ 
 & Q: what team did he play for? \\ 
 &  \\ 
 & A: \\ 
 \midrule Target Completion $\to$ & the Chicago Bears \\
\bottomrule \end{tabularx} }  \caption{Formatted dataset example for QuAC}  \label{eval:Quac}  \end{figure}
\begin{figure} { \tt \footnotesize \begin{tabularx}{\linewidth}{r X} \toprule Context $\to$ & Please unscramble the letters into a word, and write that word: \\
& r e!c.i p r o.c a/l = \\ 
 \midrule Target Completion $\to$ & reciprocal\\ \bottomrule \end{tabularx} }  \caption{Formatted dataset example for Symbol Insertion}  \label{eval:RI}  \end{figure}
\begin{figure} { \tt \footnotesize \begin{tabularx}{\linewidth}{r X} \toprule Context $\to$ & Please unscramble the letters into a word, and write that word:\\
& taefed = \\ 
 \midrule Target Completion $\to$ & defeat\\ \bottomrule \end{tabularx} }  \caption{Formatted dataset example for Reversed Words}  \label{eval:RW}  \end{figure}
\begin{figure} { \tt \footnotesize \begin{tabularx}{\linewidth}{r X} \toprule Context $\to$ & Title: The\_Blitz \\ 
 &  \\ 
 & Background: From the German point of view, March 1941 saw an improvement. The Luftwaffe flew 4,000 sorties that month, including 12 major and three heavy attacks. The electronic war intensified but the Luftwaffe flew major inland missions only on moonlit nights. Ports were easier to find and made better targets. To confuse the British, radio silence was observed until the bombs fell. X- and Y-Ger\"at beams were placed over false targets and switched only at the last minute. Rapid frequency changes were introduced for X-Ger\"at, whose wider band of frequencies and greater tactical flexibility ensured it remained effective at a time when British selective jamming was degrading the effectiveness of Y-Ger\"at. \\ 
 &  \\ 
 & Q: How many sorties were flown in March 1941? \\ 
 &  \\ 
 & A: 4,000 \\ 
 &  \\ 
 & Q: When did the Luftwaffe fly inland missions? \\ 
 &  \\ 
 & A: \\ 
 \midrule Target Completion $\to$ & only on moonlit nights \\
\bottomrule \end{tabularx} }  \caption{Formatted dataset example for SQuADv2}  \label{eval:SQuADv2}  \end{figure}
\begin{figure} { \tt \footnotesize \begin{tabularx}{\linewidth}{r X} \toprule Context $\to$ & Normal force -- In a simple case such as an object resting upon a table, the normal force on the object is equal but in opposite direction to the gravitational force applied on the object (or the weight of the object), that is, N = m g ({\textbackslash}displaystyle N=mg), where m is mass, and g is the gravitational field strength (about 9.81 m/s on Earth). The normal force here represents the force applied by the table against the object that prevents it from sinking through the table and requires that the table is sturdy enough to deliver this normal force without breaking. However, it is easy to assume that the normal force and weight are action-reaction force pairs (a common mistake). In this case, the normal force and weight need to be equal in magnitude to explain why there is no upward acceleration of the object. For example, a ball that bounces upwards accelerates upwards because the normal force acting on the ball is larger in magnitude than the weight of the ball. \\ 
 & question: is the normal force equal to the force of gravity? \\ 
 & answer: \\ 
 \midrule Target Completion $\to$ & yes\\ \bottomrule \end{tabularx} }  \caption{Formatted dataset example for BoolQ}  \label{eval:BoolQ}  \end{figure}
\begin{figure} { \tt \footnotesize \begin{tabularx}{\linewidth}{r X} \toprule Context $\to$ & The trend toward lower rents may seem surprising given that some communities in New York are bemoaning the loss of favorite local businesses to high rents. But, despite the recent softening, for many of these retailers there's still been too big a jump from the rental rates of the late 1970s, when their leases were signed. Certainly, the recent drop in prices doesn't mean Manhattan comes cheap. \\ 
 & question: Manhattan comes cheap. true, false, or neither? \\ 
 & answer: \\ 
 \midrule Target Completion $\to$ & false\\ \bottomrule \end{tabularx} }  \caption{Formatted dataset example for CB}  \label{eval:CB}  \end{figure}
\begin{figure} { \tt \footnotesize \begin{tabularx}{\linewidth}{r X} \toprule Context $\to$ & The bet, which won him dinner for four, was regarding the existence and mass of the top quark, an elementary particle discovered in 1995. \\ 
 & question: The Top Quark is the last of six flavors of quarks predicted by the standard model theory of particle physics. True or False? \\ 
 & answer: \\ 
 \midrule Target Completion $\to$ & False\\ \bottomrule \end{tabularx} }  \caption{Formatted dataset example for RTE}  \label{eval:RTE}  \end{figure}
\begin{figure} { \tt \footnotesize \begin{tabularx}{\linewidth}{r X} \toprule Context $\to$ & An outfitter provided everything needed for the safari. \\ 
 & Before his first walking holiday, he went to a specialist outfitter to buy some boots. \\ 
 & question: Is the word `outfitter' used in the same way in the two sentences above? \\ 
 & answer: \\ 
 \midrule Target Completion $\to$ & no\\ \bottomrule \end{tabularx} }  \caption{Formatted dataset example for WiC}  \label{eval:WiC}  \end{figure}
\begin{figure} { \tt \footnotesize \begin{tabularx}{\linewidth}{r X} \toprule Context $\to$ & Final Exam with Answer Key \\ 
 & Instructions: Please carefully read the following passages. For each passage, you must identify which noun the pronoun marked in *bold* refers to. \\ 
 & ===== \\ 
 & Passage: Mr. Moncrieff visited Chester's luxurious New York apartment, thinking that it belonged to his son Edward. The result was that Mr. Moncrieff has decided to cancel Edward's allowance on the ground that he no longer requires *his* financial support. \\ 
 & Question: In the passage above, what does the pronoun "*his*" refer to? \\ 
 & Answer: \\ 
 \midrule Target Completion $\to$ & mr. moncrieff\\ \bottomrule \end{tabularx} }  \caption{Formatted dataset example for WSC}  \label{eval:WSC}  \end{figure}
\begin{figure} { \tt \footnotesize \begin{tabularx}{\linewidth}{r X} \toprule Context $\to$ & Q: `Nude Descending A Staircase' is perhaps the most famous painting by which 20th century artist? \\ 
 &  \\ 
 & A: \\ 
 \midrule Target Completion $\to$ & MARCEL DUCHAMP \\
 Target Completion $\to$ & r mutt \\
 Target Completion $\to$ & duchamp \\
 Target Completion $\to$ & marcel duchamp \\
 Target Completion $\to$ & R.Mutt \\
 Target Completion $\to$ & Marcel duChamp \\
 Target Completion $\to$ & Henri-Robert-Marcel Duchamp \\
 Target Completion $\to$ & Marcel du Champ \\
 Target Completion $\to$ & henri robert marcel duchamp \\
 Target Completion $\to$ & Duchampian \\
 Target Completion $\to$ & Duchamp \\
 Target Completion $\to$ & duchampian \\
 Target Completion $\to$ & marcel du champ \\
 Target Completion $\to$ & Marcel Duchamp \\
 Target Completion $\to$ & MARCEL DUCHAMP\\ \bottomrule \end{tabularx} }  \caption{Formatted dataset example for TriviaQA. TriviaQA allows for multiple valid completions.}  \label{eval:TriviaQA}  \end{figure}
\begin{figure} { \tt \footnotesize \begin{tabularx}{\linewidth}{r X} \toprule Context $\to$ & Q: What school did burne hogarth establish? \\ 
 &  \\ 
 & A: \\ 
 \midrule Target Completion $\to$ & School of Visual Arts \\ \bottomrule \end{tabularx} }  \caption{Formatted dataset example for WebQA}  \label{eval:WebQA-no-dev}  \end{figure}
\begin{figure} { \tt \footnotesize \begin{tabularx}{\linewidth}{r X} \toprule Context $\to$ & Keinesfalls d\"urfen diese f\"ur den kommerziellen Gebrauch verwendet werden. = \\ 
 \midrule
Target Completion $\to$ & In no case may they be used for commercial purposes. \\
\bottomrule \end{tabularx} }  \caption{Formatted dataset example for De$\to$En.  This is the format for one- and few-shot learning, for this and other langauge tasks, the format for zero-shot learning is ``Q: What is the \{language\} translation of \{sentence\} A: \{translation\}." }  \label{eval:De_En}  \end{figure}
\begin{figure} { \tt \footnotesize \begin{tabularx}{\linewidth}{r X} \toprule Context $\to$ & In no case may they be used for commercial purposes. = \\ 
 \midrule Target Completion $\to$ & Keinesfalls d\"urfen diese f\"ur den kommerziellen Gebrauch verwendet werden. \\
\bottomrule \end{tabularx} }  \caption{Formatted dataset example for En$\to$De}  \label{eval:En_De}  \end{figure}
\begin{figure} { \tt \footnotesize \begin{tabularx}{\linewidth}{r X} \toprule Context $\to$ & Analysis of instar distributions of larval I. verticalis collected from a series of ponds also indicated that males were in more advanced instars than females. = \\ 
 \midrule Target Completion $\to$ & L'analyse de la distribution de fr\'equence des stades larvaires d'I. verticalis dans une s\'erie d'\'etangs a \'egalement d\'emontr\'e que les larves m\^ales \'etaient \`a des stades plus avanc\'es que les larves femelles. \\
\bottomrule \end{tabularx} }  \caption{Formatted dataset example for En$\to$Fr}  \label{eval:En_Fr}  \end{figure}
\begin{figure} { \tt \footnotesize \begin{tabularx}{\linewidth}{r X} \toprule Context $\to$ & L'analyse de la distribution de fr\'equence des stades larvaires d'I. verticalis dans une s\'erie d'\'etangs a \'egalement d\'emontr\'e que les larves m\^ales \'etaient \`a des stades plus avanc\'es que les larves femelles. = \\ 
 \midrule Target Completion $\to$ & Analysis of instar distributions of larval I. verticalis collected from a series of ponds also indicated that males were in more advanced instars than females. \\
\bottomrule \end{tabularx} }  \caption{Formatted dataset example for Fr$\to$En}  \label{eval:Fr_En}  \end{figure}
\begin{figure} { \tt \footnotesize \begin{tabularx}{\linewidth}{r X} \toprule Context $\to$ & The truth is that you want, at any price, and against the wishes of the peoples of Europe, to continue the negotiations for Turkey's accession to the European Union, despite Turkey's continuing refusal to recognise Cyprus and despite the fact that the democratic reforms are at a standstill. = \\ 
 \midrule Target Completion $\to$ & Adev\u{a}rul este c\u{a} v\u{a} dori\c{t}i, cu orice pre\c{t} \c{s}i {\^\i}mpotriva dorin\c{t}ei europenilor, s\u{a} continua\c{t}i negocierile de aderare a Turciei la Uniunea European\u{a}, {\^\i}n ciuda refuzului continuu al Turciei de a recunoa\c{s}te Ciprul \c{s}i {\^\i}n ciuda faptului c\u{a} reformele democratice au ajuns {\^\i}ntr-un punct mort. \\
\bottomrule \end{tabularx} }  \caption{Formatted dataset example for En$\to$Ro}  \label{eval:En_Ro}  \end{figure}
\begin{figure} { \tt \footnotesize \begin{tabularx}{\linewidth}{r X} \toprule Context $\to$ & Adev\u{a}rul este c\u{a} v\u{a} dori\c{t}i, cu orice pre\c{t} \c{s}i {\^\i}mpotriva dorin\c{t}ei europenilor, s\u{a} continua\c{t}i negocierile de aderare a Turciei la Uniunea European\u{a}, {\^\i}n ciuda refuzului continuu al Turciei de a recunoa\c{s}te Ciprul \c{s}i {\^\i}n ciuda faptului c\u{a} reformele democratice au ajuns {\^\i}ntr-un punct mort. = \\ 
 \midrule Target Completion $\to$ & The truth is that you want, at any price, and against the wishes of the peoples of Europe, to continue the negotiations for Turkey's accession to the European Union, despite Turkey's continuing refusal to recognise Cyprus and despite the fact that the democratic reforms are at a standstill. \\

\bottomrule \end{tabularx} }  \caption{Formatted dataset example for Ro$\to$En}  \label{eval:Ro_En}  \end{figure}

\begin{figure} { \tt \footnotesize \begin{tabularx}{\linewidth}{r X} \toprule Context $\to$ & Q: What is (2 * 4) * 6? \\
 & A: \\
 \midrule Target Completion $\to$ & 48 \\
 \bottomrule \end{tabularx} }  \caption{Formatted dataset example for Arithmetic 1DC}  \label{eval:single_digit_three_ops}  \end{figure}
\begin{figure} { \tt \footnotesize \begin{tabularx}{\linewidth}{r X} \toprule Context $\to$ & Q: What is 17 minus 14? \\
 & A: \\
 \midrule Target Completion $\to$ & 3 \\
 \bottomrule \end{tabularx} }  \caption{Formatted dataset example for Arithmetic 2D-}  \label{eval:two_digit_subtraction}  \end{figure}
\begin{figure} { \tt \footnotesize \begin{tabularx}{\linewidth}{r X} \toprule Context $\to$ & Q: What is 98 plus 45? \\
 & A: \\
 \midrule Target Completion $\to$ & 143 \\
 \bottomrule \end{tabularx} }  \caption{Formatted dataset example for Arithmetic 2D+}  \label{eval:two_digit_addition}  \end{figure}
    \begin{figure} { \tt \footnotesize \begin{tabularx}{\linewidth}{r X} \toprule Context $\to$ & Q: What is 95 times 45? \\
 & A: \\
 \midrule Target Completion $\to$ & 4275 \\
 \bottomrule \end{tabularx} }  \caption{Formatted dataset example for Arithmetic 2Dx}  \label{eval:two_digit_multiplication}  \end{figure}
\begin{figure} { \tt \footnotesize \begin{tabularx}{\linewidth}{r X} \toprule Context $\to$ & Q: What is 509 minus 488? \\
 & A: \\
 \midrule Target Completion $\to$ & 21 \\
 \bottomrule \end{tabularx} }  \caption{Formatted dataset example for Arithmetic 3D-}  \label{eval:three_digit_subtraction}  \end{figure}
\begin{figure} { \tt \footnotesize \begin{tabularx}{\linewidth}{r X} \toprule Context $\to$ & Q: What is 556 plus 497? \\
 & A: \\
 \midrule Target Completion $\to$ & 1053 \\
 \bottomrule \end{tabularx} }  \caption{Formatted dataset example for Arithmetic 3D+}  \label{eval:three_digit_addition}  \end{figure}
\begin{figure} { \tt \footnotesize \begin{tabularx}{\linewidth}{r X} \toprule Context $\to$ & Q: What is 6209 minus 3365? \\
 & A: \\
 \midrule Target Completion $\to$ & 2844 \\
 \bottomrule \end{tabularx} }  \caption{Formatted dataset example for Arithmetic 4D-}  \label{eval:four_digit_subtraction}  \end{figure}
\begin{figure} { \tt \footnotesize \begin{tabularx}{\linewidth}{r X} \toprule Context $\to$ & Q: What is 9923 plus 617? \\
 & A: \\
 \midrule Target Completion $\to$ & 10540 \\
 \bottomrule \end{tabularx} }  \caption{Formatted dataset example for Arithmetic 4D+}  \label{eval:four_digit_addition}  \end{figure}

\begin{figure}
{ \tt \footnotesize
\begin{tabularx}{\linewidth}{r X}
\toprule
Context $\to$ & Q:  What is 40649 minus 78746? \\
 & A: \\
 \midrule Target Completion $\to$ & -38097 \\
 \bottomrule \end{tabularx}
 }
 \caption{Formatted dataset example for Arithmetic 5D$-$}
 \label{eval:five_digit_subtraction}
 \end{figure}
\begin{figure} { \tt \footnotesize \begin{tabularx}{\linewidth}{r X} \toprule Context $\to$ & Q: What is 65360 plus 16204? \\
 & A: \\
 \midrule Target Completion $\to$ & 81564 \\
 \bottomrule \end{tabularx} }  \caption{Formatted dataset example for Arithmetic 5D+}  \label{eval:five_digit_addition}  \end{figure}

\clearpage
\newpage
\section{Results on All Tasks for All Model Sizes}
\label{appendix:results_on_all_tasks}


\begin{table}[!h]
{
    \setlength\tabcolsep{1pt}
    \centering
    \tiny
    \begin{adjustwidth}{-.8in}{-.8in}
    \begin{center}
    \begin{tabular}{llccl<{\hspace{4ex}}llllllll<{\hspace{4 ex}}llllllll<{\hspace{4ex}}llllllllc}
\toprule
& & & & & \multicolumn{8}{c}{Zero-Shot} & \multicolumn{8}{c}{One-Shot} & \multicolumn{8}{c}{Few-Shot}\\
\cmidrule(r{4ex}){6-13} \cmidrule(r{4ex}){14-21} \cmidrule(r){22-30} 
Name &   Metric  & Split &  \shortstack{Fine-tune\\SOTA} &    K &     Small &  Med &  Large &    XL &  2.7B &  6.7B &   13B &  175B &    Small &  Med &  Large &    XL &  2.7B &  6.7B &   13B &  175B &    Small &  Med &  Large &    XL &  2.7B &  6.7B &   13B &   175B & \shortstack{175B \\(test server)} \\ \midrule

     HellaSwag &      acc &    dev &           85.6 &   20 &      33.7 &    43.6 &   51.0 &  54.7 &  62.8 &  67.4 &  70.9 &  78.9 &     33.0 &    42.9 &   50.5 &  53.5 &  61.9 &  66.5 &  70.0 &  78.1 &     33.5 &    43.1 &   51.3 &  54.9 &  62.9 &  67.3 &  71.3 &   79.3 &                      \\
          LAMBADA &      acc &   test &           68.0 &   15 &      42.7 &    54.3 &   60.4 &  63.6 &  67.1 &  70.3 &  72.5 &  76.2 &     22.0 &    47.1 &   52.6 &  58.3 &  61.1 &  65.4 &  69.0 &  72.5 &     22.0 &    40.4 &   63.2 &  57.0 &  78.1 &  79.1 &  81.3 &   86.4 &                      \\
          LAMBADA &      ppl &   test &           8.63 &   15 &      18.6 &    9.09 &   6.53 &  5.44 &  4.60 &  4.00 &  3.56 &  3.00 &    165.0 &    11.6 &   8.29 &  6.46 &  5.53 &  4.61 &  4.06 &  3.35 &    165.0 &    27.6 &   6.63 &  7.45 &  2.89 &  2.56 &  2.56 &   1.92 &                      \\
      StoryCloze &      acc &   test &           91.8 &   70 &      63.3 &    68.5 &   72.4 &  73.4 &  77.2 &  77.7 &  79.5 &  83.2 &     62.3 &    68.7 &   72.3 &  74.2 &  77.3 &  78.7 &  79.7 &  84.7 &     62.3 &    70.2 &   73.9 &  76.1 &  80.2 &  81.2 &  83.0 &   87.7 &                      \\
                  &          &        &                &      &           &         &        &       &       &       &       &       &          &         &        &       &       &       &       &       &          &         &        &       &       &       &       &        &                      \\
              NQs &      acc &   test &           44.5 &   64 &      0.64 &    1.75 &   2.71 &  4.40 &  6.01 &  5.79 &  7.84 &  14.6 &     1.19 &    3.07 &   4.79 &  5.43 &  8.73 &  9.78 &  13.7 &  23.0 &     1.72 &    4.46 &   7.89 &  9.72 &  13.2 &  17.0 &  21.0 &   29.9 &                      \\
         TriviaQA &      acc &    dev &           68.0 &   64 &      4.15 &    7.61 &   14.0 &  19.7 &  31.3 &  38.7 &  41.8 &  64.3 &     4.19 &    12.9 &   20.5 &  26.5 &  35.9 &  44.4 &  51.3 &  68.0 &     6.96 &    16.3 &   26.5 &  32.1 &  42.3 &  51.6 &  57.5 &   71.2 &                 71.2 \\
            WebQs &      acc &   test &           45.5 &   64 &      1.77 &    3.20 &   4.33 &  4.63 &  7.92 &  7.73 &  8.22 &  14.4 &     2.56 &    6.20 &   8.51 &  9.15 &  14.5 &  15.1 &  19.0 &  25.3 &     5.46 &    12.6 &   15.9 &  19.6 &  24.8 &  27.7 &  33.5 &   41.5 &                      \\
                  &          &        &                &      &           &         &        &       &       &       &       &       &          &         &        &       &       &       &       &       &          &         &        &       &       &       &       &        &                      \\
     Ro$\to$En 16 &  BLEU-mb &   test &           39.9 &   64 &      2.08 &    2.71 &   3.09 &  3.15 &  16.3 &  8.34 &  20.2 &  19.9 &     0.55 &    15.4 &   23.0 &  26.3 &  30.6 &  33.2 &  35.6 &  38.6 &     1.25 &    20.7 &   25.8 &  29.2 &  33.1 &  34.8 &  37.0 &   39.5 &                      \\
     Ro$\to$En 16 &  BLEU-sb &   test &                &   64 &      2.39 &    3.08 &   3.49 &  3.56 &  16.8 &  8.75 &  20.8 &  20.9 &     0.65 &    15.9 &   23.6 &  26.8 &  31.3 &  34.2 &  36.7 &  40.0 &     1.40 &    21.3 &   26.6 &  30.1 &  34.3 &  36.2 &  38.4 &   41.3 &                      \\
     En$\to$Ro 16 &  BLEU-mb &   test &           38.5 &   64 &      2.14 &    2.65 &   2.53 &  2.50 &  3.46 &  4.24 &  5.32 &  14.1 &     0.35 &    3.30 &   7.89 &  8.72 &  13.2 &  15.1 &  17.3 &  20.6 &     1.25 &    5.90 &   9.33 &  10.7 &  14.3 &  16.3 &  18.0 &   21.0 &                      \\
     En$\to$Ro 16 &  BLEU-sb &   test &                &   64 &      2.61 &    3.11 &   3.07 &  3.09 &  4.26 &  5.31 &  6.43 &  18.0 &     0.55 &    3.90 &   9.15 &  10.3 &  15.7 &  18.2 &  20.8 &  24.9 &     1.64 &    7.40 &   10.9 &  12.9 &  17.2 &  19.6 &  21.8 &   25.8 &                      \\
     Fr$\to$En 14 &  BLEU-mb &   test &           35.0 &   64 &      1.81 &    2.53 &   3.47 &  3.13 &  20.6 &  15.1 &  21.8 &  21.2 &     1.28 &    15.9 &   23.7 &  26.3 &  29.0 &  30.5 &  30.2 &  33.7 &     4.98 &    25.5 &   28.5 &  31.1 &  33.7 &  34.9 &  36.6 &   39.2 &                      \\
     Fr$\to$En 14 &  BLEU-sb &   test &                &   64 &      2.29 &    2.99 &   3.90 &  3.60 &  21.2 &  15.5 &  22.4 &  21.9 &     1.50 &    16.3 &   24.4 &  27.0 &  30.0 &  31.6 &  31.4 &  35.6 &     5.30 &    26.2 &   29.5 &  32.2 &  35.1 &  36.4 &  38.3 &   41.4 &                      \\
     En$\to$Fr 14 &  BLEU-mb &   test &           45.6 &   64 &      1.74 &    2.16 &   2.73 &  2.15 &  15.1 &  8.82 &  12.0 &  25.2 &     0.49 &    8.00 &   14.8 &  15.9 &  20.3 &  23.3 &  24.9 &  28.3 &     4.08 &    14.5 &   19.3 &  21.5 &  24.9 &  27.3 &  29.5 &   32.6 &                      \\
     En$\to$Fr 14 &  BLEU-sb &   test &           45.9 &   64 &      2.44 &    2.75 &   3.54 &  2.82 &  19.3 &  11.4 &  15.3 &  31.3 &     0.81 &    10.0 &   18.2 &  19.3 &  24.7 &  28.3 &  30.1 &  34.1 &     5.31 &    18.0 &   23.6 &  26.1 &  30.3 &  33.3 &  35.5 &   39.9 &                      \\
     De$\to$En 16 &  BLEU-mb &   test &           40.2 &   64 &      2.06 &    2.87 &   3.41 &  3.63 &  21.5 &  17.3 &  23.0 &  27.2 &     0.83 &    16.2 &   22.5 &  24.7 &  28.2 &  30.7 &  33.0 &  30.4 &     3.25 &    22.7 &   26.2 &  29.2 &  32.7 &  34.8 &  37.3 &   40.6 &                      \\
     De$\to$En 16 &  BLEU-sb &   test &                &   64 &      2.39 &    3.27 &   3.85 &  4.04 &  22.5 &  18.2 &  24.4 &  28.6 &     0.93 &    17.1 &   23.4 &  25.8 &  29.2 &  31.9 &  34.5 &  32.1 &     3.60 &    23.8 &   27.5 &  30.5 &  34.1 &  36.5 &  39.1 &   43.0 &                      \\
     En$\to$De 16 &  BLEU-mb &   test &            41.2 &   64 &      1.70 &    2.27 &   2.31 &  2.43 &  12.9 &  8.66 &  10.4 &  24.6 &     0.50 &    7.00 &   12.9 &  13.1 &  18.3 &  20.9 &  22.5 &  26.2 &     3.42 &    12.3 &   15.4 &  17.1 &  20.9 &  23.0 &  26.6 &   29.7 &                      \\
     En$\to$De 16 &  BLEU-sb &   test &           41.2 &   64 &      2.09 &    2.65 &   2.75 &  2.92 &  13.7 &  9.36 &  11.0 &  25.3 &     0.54 &    7.40 &   13.4 &  13.4 &  18.8 &  21.7 &  23.3 &  27.3 &     3.78 &    12.9 &   16.1 &  17.7 &  21.7 &  24.1 &  27.7 &   30.9 &                      \\
                  &          &        &                &      &           &         &        &       &       &       &       &       &          &         &        &       &       &       &       &       &          &         &        &       &       &       &       &        &                      \\
         Winograd &      acc &   test &           93.8 &    7 &      66.3 &    72.9 &   74.7 &  76.9 &  82.4 &  85.7 &  87.9 &  88.3 &     63.4 &    68.5 &   72.9 &  76.9 &  82.4 &  84.6 &  86.1 &  89.7 &     63.4 &    67.4 &   73.6 &  76.9 &  84.3 &  85.4 &  82.4 &   88.6 &                      \\
      Winogrande &      acc &    dev &           84.6 &   50 &      52.0 &    52.1 &   57.4 &  58.7 &  62.3 &  64.5 &  67.9 &  70.2 &     51.3 &    53.0 &   58.3 &  59.1 &  61.7 &  65.8 &  66.9 &  73.2 &     51.3 &    52.6 &   57.5 &  59.1 &  62.6 &  67.4 &  70.0 &   77.7 &                      \\
                  &          &        &                &      &           &         &        &       &       &       &       &       &          &         &        &       &       &       &       &       &          &         &        &       &       &       &       &        &                      \\
             PIQA &      acc &    dev &           77.1 &   50 &      64.6 &    70.2 &   72.9 &  75.1 &  75.6 &  78.0 &  78.5 &  81.0 &     64.3 &    69.3 &   71.8 &  74.4 &  74.3 &  76.3 &  77.8 &  80.5 &     64.3 &    69.4 &   72.0 &  74.3 &  75.4 &  77.8 &  79.9 &   82.3 &                 82.8 \\
  ARC (Challenge) &      acc &   test &           78.5 &   50 &      26.6 &    29.5 &   31.8 &  35.5 &  38.0 &  41.4 &  43.7 &  51.4 &     25.5 &    30.2 &   31.6 &  36.4 &  38.4 &  41.5 &  43.1 &  53.2 &     25.5 &    28.4 &   32.3 &  36.7 &  39.5 &  43.7 &  44.8 &   51.5 &                      \\
      ARC (Easy) &      acc &   test &           92.0 &   50 &      43.6 &    46.5 &   53.0 &  53.8 &  58.2 &  60.2 &  63.8 &  68.8 &     42.7 &    48.2 &   54.6 &  55.9 &  60.3 &  62.6 &  66.8 &  71.2 &     42.7 &    51.0 &   58.1 &  59.1 &  62.1 &  65.8 &  69.1 &   70.1 &                      \\
      OpenBookQA &      acc &   test &           87.2 &  100 &      35.6 &    43.2 &   45.2 &  46.8 &  53.0 &  50.4 &  55.6 &  57.6 &     37.0 &    39.8 &   46.2 &  46.4 &  53.4 &  53.0 &  55.8 &  58.8 &     37.0 &    43.6 &   48.0 &  50.6 &  55.6 &  55.2 &  60.8 &   65.4 &                      \\
                  &          &        &                &      &           &         &        &       &       &       &       &       &          &         &        &       &       &       &       &       &          &         &        &       &       &       &       &        &                      \\
             Quac &       f1 &    dev &           74.4 &    5 &      21.2 &    26.8 &   31.0 &  30.1 &  34.7 &  36.1 &  38.4 &  41.5 &     21.1 &    26.9 &   31.9 &  32.3 &  37.4 &  39.0 &  40.6 &  43.4 &     21.6 &    27.6 &   32.9 &  34.2 &  38.2 &  39.9 &  40.9 &   44.3 &                      \\
          RACE-h &      acc &   test &           90.0 &   10 &      35.2 &    37.9 &   40.1 &  40.9 &  42.4 &  44.1 &  44.6 &  45.5 &     34.3 &    37.7 &   40.0 &  42.0 &  43.8 &  44.3 &  44.6 &  45.9 &     34.3 &    37.0 &   40.4 &  41.4 &  42.3 &  44.7 &  45.1 &   46.8 &                      \\
          RACE-m &      acc &   test &           93.1 &   10 &      42.1 &    47.2 &   52.1 &  52.3 &  54.7 &  54.4 &  56.7 &  58.4 &     42.3 &    47.3 &   51.7 &  55.2 &  56.1 &  54.7 &  56.9 &  57.4 &     42.3 &    47.0 &   52.7 &  53.0 &  55.6 &  55.4 &  58.1 &   58.1 &                      \\
          SQuADv2 &       em &    dev &           90.7 &   16 &      22.6 &    32.8 &   33.9 &  43.1 &  43.6 &  45.4 &  49.0 &  52.6 &     25.1 &    37.5 &   37.9 &  47.9 &  47.9 &  51.1 &  56.0 &  60.1 &     27.5 &    40.5 &   39.2 &  53.5 &  50.0 &  56.6 &  62.6 &   64.9 &                      \\
          SQuADv2 &       f1 &    dev &           93.0 &   16 &      28.3 &    40.2 &   41.4 &  50.3 &  51.0 &  52.7 &  56.3 &  59.5 &     30.1 &    43.6 &   44.1 &  54.0 &  54.1 &  57.1 &  61.8 &  65.4 &     32.1 &    45.5 &   44.9 &  58.7 &  55.9 &  62.1 &  67.7 &   69.8 &                      \\
             CoQA &       f1 &    dev &           90.7 &    5 &      34.5 &    55.0 &   61.8 &  65.3 &  71.1 &  72.8 &  76.3 &  81.5 &     30.6 &    52.1 &   61.6 &  66.1 &  71.8 &  75.1 &  77.9 &  84.0 &     31.1 &    52.0 &   62.7 &  66.8 &  73.2 &  77.3 &  79.9 &   85.0 &                      \\
             DROP &       f1 &    dev &           89.1 &   20 &      9.40 &    13.6 &   14.4 &  16.4 &  19.7 &  17.0 &  24.0 &  23.6 &     11.7 &    18.1 &   20.9 &  23.0 &  26.4 &  27.3 &  29.2 &  34.3 &     12.9 &    18.7 &   24.0 &  25.6 &  29.7 &  29.7 &  32.3 &   36.5 &                      \\
                  &          &        &                &      &           &         &        &       &       &       &       &       &          &         &        &       &       &       &       &       &          &         &        &       &       &       &       &        &                      \\
            BoolQ &      acc &    dev &           91.0 &   32 &      49.7 &    60.3 &   58.9 &  62.4 &  67.1 &  65.4 &  66.2 &  60.5 &     52.6 &    61.7 &   60.4 &  63.7 &  68.4 &  68.7 &  69.0 &  76.7 &     43.1 &    60.6 &   62.0 &  64.1 &  70.3 &  70.0 &  70.2 &   77.5 &                 76.4 \\
              CB &      acc &    dev &           96.9 &   32 &      0.00 &    32.1 &   8.93 &  19.6 &  19.6 &  28.6 &  19.6 &  46.4 &     55.4 &    53.6 &   53.6 &  48.2 &  57.1 &  33.9 &  55.4 &  64.3 &     42.9 &    58.9 &   53.6 &  69.6 &  67.9 &  60.7 &  66.1 &   82.1 &                 75.6 \\
              CB &       f1 &    dev &           93.9 &   32 &      0.00 &    29.3 &   11.4 &  17.4 &  22.4 &  25.1 &  20.3 &  42.8 &     60.1 &    39.8 &   45.6 &  37.5 &  45.7 &  28.5 &  44.6 &  52.5 &     26.1 &    40.4 &   32.6 &  48.3 &  45.7 &  44.6 &  46.0 &   57.2 &                 52.0 \\
             Copa &      acc &    dev &           94.8 &   32 &      66.0 &    68.0 &   73.0 &  77.0 &  76.0 &  80.0 &  84.0 &  91.0 &     62.0 &    64.0 &   66.0 &  74.0 &  76.0 &  82.0 &  86.0 &  87.0 &     67.0 &    64.0 &   72.0 &  77.0 &  83.0 &  83.0 &  86.0 &   92.0 &                 92.0 \\
              RTE &      acc &    dev &           92.5 &   32 &      47.7 &    49.8 &   48.4 &  56.0 &  46.6 &  55.2 &  62.8 &  63.5 &     53.1 &    47.3 &   49.5 &  49.5 &  54.9 &  54.9 &  56.3 &  70.4 &     52.3 &    48.4 &   46.9 &  50.9 &  56.3 &  49.5 &  60.6 &   72.9 &                 69.0 \\
              WiC &      acc &    dev &           76.1 &   32 &      0.00 &    0.00 &   0.00 &  0.00 &  0.00 &  0.00 &  0.00 &  0.00 &     50.0 &    50.3 &   50.3 &  49.2 &  49.4 &  50.3 &  50.0 &  48.6 &     49.8 &    55.0 &   53.0 &  53.0 &  51.6 &  53.1 &  51.1 &   55.3 &                 49.4 \\
              WSC &      acc &    dev &           93.8 &   32 &      59.6 &    56.7 &   65.4 &  61.5 &  66.3 &  60.6 &  64.4 &  65.4 &     58.7 &    58.7 &   60.6 &  62.5 &  66.3 &  60.6 &  66.3 &  69.2 &     58.7 &    60.6 &   54.8 &  49.0 &  62.5 &  67.3 &  75.0 &   75.0 &                 80.1 \\
          MultiRC &      acc &    dev &           62.3 &   32 &      4.72 &    9.65 &   12.3 &  13.6 &  14.3 &  18.4 &  24.2 &  27.6 &     4.72 &    9.65 &   12.3 &  13.6 &  14.3 &  18.4 &  24.2 &  27.6 &     6.09 &    11.8 &   16.8 &  20.8 &  24.7 &  23.8 &  25.0 &   32.5 &                 30.5 \\
          MultiRC &      f1a &    dev &           88.2 &   32 &      57.0 &    59.7 &   60.4 &  59.9 &  60.0 &  64.5 &  71.4 &  72.9 &     57.0 &    59.7 &   60.4 &  59.9 &  60.0 &  64.5 &  71.4 &  72.9 &     45.0 &    55.9 &   64.2 &  65.4 &  69.5 &  66.4 &  69.3 &   74.8 &                 75.4 \\
          ReCoRD &      acc &    dev &           92.5 &   32 &      70.8 &    78.5 &   82.1 &  84.1 &  86.2 &  88.6 &  89.0 &  90.2 &     69.8 &    77.0 &   80.7 &  83.0 &  85.9 &  88.0 &  88.8 &  90.2 &     69.8 &    77.2 &   81.3 &  83.1 &  86.6 &  87.9 &  88.9 &   89.0 &                 90.2 \\
          ReCoRD &       f1 &    dev &           93.3 &   32 &      71.9 &    79.2 &   82.8 &  85.2 &  87.3 &  89.5 &  90.4 &  91.0 &     70.7 &    77.8 &   81.6 &  83.9 &  86.8 &  88.8 &  89.7 &  91.2 &     70.7 &    77.9 &   82.1 &  84.0 &  87.5 &  88.8 &  89.8 &   90.1 &                 91.1 \\
        SuperGLUE &  average &    dev &           89.0 &      &      40.6 &    47.4 &   46.8 &  49.6 &  50.1 &  52.3 &  54.4 &  58.2 &     54.4 &    55.1 &   56.7 &  57.8 &  61.2 &  59.7 &  64.3 &  68.9 &     50.2 &    56.2 &   56.8 &  60.0 &  64.3 &  63.6 &  66.9 &   73.2 &                 71.8 \\
                  &          &        &                &      &           &         &        &       &       &       &       &       &          &         &        &       &       &       &       &       &          &         &        &       &       &       &       &        &                      \\
          ANLI R1 &      acc &   test &           73.8 &   50 &      33.4 &    34.2 &   33.4 &  33.4 &  34.2 &  32.3 &  33.2 &  34.6 &     32.1 &    31.6 &   31.9 &  34.6 &  30.6 &  31.6 &  32.7 &  32.0 &     32.1 &    32.5 &   30.9 &  32.5 &  33.5 &  33.1 &  33.3 &   36.8 &                      \\
          ANLI R2 &      acc &   test &           50.7 &   50 &      33.2 &    31.9 &   33.3 &  33.3 &  33.8 &  33.5 &  33.5 &  35.4 &     35.7 &    33.7 &   33.2 &  32.7 &  32.7 &  33.9 &  33.9 &  33.9 &     35.7 &    33.8 &   32.1 &  31.4 &  32.6 &  33.3 &  32.6 &   34.0 &                      \\
          ANLI R3 &      acc &   test &           48.3 &   50 &      33.6 &    34.0 &   33.8 &  33.4 &  35.3 &  34.8 &  34.4 &  34.5 &     35.0 &    32.6 &   33.0 &  33.9 &  34.1 &  33.1 &  32.5 &  35.1 &     35.0 &    34.4 &   35.1 &  36.0 &  32.7 &  33.9 &  34.5 &   40.2 &                      \\
                  &          &        &                &      &           &         &        &       &       &       &       &       &          &         &        &       &       &       &       &       &          &         &        &       &       &       &       &        &                      \\
              2D+ &      acc &    n/a &                &   50 &      0.70 &    0.65 &   0.70 &  0.85 &  1.10 &  2.54 &  15.4 &  76.9 &     2.00 &    0.55 &   3.15 &  4.00 &  12.1 &  19.6 &  73.0 &  99.6 &     2.00 &    4.10 &   3.50 &  4.50 &  8.90 &  11.9 &  55.5 &  100.0 &                      \\
              2D- &      acc &    n/a &                &   50 &      1.25 &    1.25 &   1.25 &  1.25 &  1.60 &  7.60 &  12.6 &  58.0 &     1.15 &    0.95 &   1.45 &  1.95 &  3.85 &  11.5 &  44.6 &  86.4 &     1.15 &    1.45 &   2.25 &  2.70 &  7.35 &  13.6 &  52.4 &   98.9 &                      \\
              3D+ &      acc &    n/a &                &   50 &      0.10 &    0.10 &   0.05 &  0.10 &  0.10 &  0.25 &  1.40 &  34.2 &     0.15 &    0.00 &   0.10 &  0.30 &  0.45 &  0.95 &  15.4 &  65.5 &     0.15 &    0.45 &   0.30 &  0.55 &  0.75 &  0.90 &  8.40 &   80.4 &                      \\
              3D- &      acc &    n/a &                &   50 &      0.05 &    0.05 &   0.05 &  0.05 &  0.05 &  0.45 &  1.35 &  48.3 &     0.05 &    0.15 &   0.25 &  0.30 &  0.55 &  1.60 &  6.15 &  78.7 &     0.05 &    0.10 &   0.15 &  0.35 &  0.65 &  1.05 &  9.20 &   94.2 &                      \\
              4D+ &      acc &    n/a &                &   50 &      0.05 &    0.05 &   0.00 &  0.00 &  0.05 &  0.05 &  0.15 &  4.00 &     0.00 &    0.00 &   0.10 &  0.00 &  0.00 &  0.10 &  0.80 &  14.0 &     0.00 &    0.05 &   0.05 &  0.00 &  0.15 &  0.15 &  0.40 &   25.5 &                      \\
              4D- &      acc &    n/a &                &   50 &      0.00 &    0.00 &   0.00 &  0.00 &  0.00 &  0.00 &  0.10 &  7.50 &     0.00 &    0.00 &   0.00 &  0.00 &  0.05 &  0.00 &  0.50 &  14.0 &     0.00 &    0.05 &   0.00 &  0.00 &  0.10 &  0.05 &  0.40 &   26.8 &                      \\
              5D+ &      acc &    n/a &                &   50 &      0.00 &    0.00 &   0.00 &  0.00 &  0.00 &  0.00 &  0.00 &  0.65 &     0.00 &    0.00 &   0.00 &  0.00 &  0.00 &  0.00 &  0.05 &  3.45 &     0.00 &    0.00 &   0.00 &  0.00 &  0.00 &  0.00 &  0.05 &   9.30 &                      \\
              5D- &      acc &    n/a &                &   50 &      0.00 &    0.00 &   0.00 &  0.00 &  0.00 &  0.00 &  0.00 &  0.80 &     0.00 &    0.00 &   0.00 &  0.00 &  0.00 &  0.00 &  0.05 &  3.75 &     0.00 &    0.00 &   0.00 &  0.00 &  0.00 &  0.00 &  0.00 &   9.90 &                      \\
              2Dx &      acc &    n/a &                &   50 &      2.20 &    2.25 &   2.65 &  2.10 &  2.55 &  5.80 &  6.15 &  19.8 &     1.35 &    2.35 &   3.35 &  2.35 &  4.75 &  9.15 &  11.0 &  27.4 &     1.35 &    2.90 &   2.70 &  2.85 &  4.25 &  6.10 &  7.05 &   29.2 &                      \\
              1DC &      acc &    n/a &                &   50 &      1.25 &    2.95 &   2.75 &  0.05 &  0.30 &  2.35 &  0.75 &  9.75 &     1.90 &    2.80 &   2.85 &  3.65 &  6.45 &  9.15 &  8.20 &  14.3 &     1.70 &    2.15 &   3.90 &  5.75 &  6.20 &  7.60 &  9.95 &   21.3 &                      \\
                  &          &        &                &      &           &         &        &       &       &       &       &       &          &         &        &       &       &       &       &       &          &         &        &       &       &       &       &        &                      \\
  Cycled Letters &      acc &    n/a &                &  100 &      0.62 &    0.71 &   2.85 &  0.00 &  0.63 &  1.35 &  2.58 &  3.66 &     1.67 &    4.36 &   5.68 &  6.46 &  6.25 &  9.41 &  15.1 &  21.7 &     4.63 &    9.27 &   10.7 &  14.5 &  16.7 &  21.9 &  27.7 &   37.9 &                      \\
      Anagrams 1 &      acc &    n/a &                &  100 &      0.10 &    0.14 &   0.40 &  0.00 &  0.27 &  0.69 &  1.16 &  2.28 &     0.21 &    0.61 &   1.12 &  1.27 &  1.60 &  2.72 &  3.72 &  8.62 &     0.50 &    1.27 &   2.13 &  3.05 &  3.81 &  5.49 &  8.38 &   15.1 &                      \\
      Anagrams 2 &      acc &    n/a &                &  100 &      0.81 &    1.21 &   2.69 &  0.01 &  1.71 &  3.75 &  4.53 &  8.91 &     1.19 &    2.62 &   4.70 &  4.77 &  6.97 &  10.2 &  14.6 &  25.9 &     1.94 &    4.80 &   7.59 &  9.87 &  12.6 &  18.9 &  25.6 &   39.7 &                      \\
 Symbol Insertion &      acc &    n/a &                &  100 &      0.00 &    0.00 &   0.10 &  0.00 &  0.05 &  0.42 &  0.89 &  8.26 &     0.03 &    0.05 &   0.57 &  1.18 &  1.67 &  3.46 &  6.62 &  45.4 &     0.11 &    0.28 &   2.19 &  4.18 &  6.61 &  11.0 &  27.3 &   67.2 &                      \\
  Reversed Words &      acc &    n/a &                &  100 &      0.00 &    0.01 &   0.01 &  0.01 &  0.02 &  0.03 &  0.03 &  0.09 &     0.02 &    0.01 &   0.01 &  0.00 &  0.05 &  0.07 &  0.11 &  0.48 &     0.00 &    0.05 &   0.00 &  0.17 &  0.24 &  0.30 &  0.42 &   0.44 &                      \\
                  &          &        &                &      &           &         &        &       &       &       &       &       &          &         &        &       &       &       &       &       &          &         &        &       &       &       &       &        &                      \\
    SAT Analogies &      acc &    n/a &                &   20 &      35.6 &    39.0 &   45.2 &  44.1 &  50.0 &  49.2 &  52.7 &  53.7 &     30.5 &    41.2 &   43.1 &  46.5 &  55.1 &  54.3 &  53.5 &  59.1 &     30.5 &    40.4 &   42.8 &  40.6 &  48.4 &  51.9 &  53.5 &   65.2 &                      \\

\bottomrule
\end{tabular}
    \end{center}
\end{adjustwidth}

\caption{Scores for every task, setting and model that we investigate in this paper. }
}
\label{table:master}
\end{table}

\begin{figure}
\begin{tabular}{lll}
\toprule
      \includegraphics[width=0.33\linewidth]{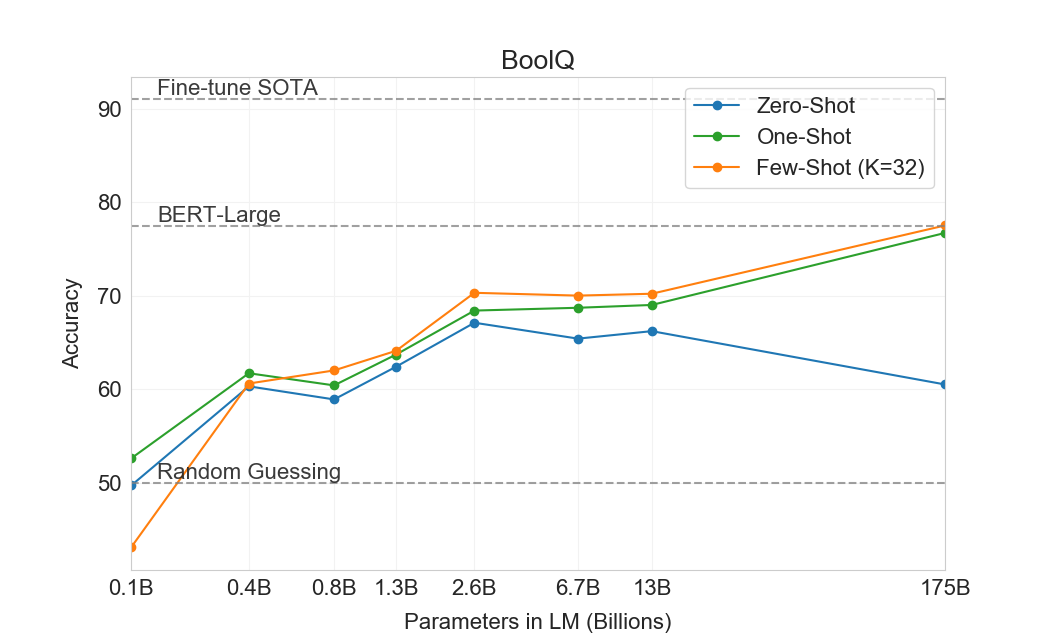} &      \includegraphics[width=0.33\linewidth]{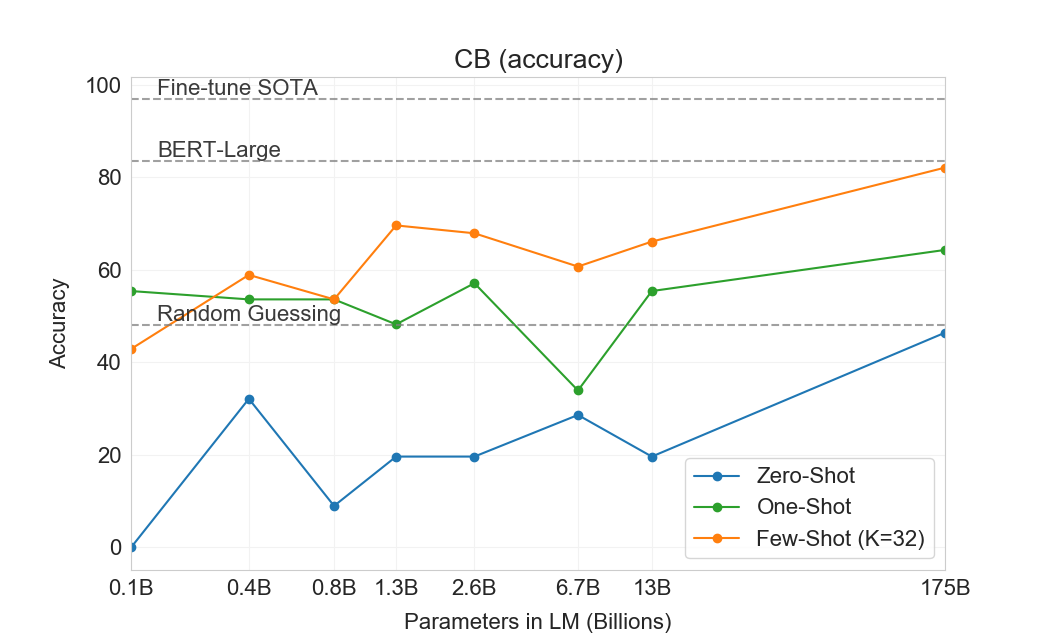} &        \includegraphics[width=0.33\linewidth]{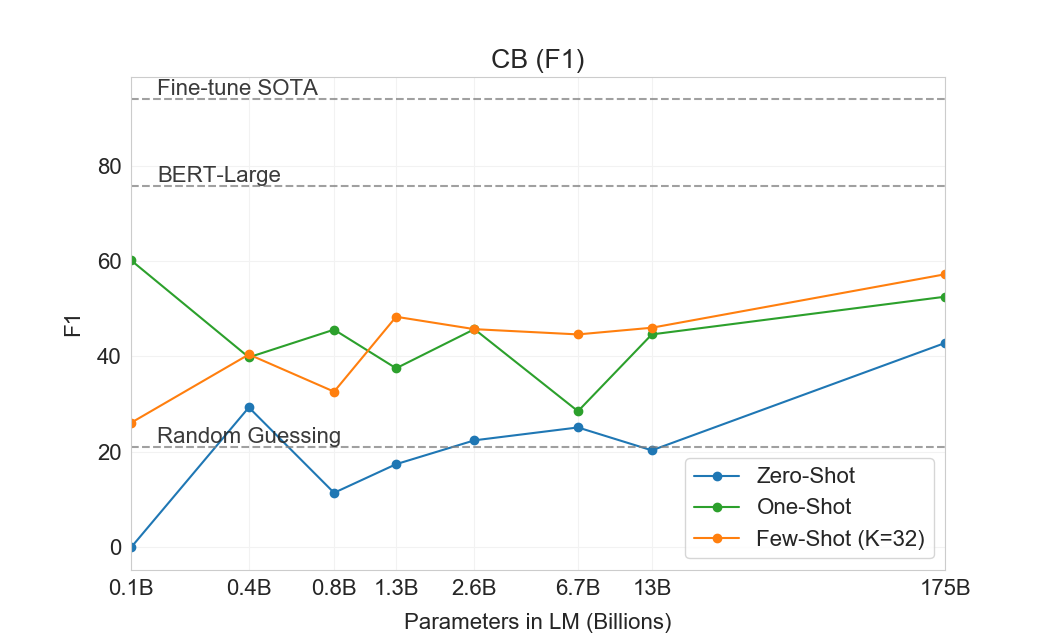} \\
       \includegraphics[width=0.33\linewidth]{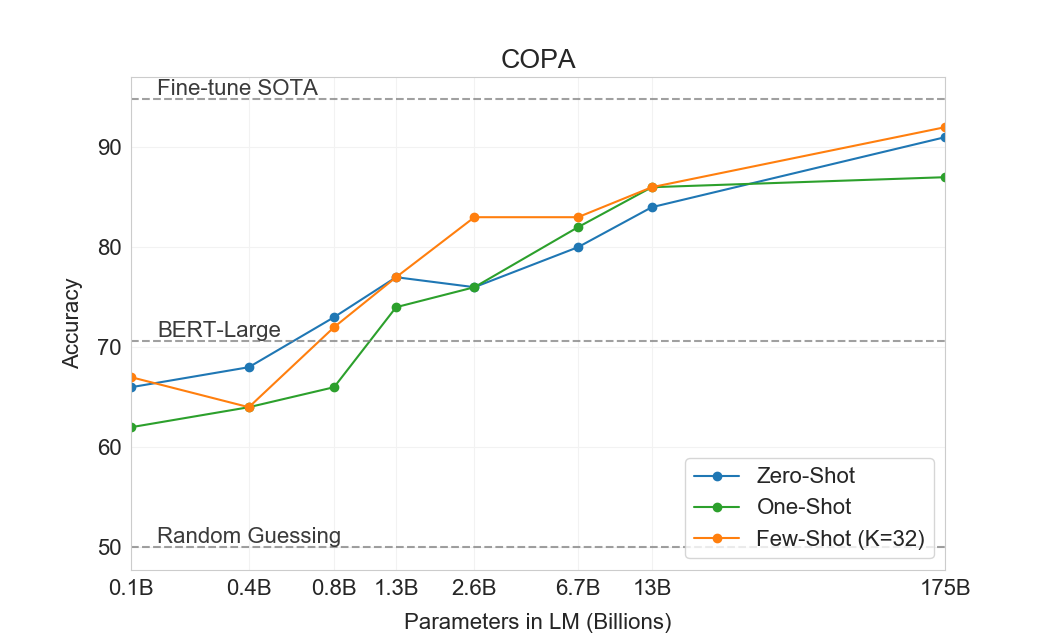} &         \includegraphics[width=0.33\linewidth]{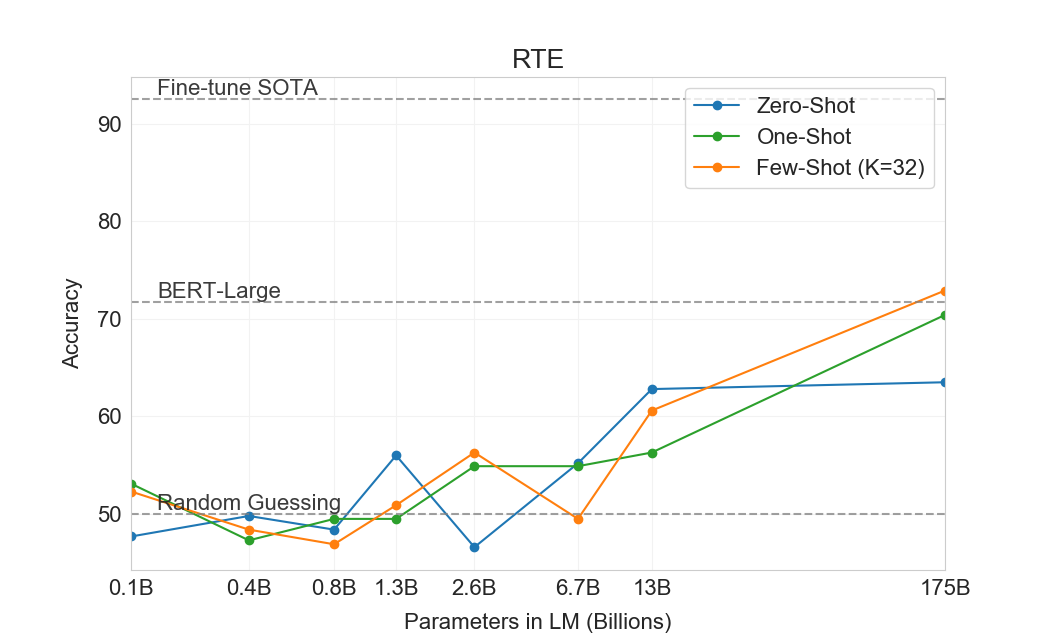} &          \includegraphics[width=0.33\linewidth]{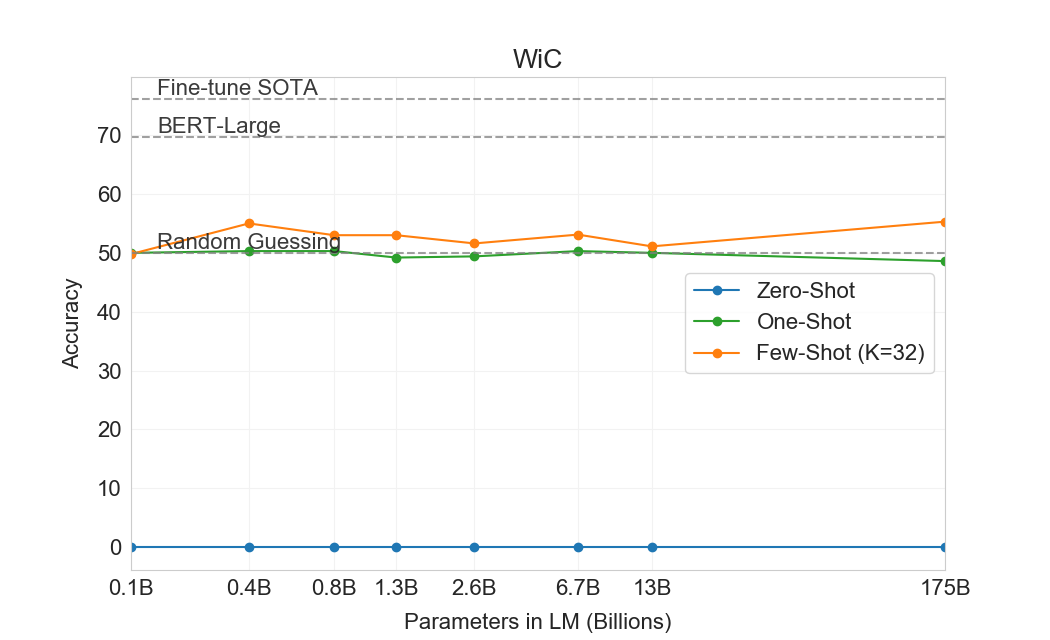} \\
        \includegraphics[width=0.33\linewidth]{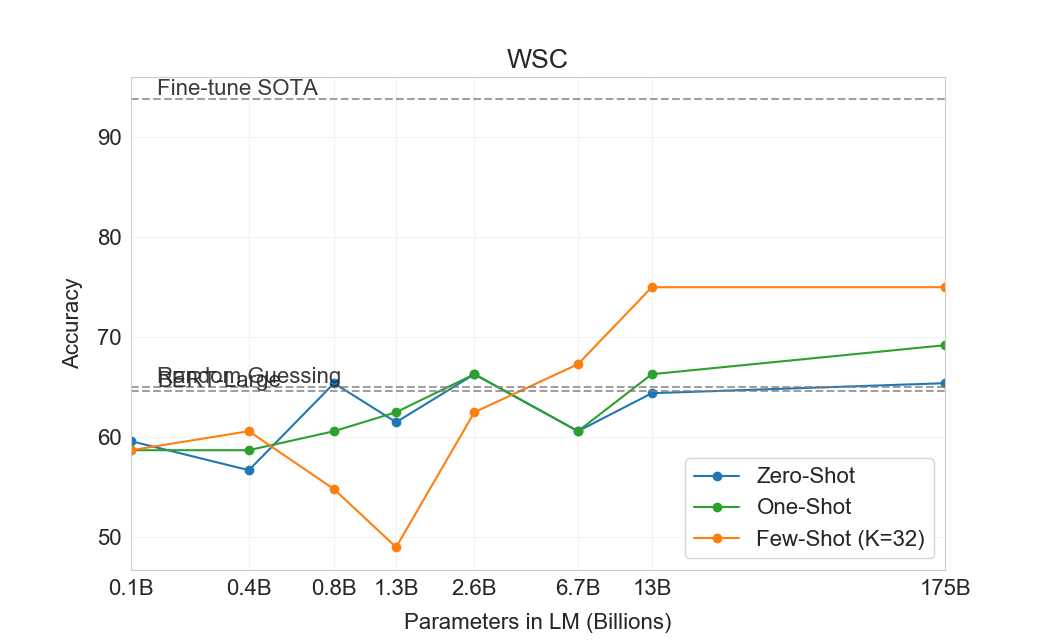} &  \includegraphics[width=0.33\linewidth]{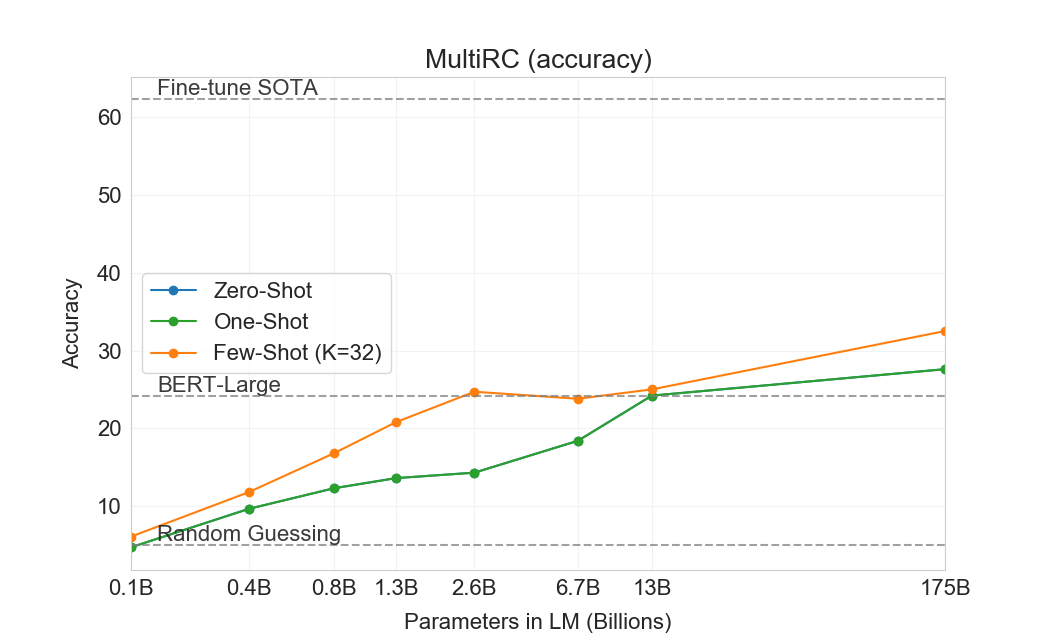} &  \includegraphics[width=0.33\linewidth]{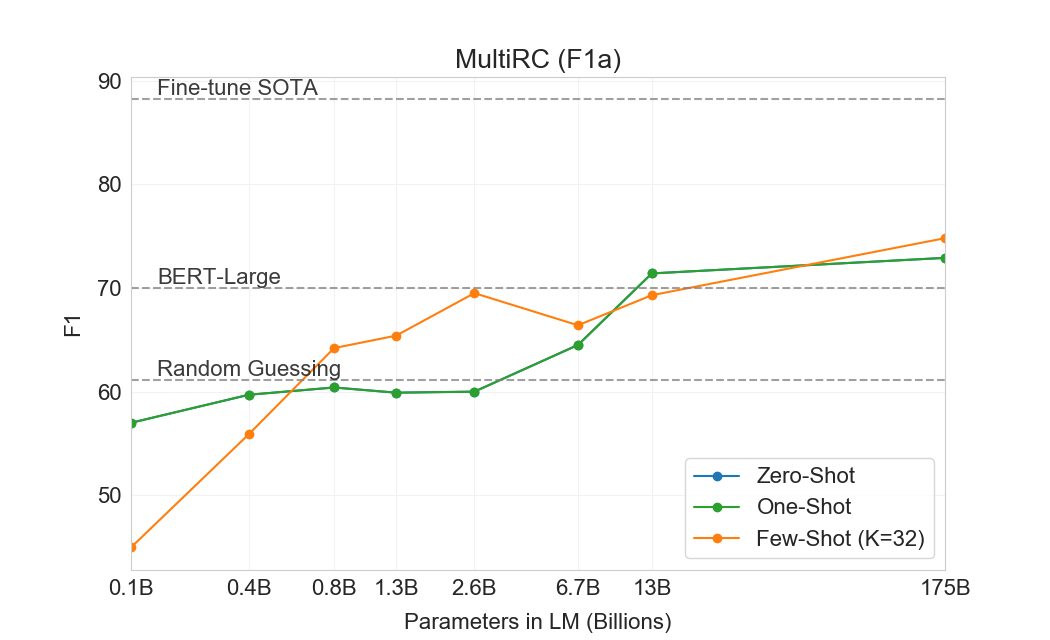} \\
 \includegraphics[width=0.33\linewidth]{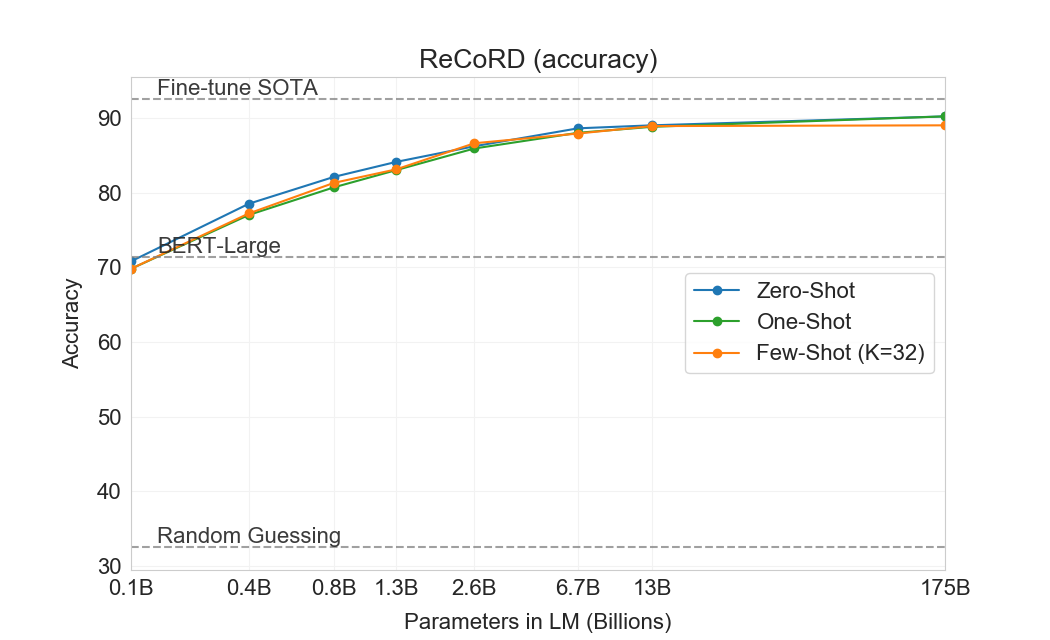} &   \includegraphics[width=0.33\linewidth]{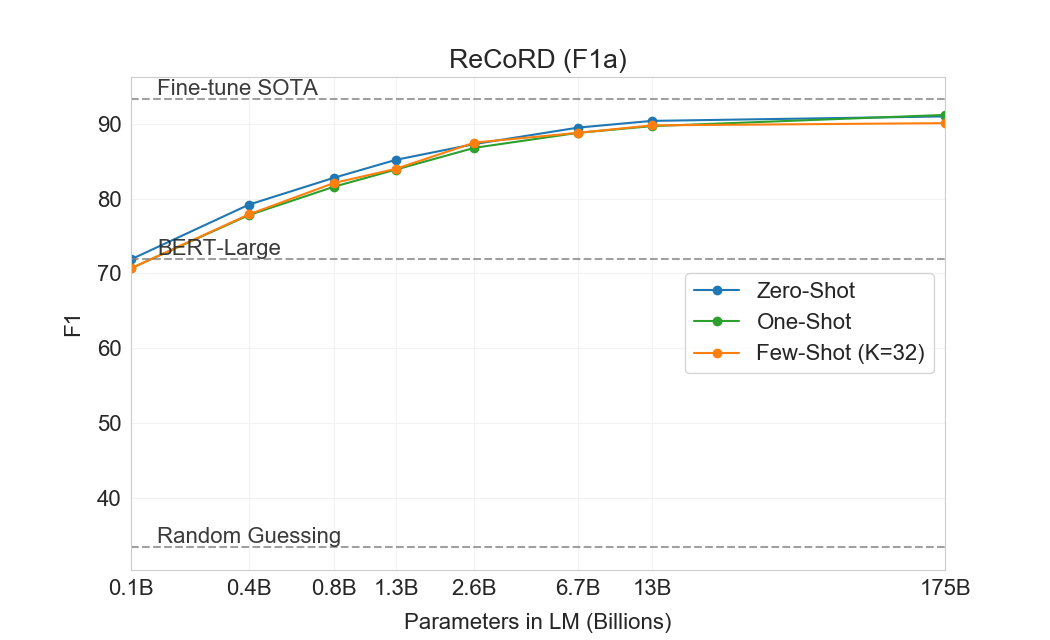} &                                                                                    \\
\bottomrule
\end{tabular}
\caption{All results for all SuperGLUE tasks.}
\label{graph:all_superglue}
\end{figure}

\begin{figure}
\centering
\begin{minipage}{.33\textwidth}
      \begin{tabular}{c}
      \toprule
      \includegraphics[height=1.3in]{graphs/scale_plots/sat_analogies.png} \\
      \bottomrule
      \end{tabular}
\captionof{figure}{Results for SAT task.}
\label{graph:all_sat}
\end{minipage}
\hfill
\begin{minipage}{.6\textwidth}
\begin{tabular}{c c}
\toprule
\includegraphics[height=1.3in]{graphs/scale_plots/winogrande.png} &
\includegraphics[height=1.3in]{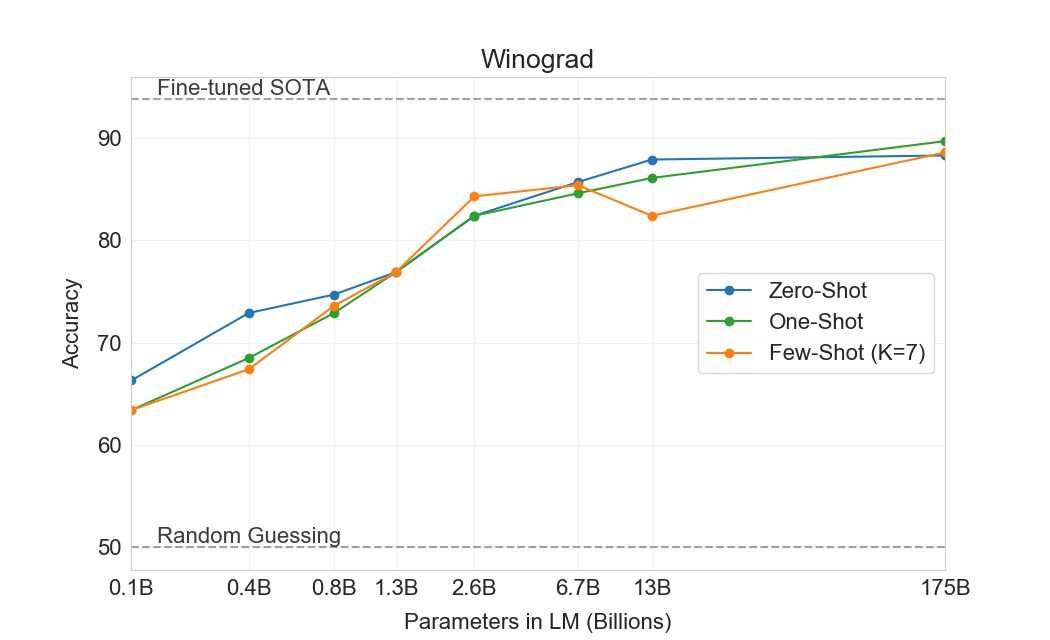} \\
\bottomrule
\end{tabular}
\caption{All results for all Winograd tasks.}
\label{figure:all_winograd}
\end{minipage}
\end{figure}


\begin{figure}
\begin{tabular}{lll}
\toprule
      \includegraphics[width=0.33\linewidth]{graphs/scale_plots/arithmetic.png} &      \includegraphics[width=0.33\linewidth]{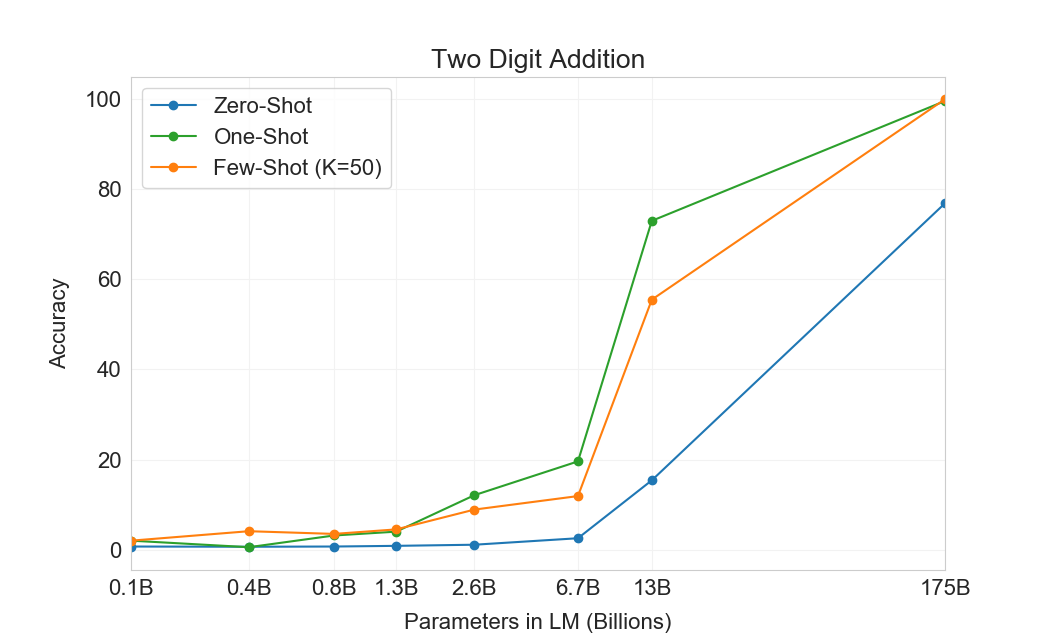} &        \includegraphics[width=0.33\linewidth]{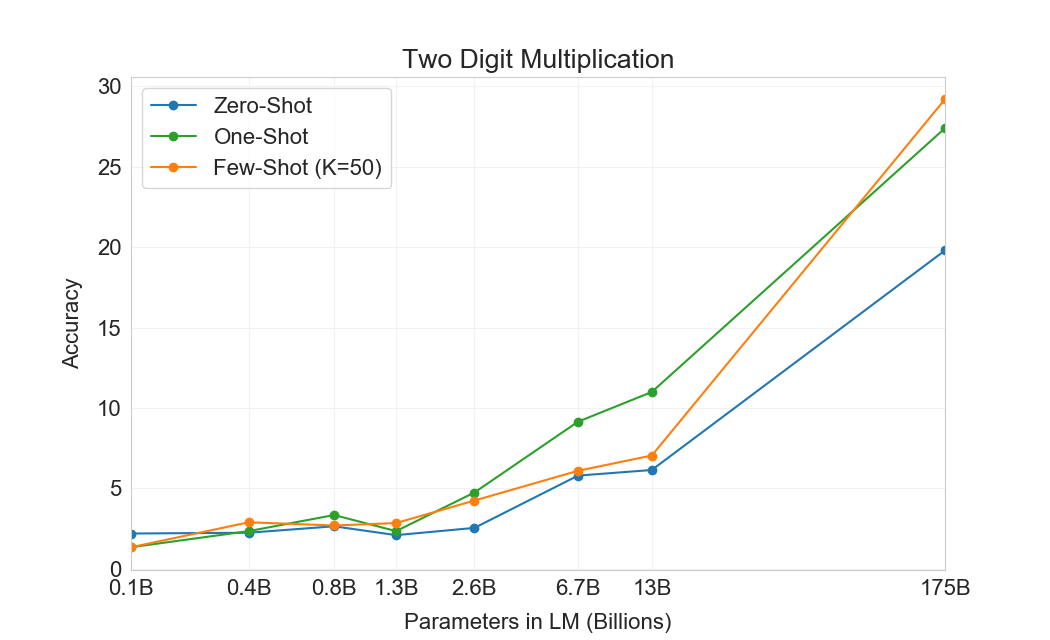} \\
       \includegraphics[width=0.33\linewidth]{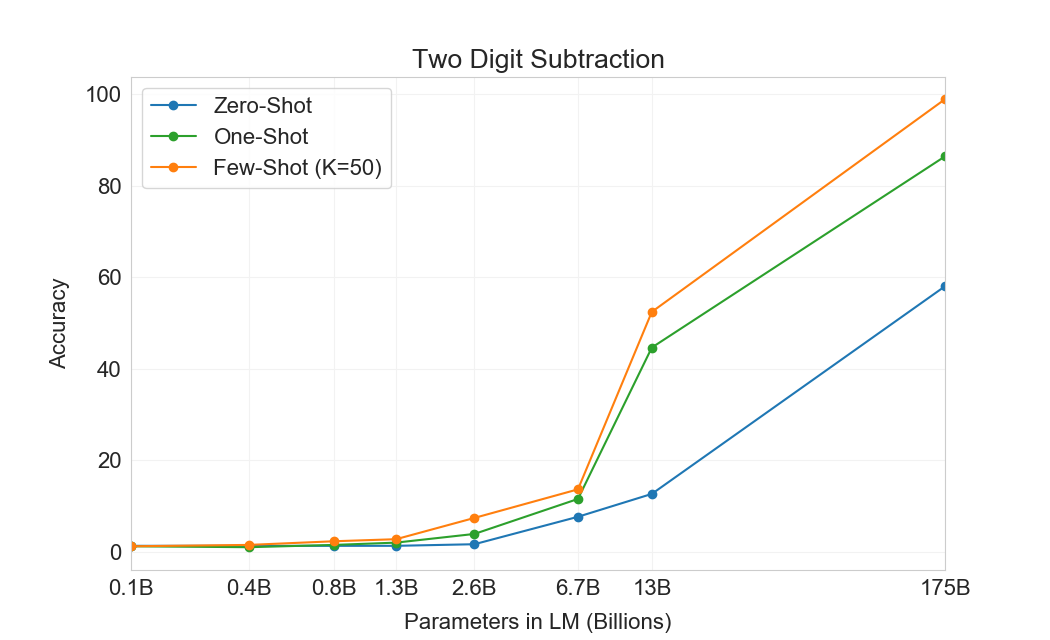} &         \includegraphics[width=0.33\linewidth]{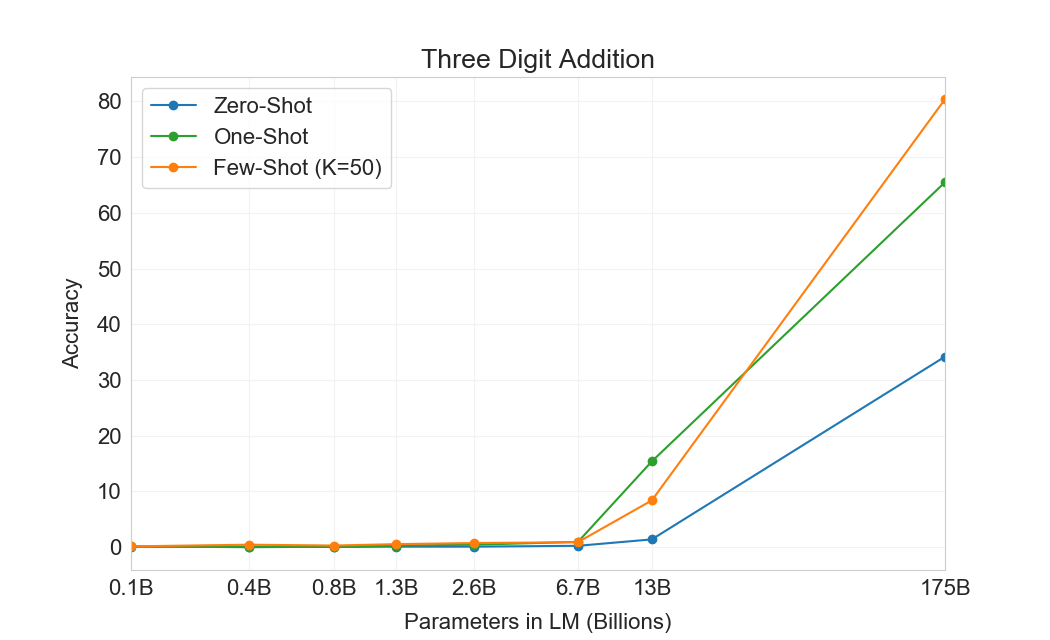} &          \includegraphics[width=0.33\linewidth]{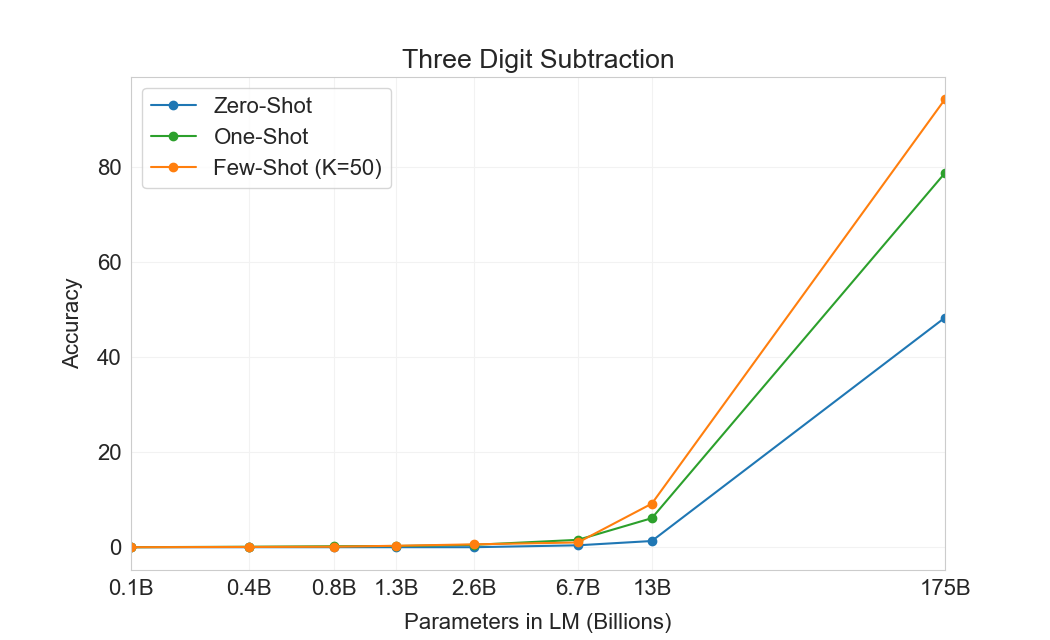} \\
        \includegraphics[width=0.33\linewidth]{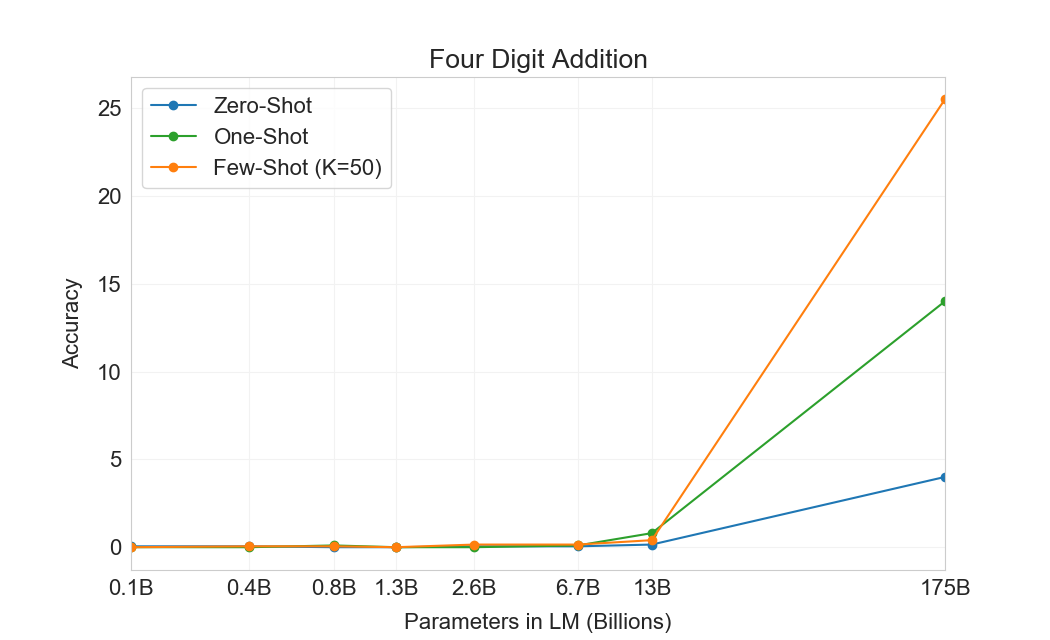} &  \includegraphics[width=0.33\linewidth]{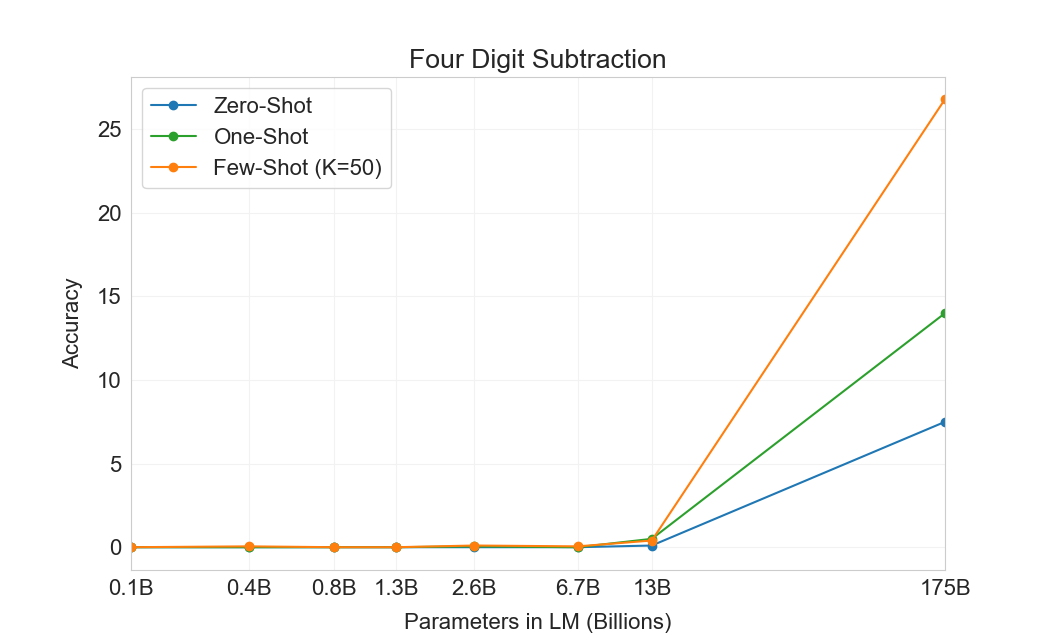} &  \includegraphics[width=0.33\linewidth]{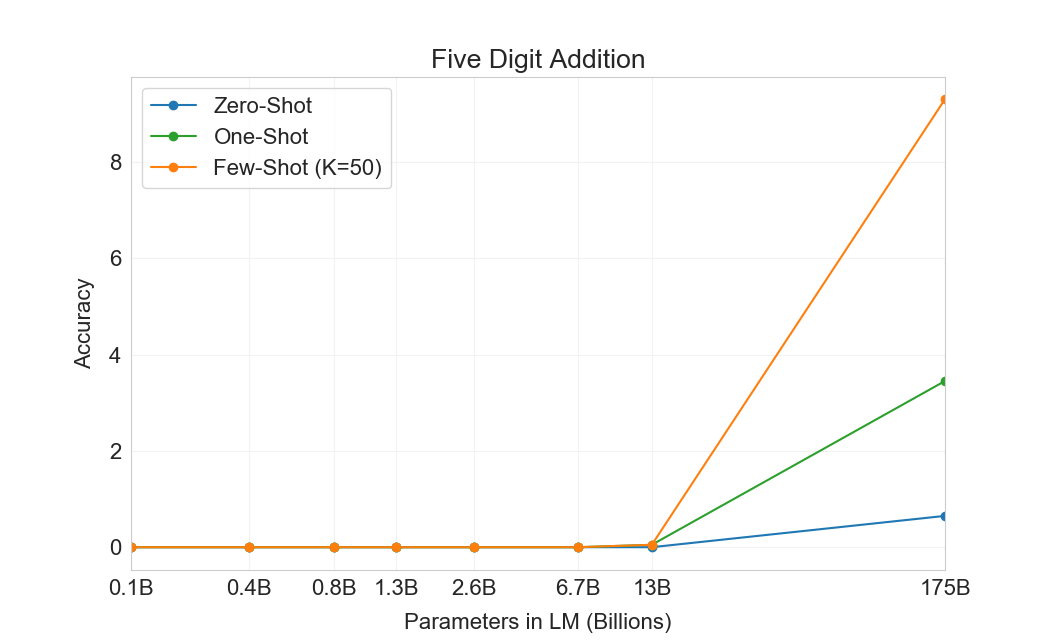} \\
 \includegraphics[width=0.33\linewidth]{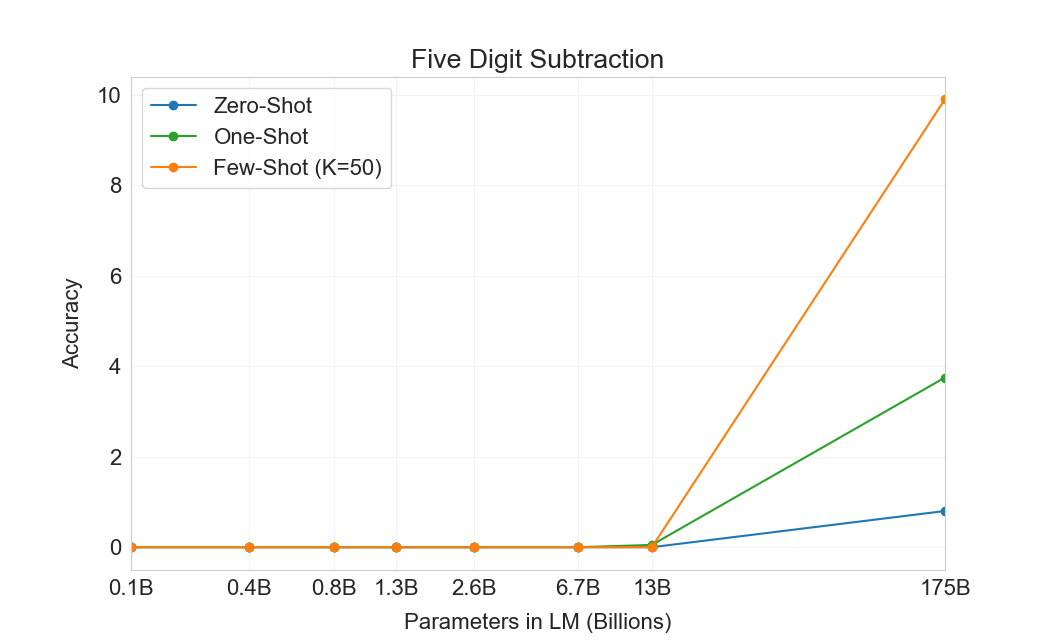} &   \includegraphics[width=0.33\linewidth]{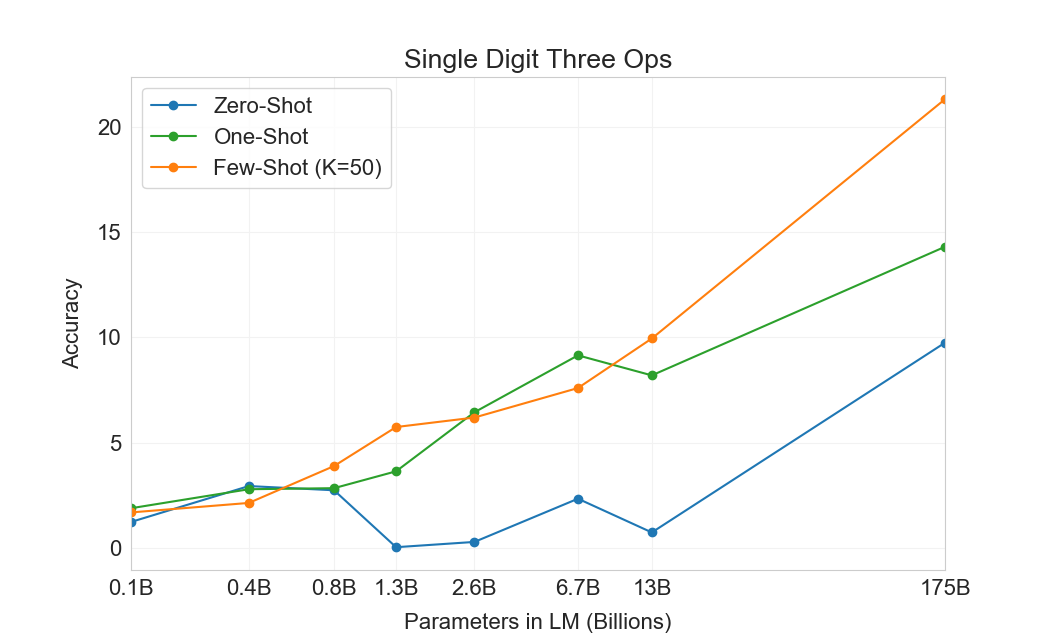} &                                                                                    \\
\bottomrule
\end{tabular}
\caption{All results for all Arithmetic tasks.}
\label{figure:all_arithmetic}
\end{figure}

\begin{figure}
\begin{tabular}{lll}
\toprule
      \includegraphics[width=0.33\linewidth]{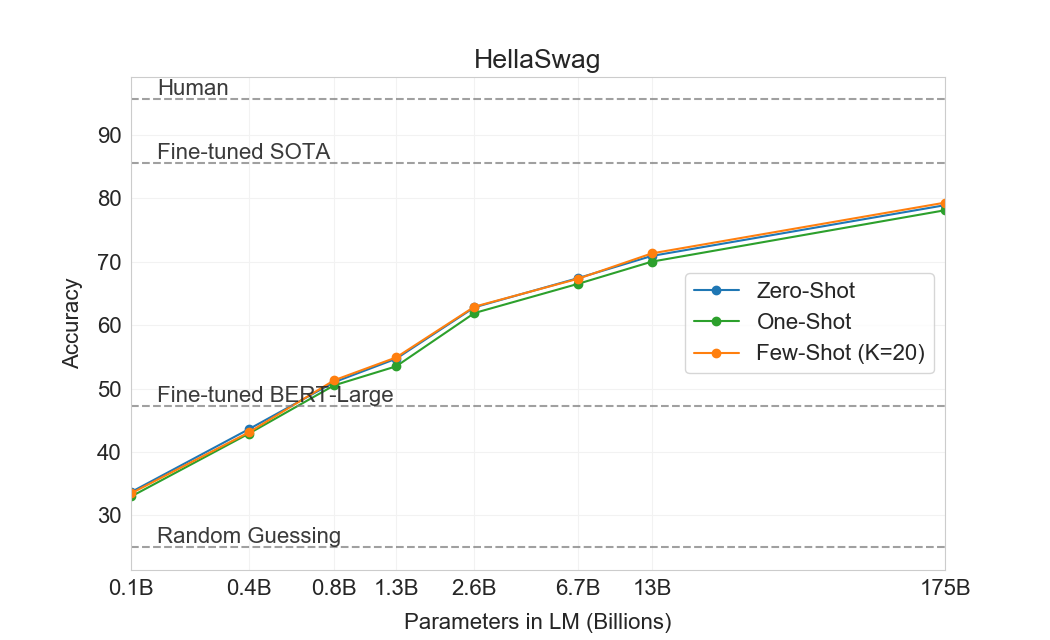} &      \includegraphics[width=0.33\linewidth]{graphs/scale_plots/lambada_acc_test.png} &               \includegraphics[width=0.33\linewidth]{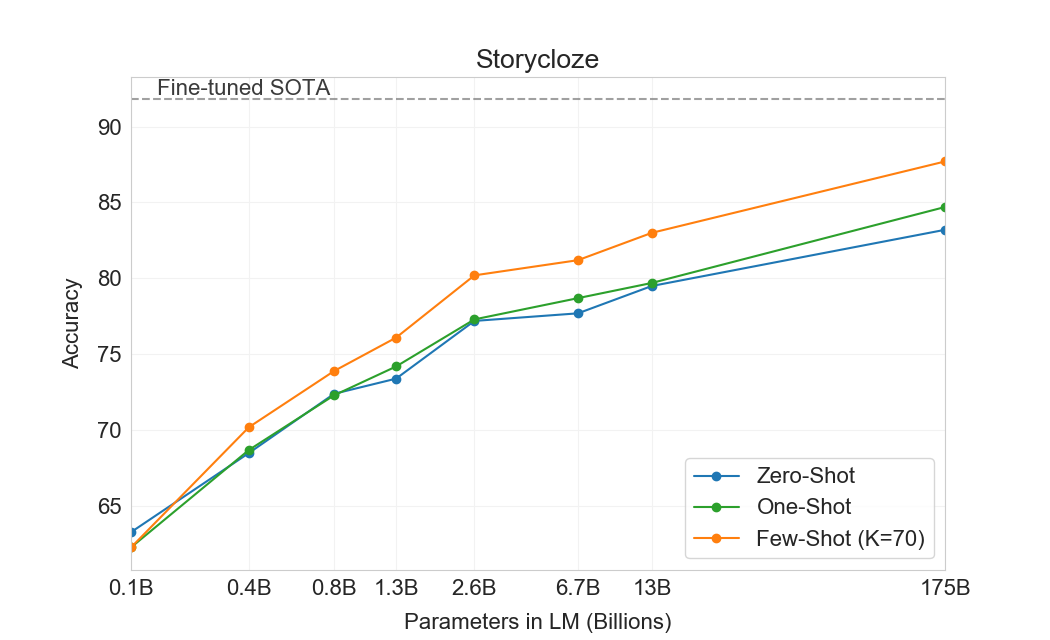} \\
\bottomrule
\end{tabular}
\caption{All results for all Cloze and Completion tasks.}
\label{graph:all_cloze_and_completion}
\end{figure}

\begin{figure}
\begin{tabular}{lll}
\toprule
      \includegraphics[width=0.33\linewidth]{graphs/scale_plots/piqa.png} &      \includegraphics[width=0.33\linewidth]{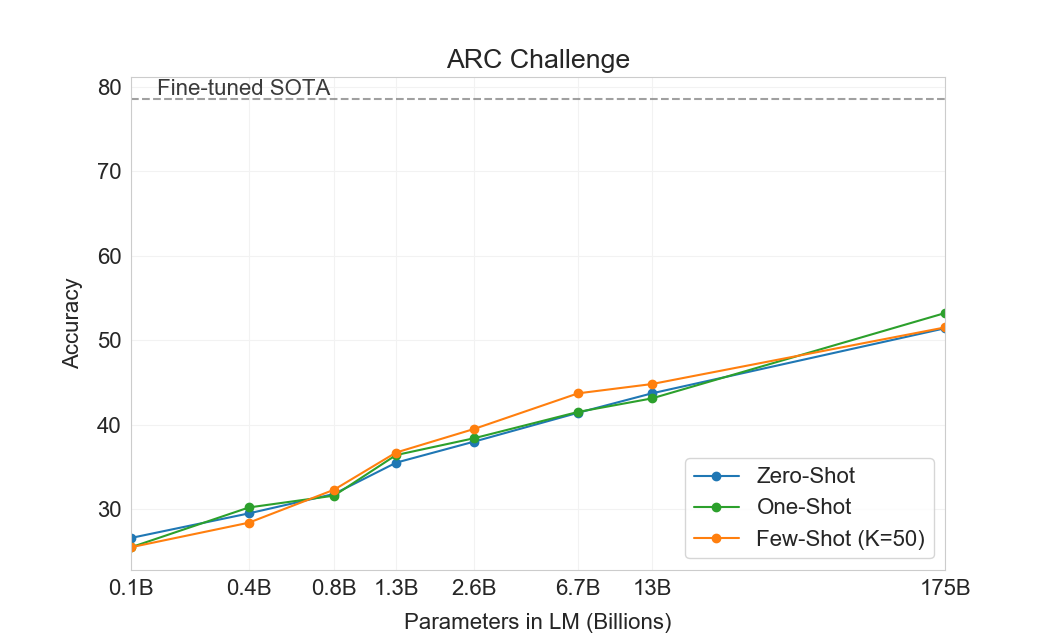} &        \includegraphics[width=0.33\linewidth]{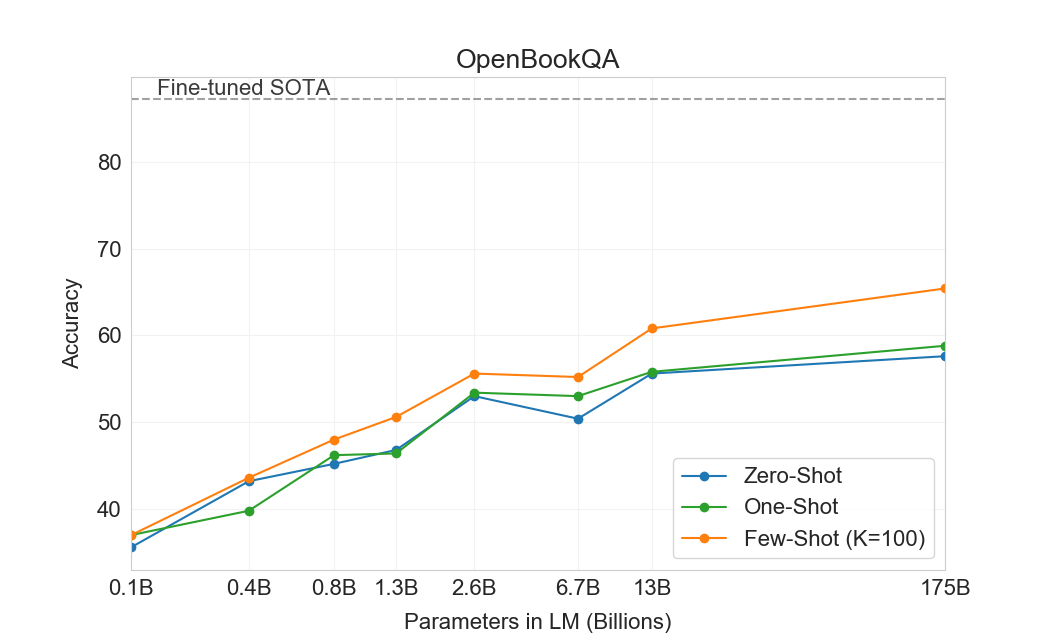} \\                                             \\
\bottomrule
\end{tabular}
\caption{All results for all Common Sense Reasoning tasks.}
\label{figure:all_common_sense}
\end{figure}

\begin{figure}
\begin{tabular}{lll}
\toprule
      \includegraphics[width=0.33\linewidth]{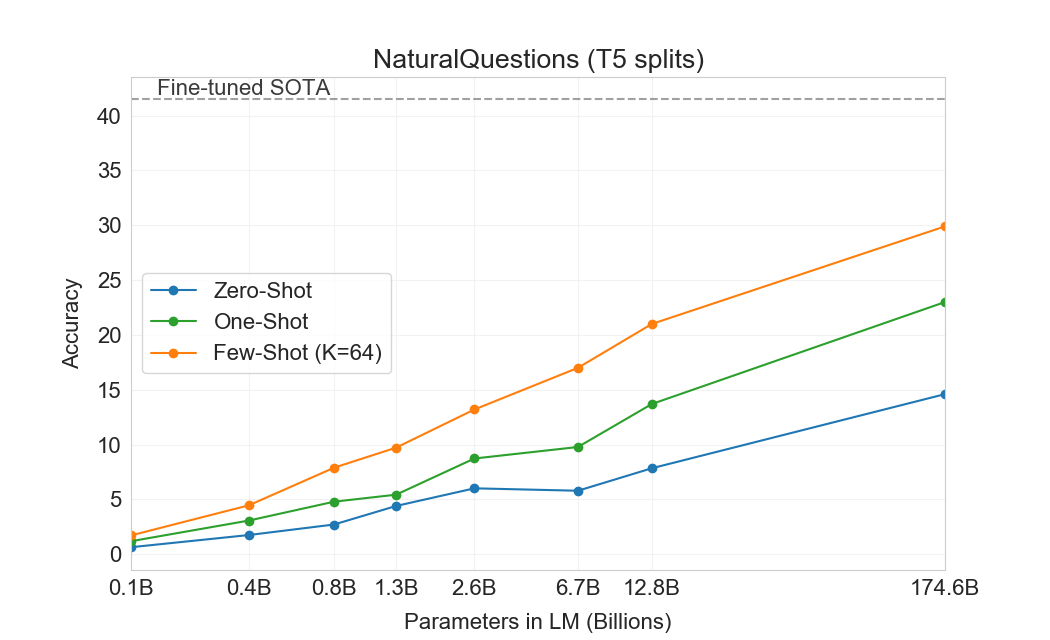} &      \includegraphics[width=0.33\linewidth]{graphs/scale_plots/triviaqa-wiki.png} &        \includegraphics[width=0.33\linewidth]{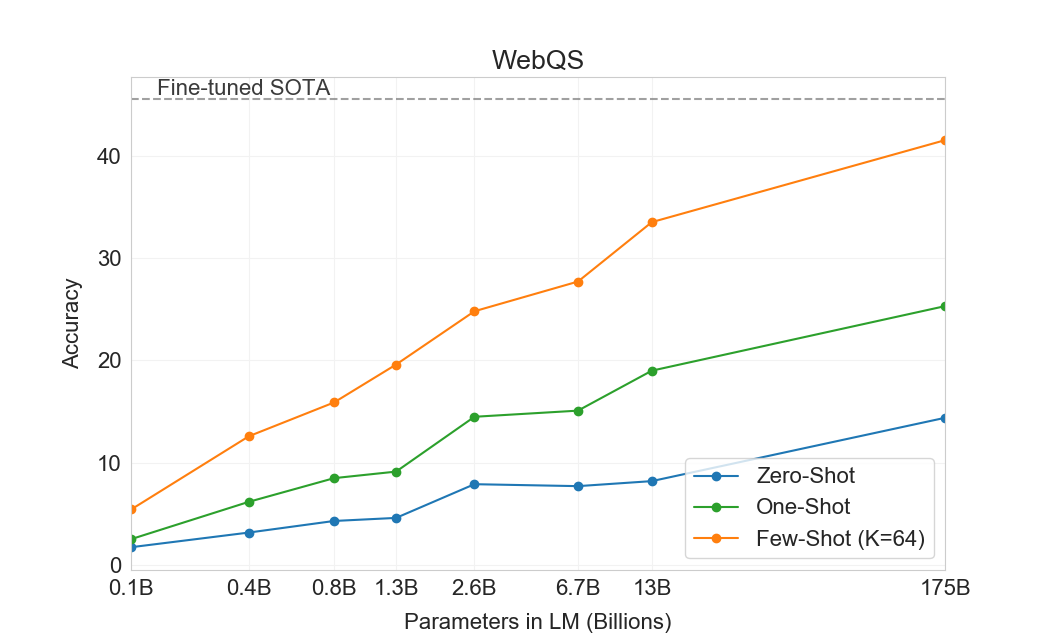} \\
\bottomrule
\end{tabular}
\caption{All results for all QA tasks.}
\label{graph:all_qa}
\end{figure}

\begin{figure}
\begin{tabular}{lll}
\toprule
      \includegraphics[width=0.33\linewidth]{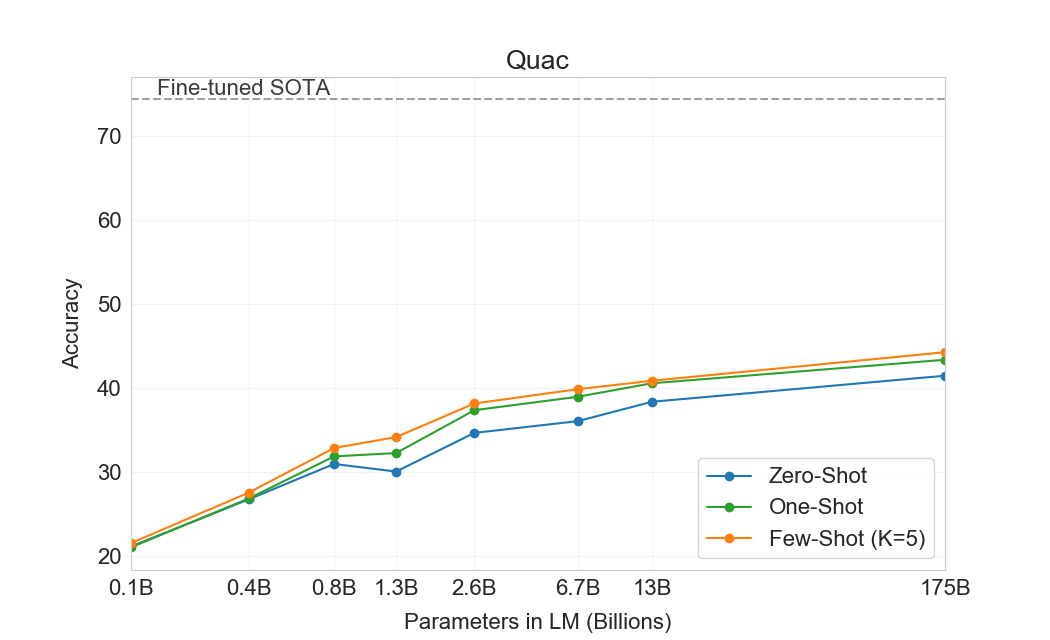} &      \includegraphics[width=0.33\linewidth]{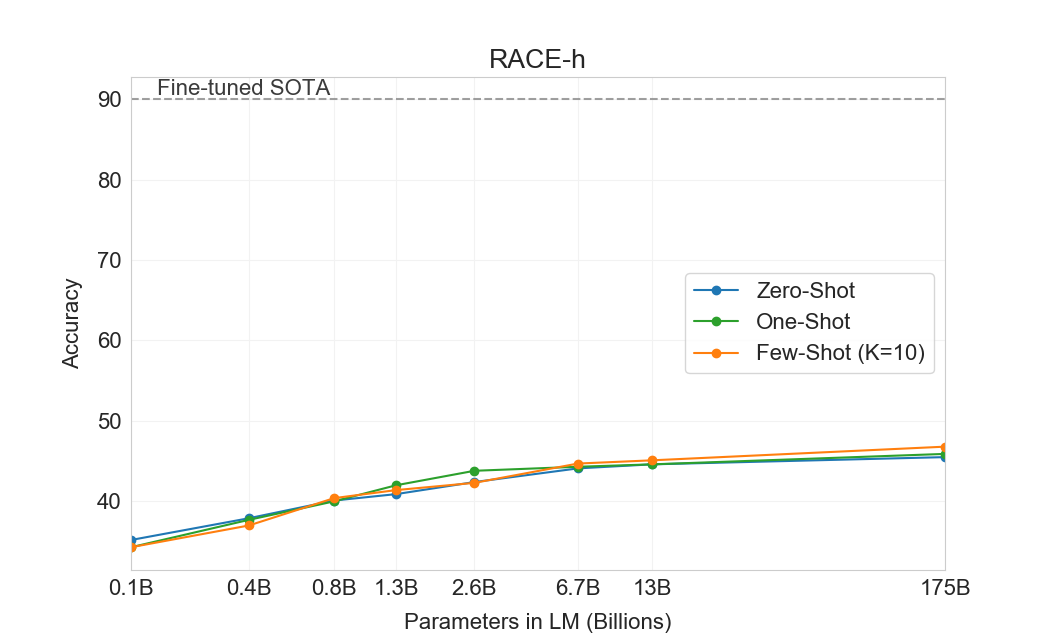} &        \includegraphics[width=0.33\linewidth]{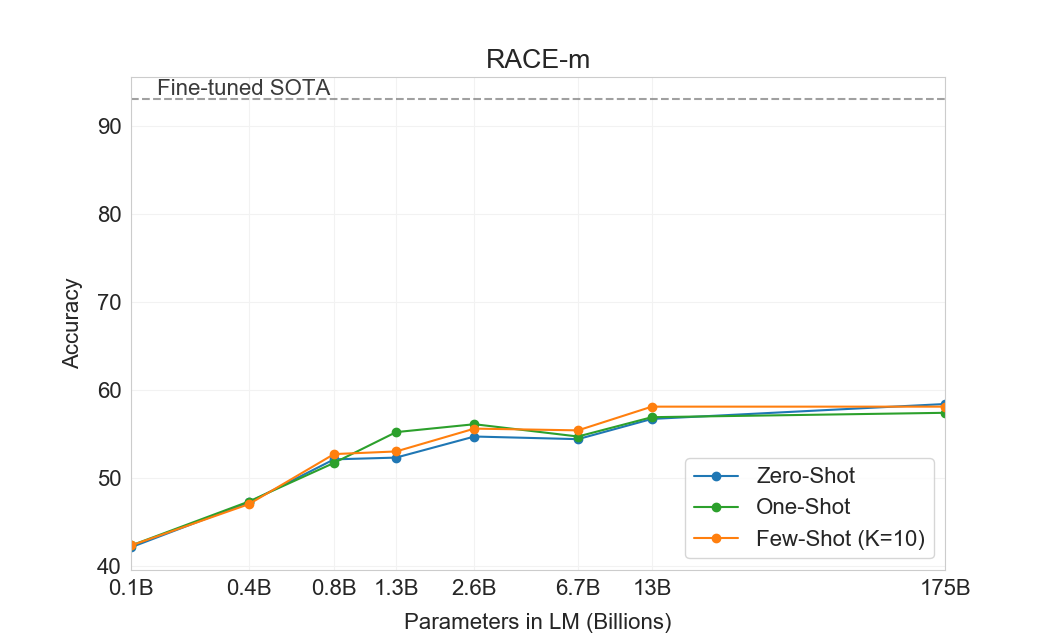} \\
       \includegraphics[width=0.33\linewidth]{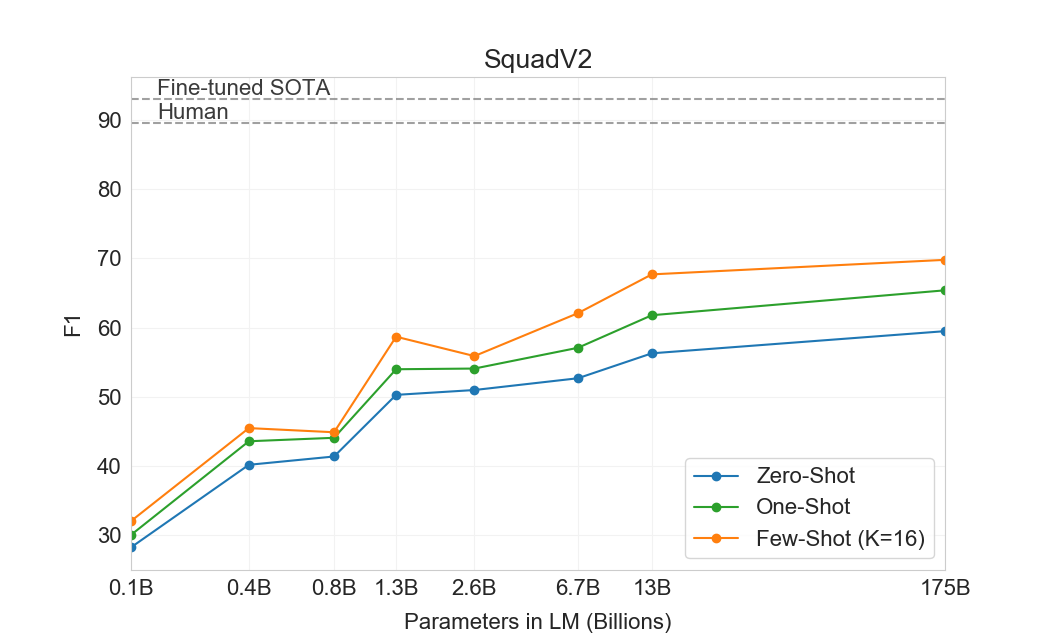} &         \includegraphics[width=0.33\linewidth]{graphs/scale_plots/coqa.png} &          \includegraphics[width=0.33\linewidth]{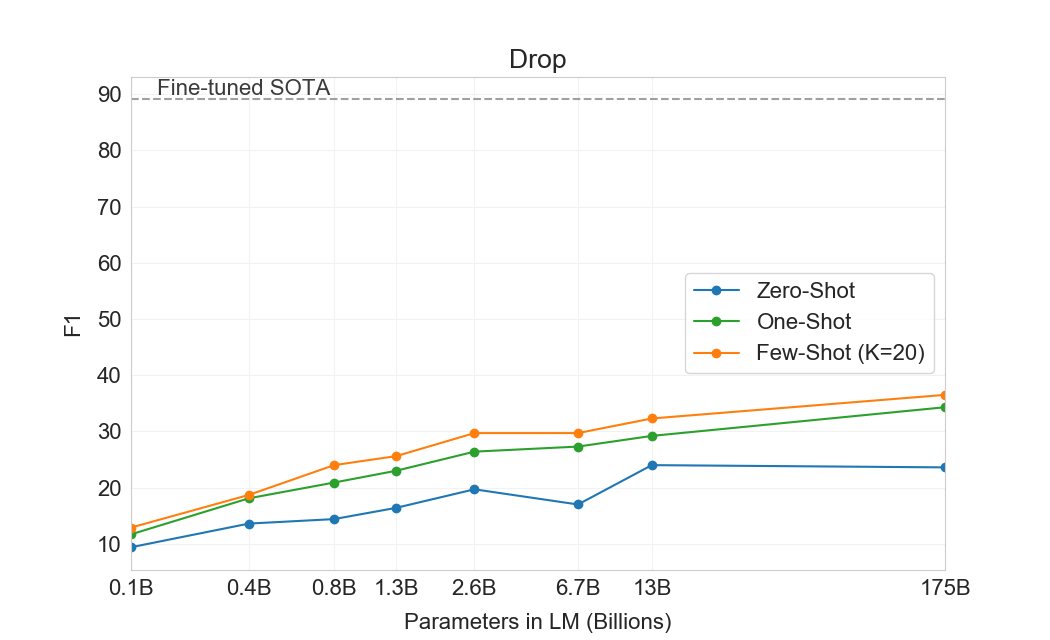} \\
\bottomrule
\end{tabular}
\caption{All results for all Reading Comprehension tasks.}
\label{figure:all_reading_comprehension}
\end{figure}

\begin{figure}
\begin{tabular}{lll}
\toprule
\includegraphics[width=0.31\linewidth]{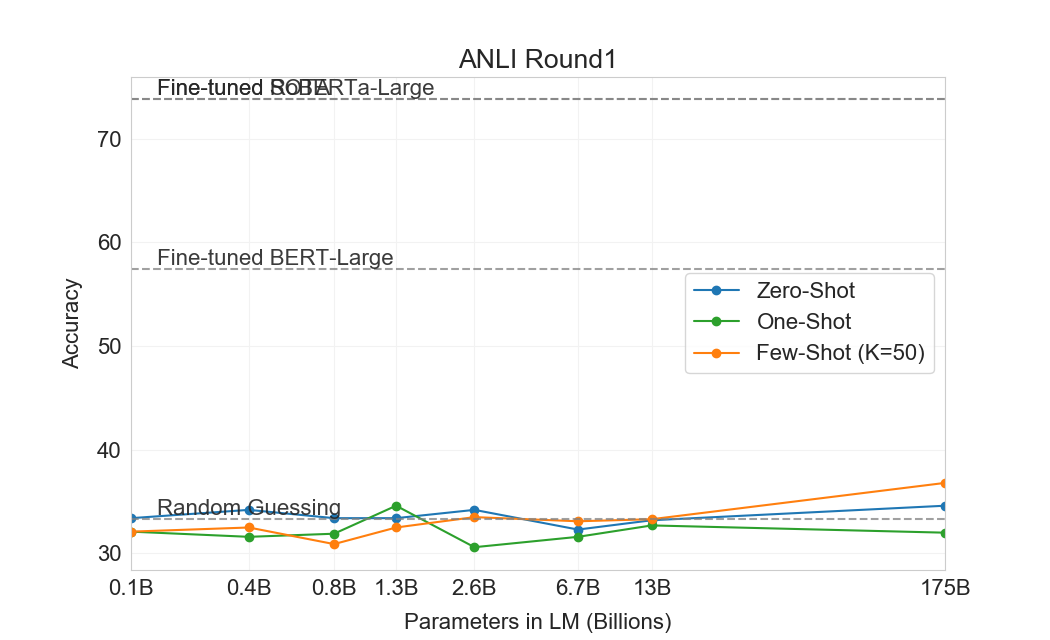} &       \includegraphics[width=0.31\linewidth]{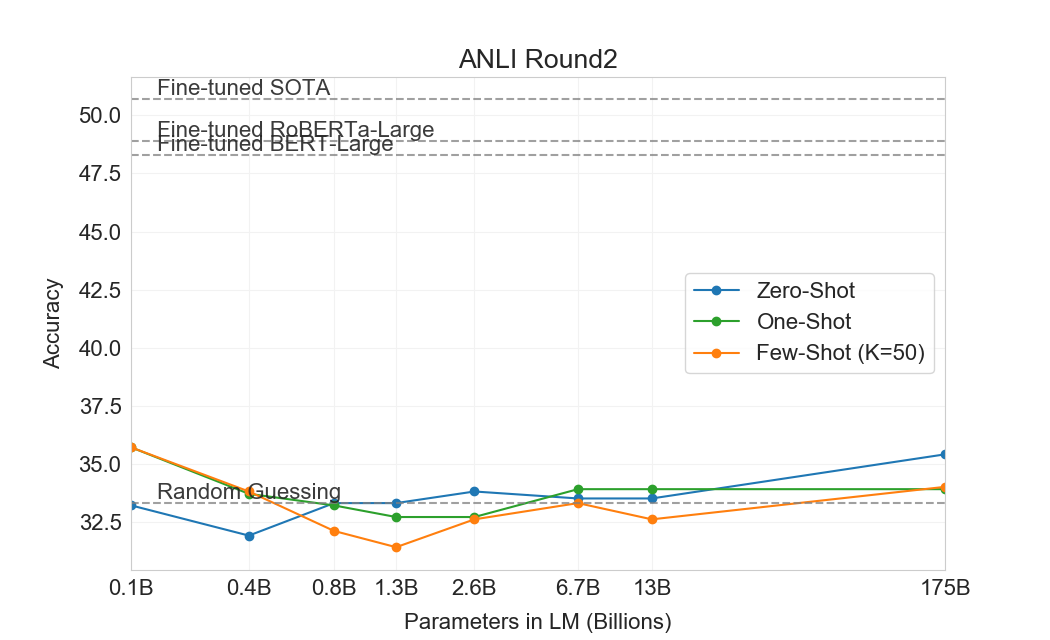} &        \includegraphics[width=0.31\linewidth]{graphs/scale_plots/anli_r3_test.png} \\
\bottomrule
\end{tabular}

\caption{All results for all ANLI rounds.}
\label{figure:all_anli}
\end{figure}

\begin{figure}
\begin{tabular}{lll}
\toprule
      \includegraphics[width=0.33\linewidth]{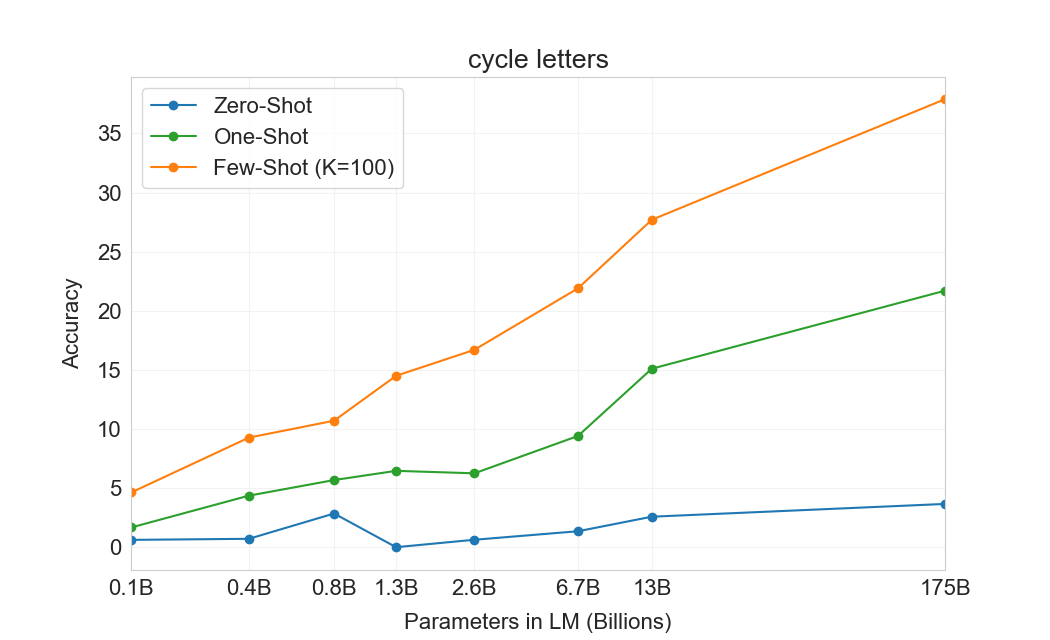} &      \includegraphics[width=0.33\linewidth]{graphs/scale_plots/wordscramble_fewshot.png} &        \includegraphics[width=0.33\linewidth]{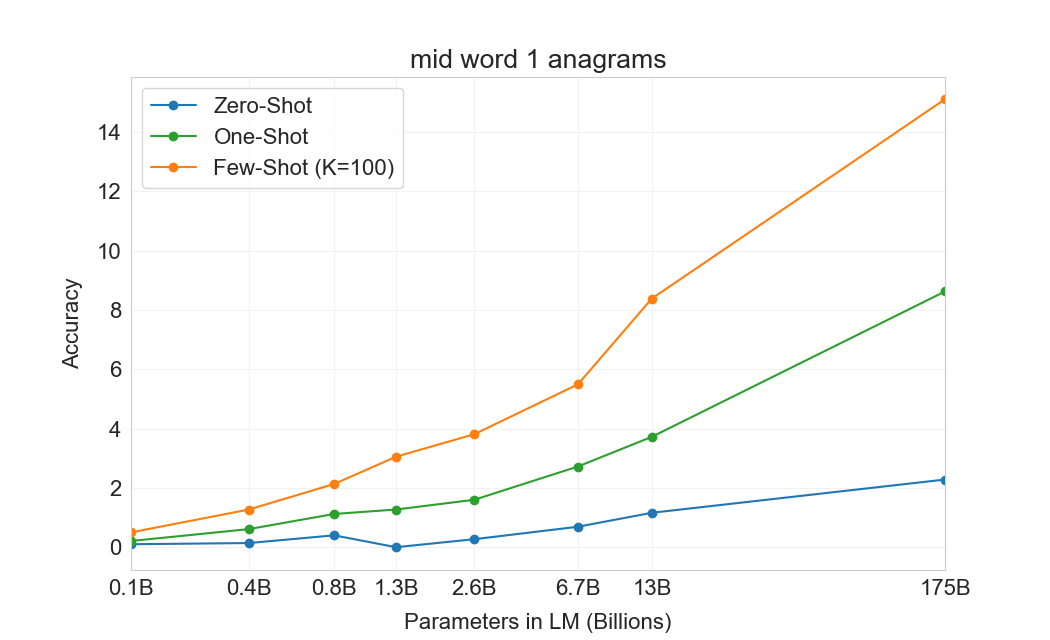} \\
       \includegraphics[width=0.33\linewidth]{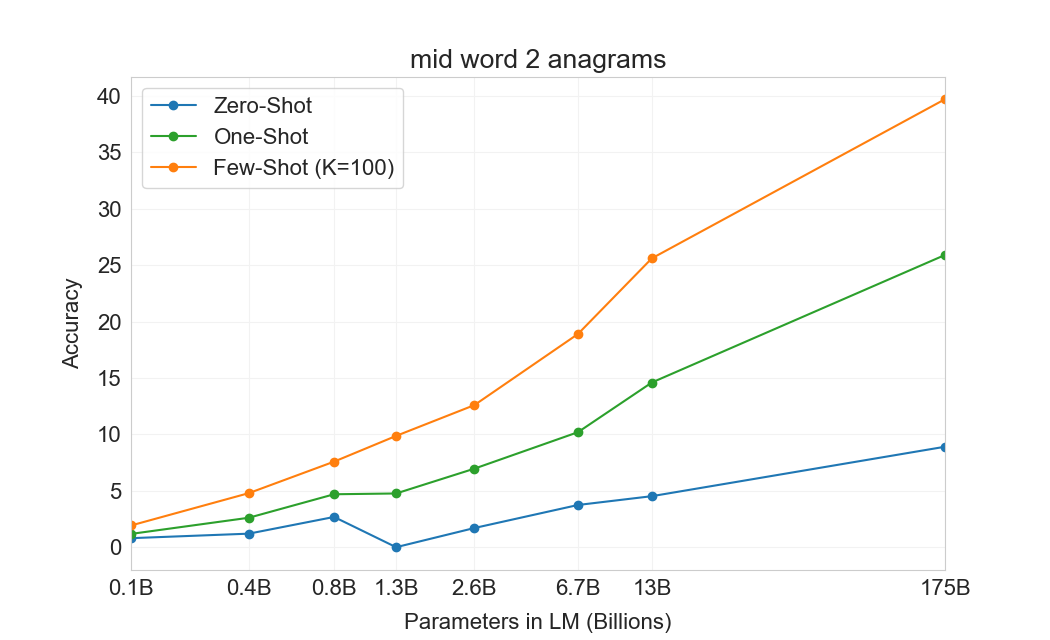} &         \includegraphics[width=0.33\linewidth]{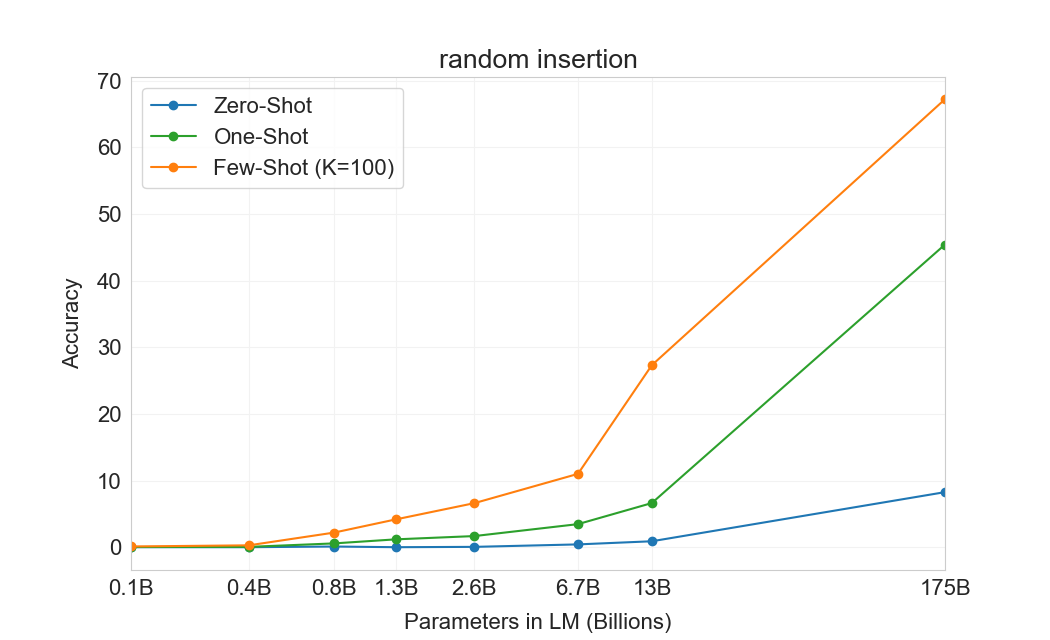} &          \includegraphics[width=0.33\linewidth]{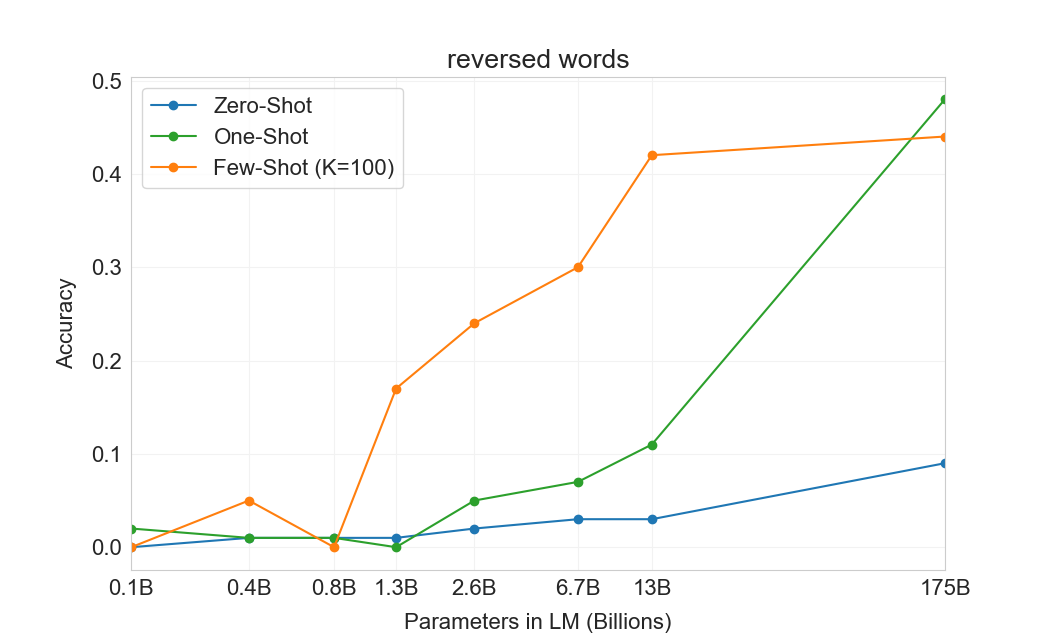} \\
\bottomrule
\end{tabular}
\caption{All results for all Scramble tasks.}
\label{figure:all_superglue}
\end{figure}

\begin{figure}
\begin{tabular}{lll}
\toprule
      \includegraphics[width=0.33\linewidth]{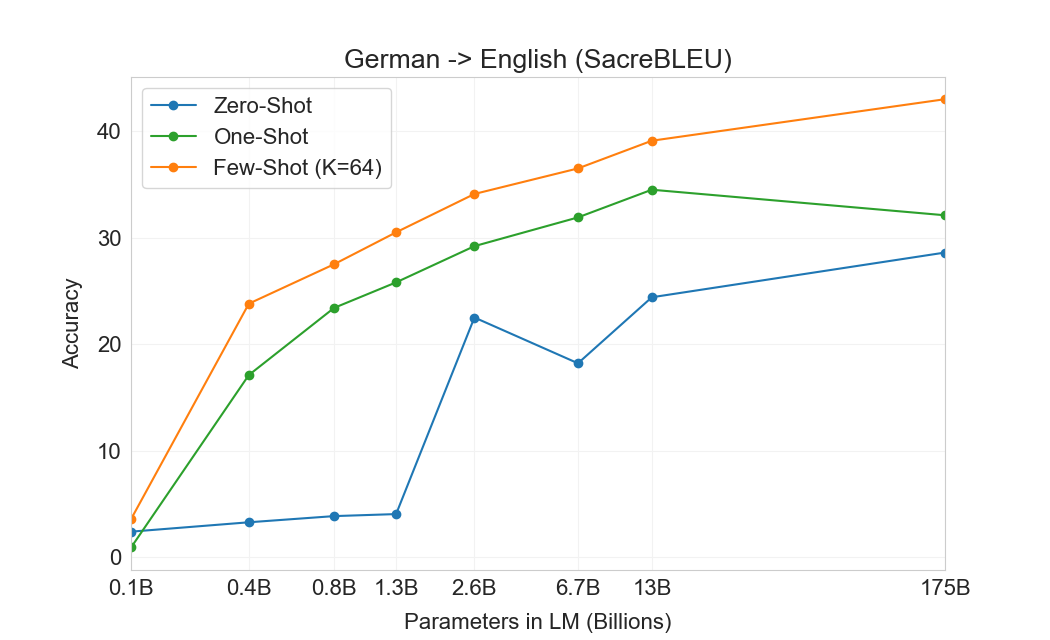} &      \includegraphics[width=0.33\linewidth]{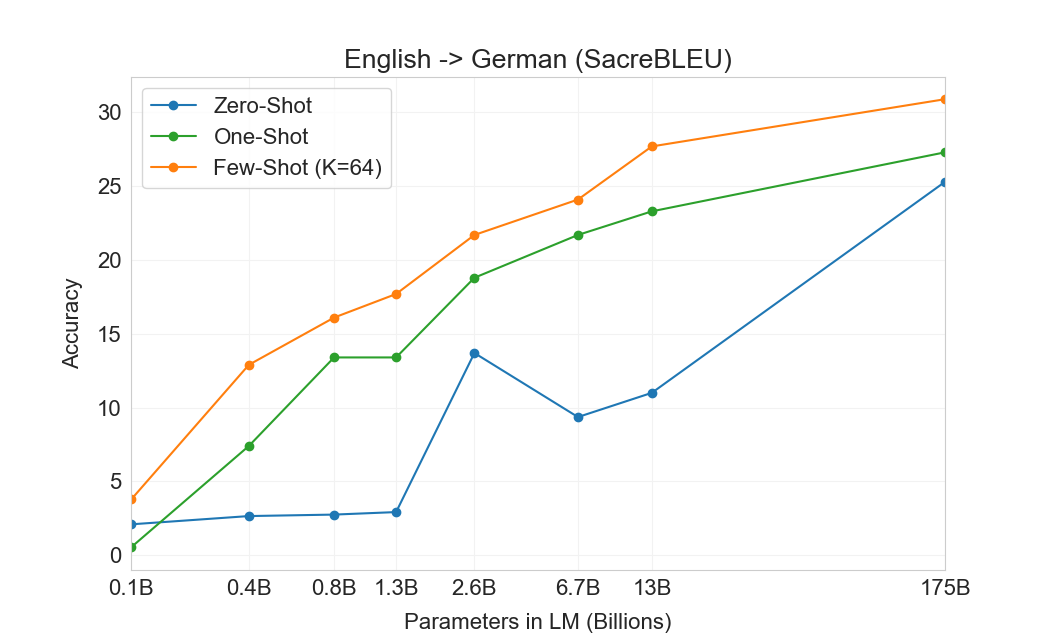} &        \includegraphics[width=0.33\linewidth]{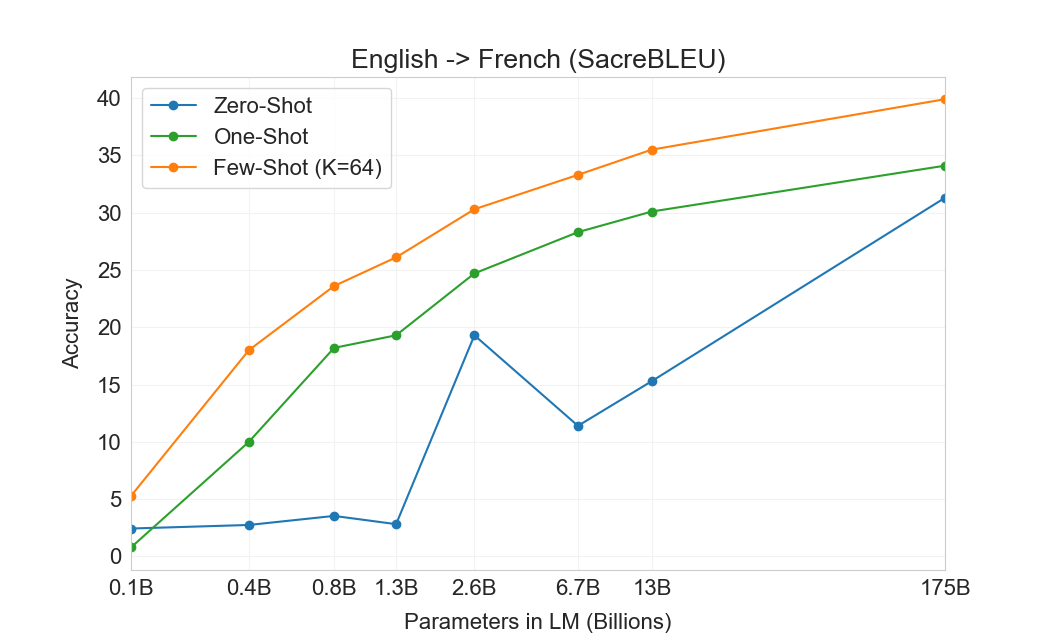} \\
       \includegraphics[width=0.33\linewidth]{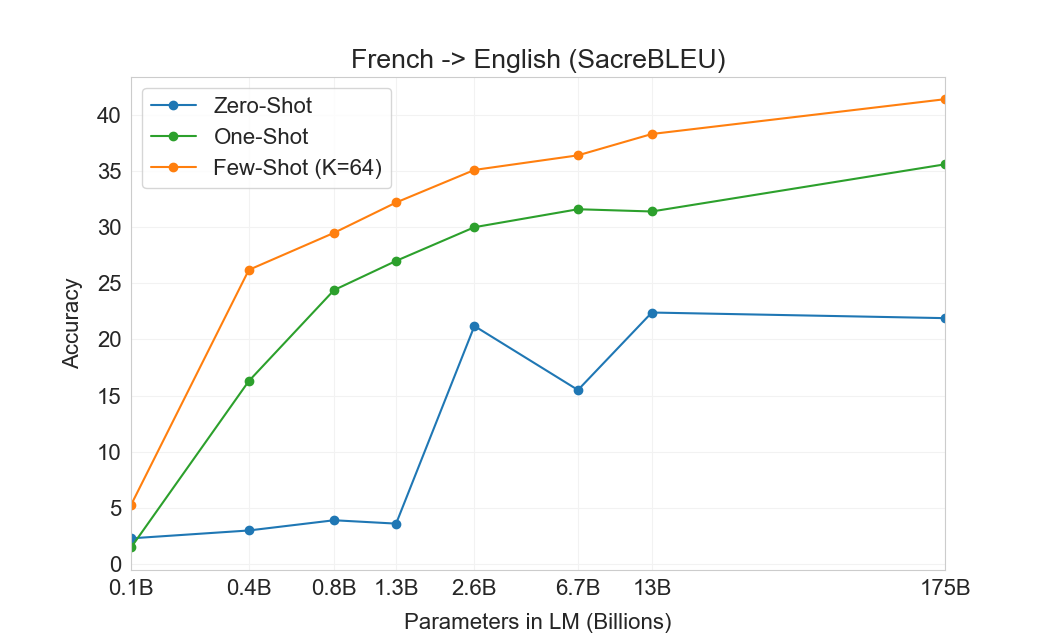} &         \includegraphics[width=0.33\linewidth]{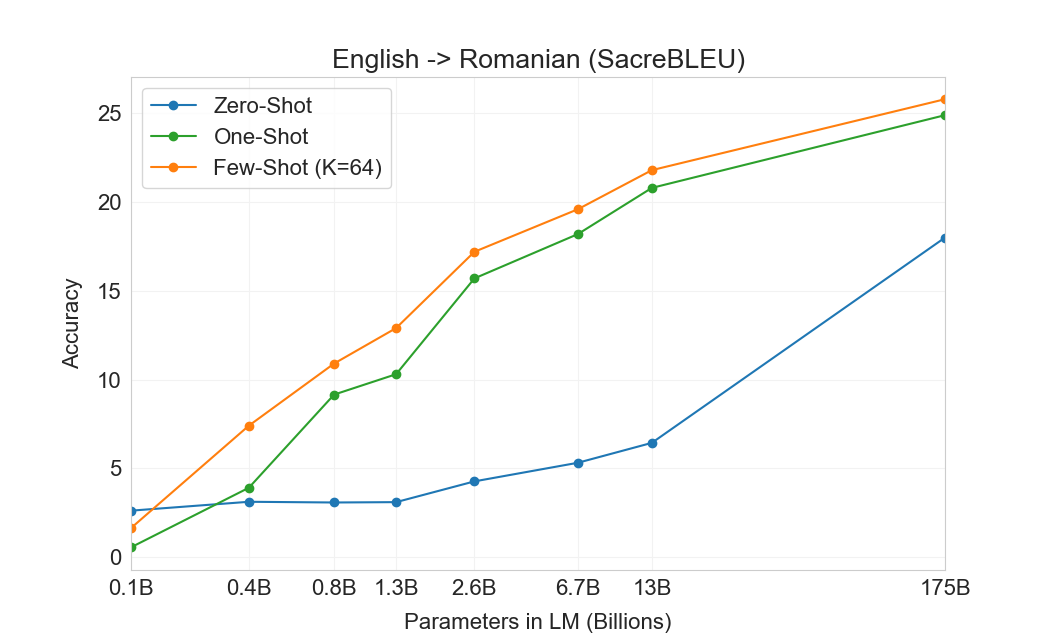} &          \includegraphics[width=0.33\linewidth]{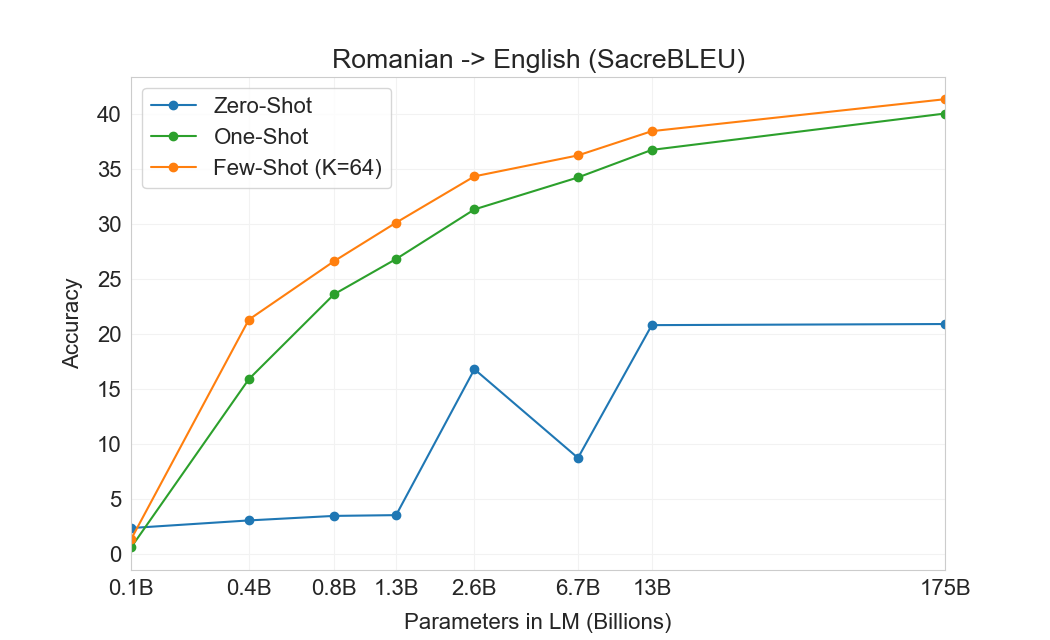} \\
       \includegraphics[width=0.33\linewidth]{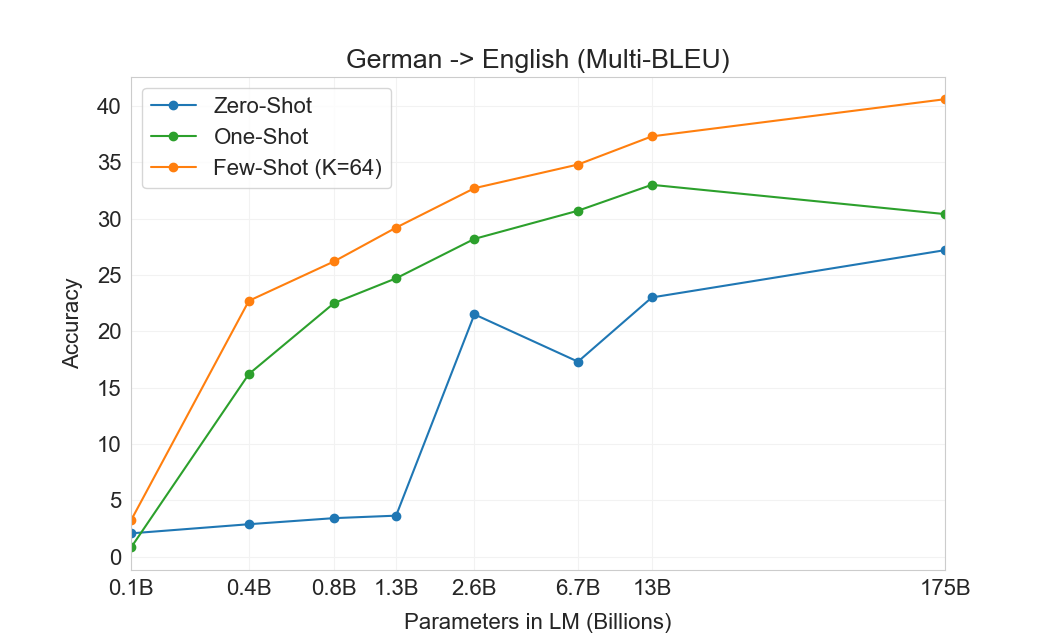} &      \includegraphics[width=0.33\linewidth]{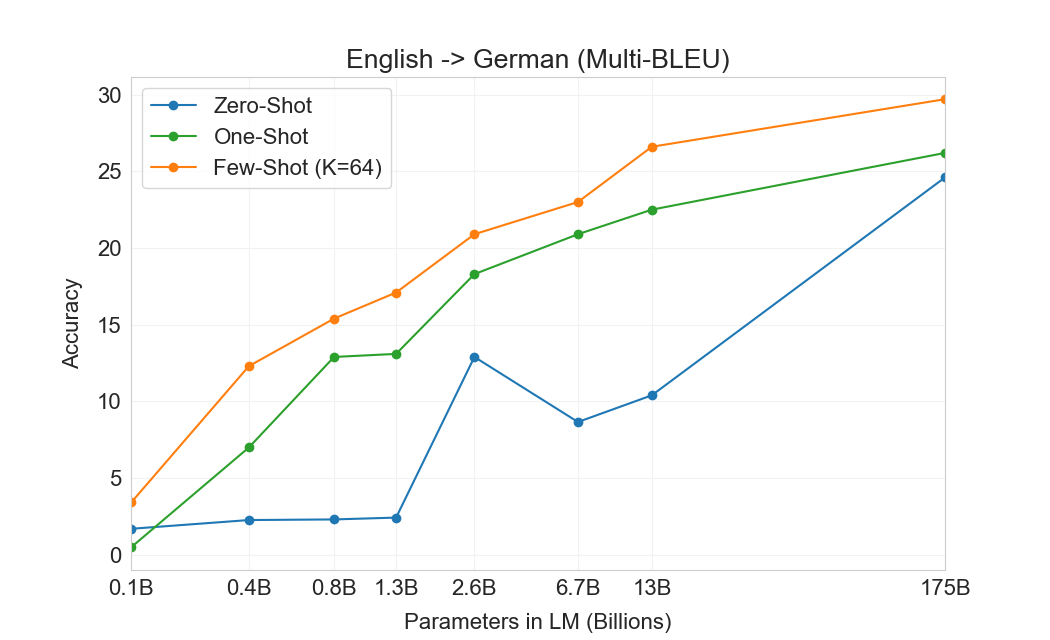} &        \includegraphics[width=0.33\linewidth]{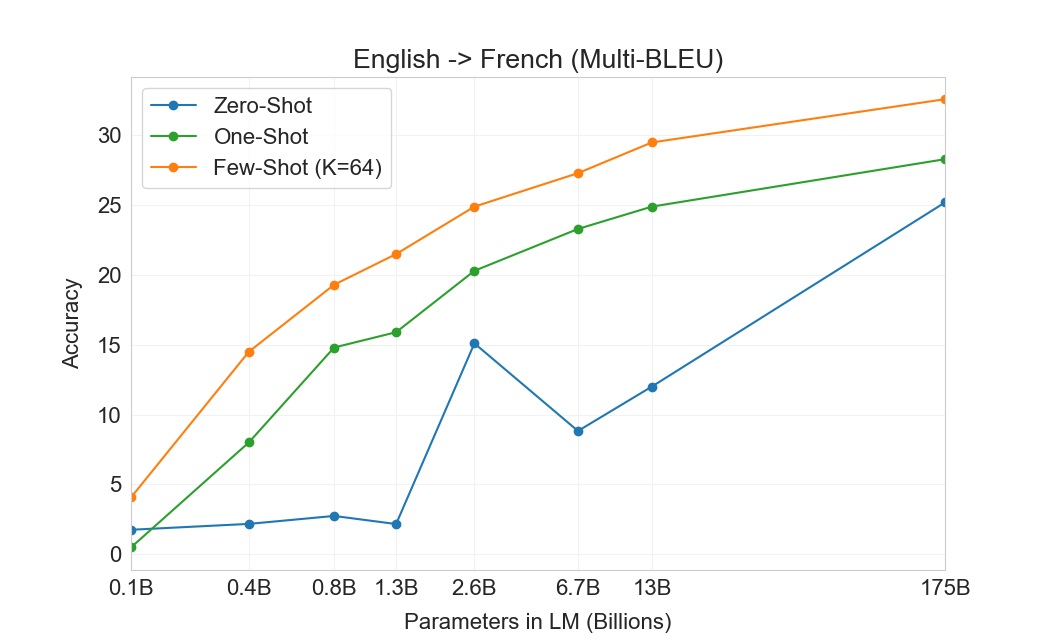} \\
       \includegraphics[width=0.33\linewidth]{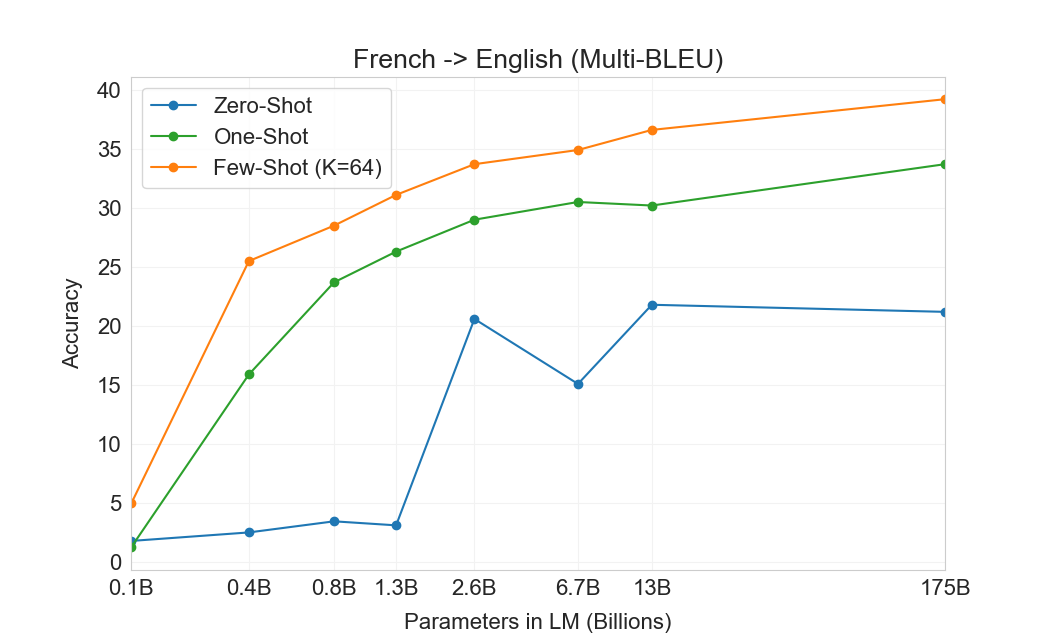} &         \includegraphics[width=0.33\linewidth]{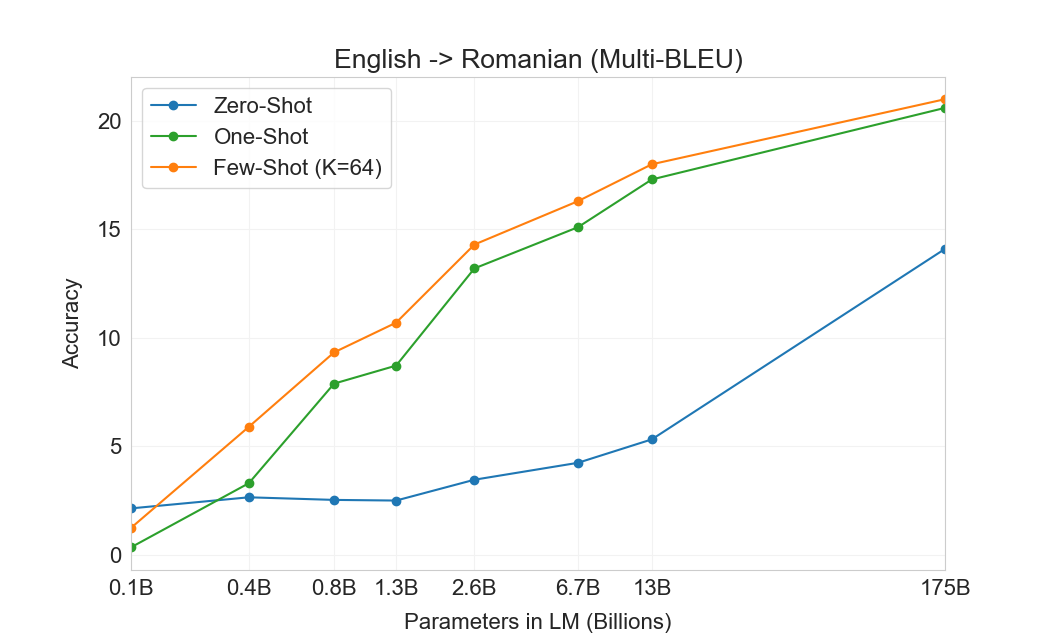} &          \includegraphics[width=0.33\linewidth]{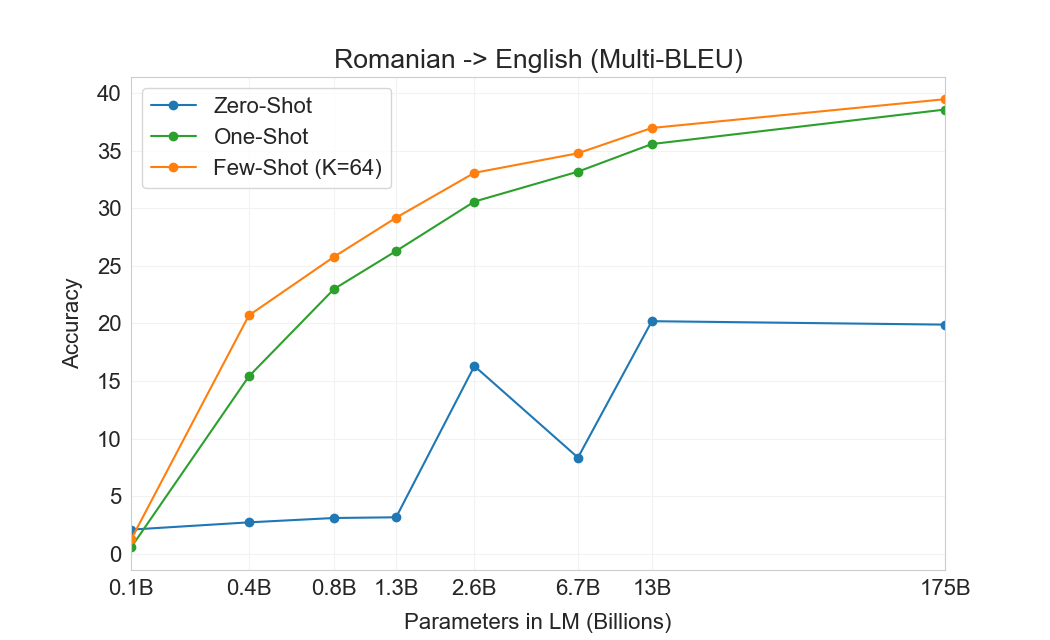} \\
\bottomrule
\end{tabular}
\caption{All results for all Translation tasks.}
\label{figure:all_translation}
\end{figure}

\clearpage
\newpage
\bibliographystyle{alpha}
\bibliography{bib}


\end{document}